\pdfoutput=1
\documentclass{article}

\usepackage{microtype}
\usepackage{graphicx}
\usepackage{booktabs}
\usepackage{array}
\newcolumntype{P}[1]{>{\raggedright\arraybackslash}p{#1}}
\usepackage{hyperref}

\usepackage[preprint]{icml2026}
\usepackage{amsmath}
\usepackage{amssymb}
\usepackage{mathtools}
\usepackage{amsthm}

\graphicspath{{figs/}}
\newcommand{\heldoutAggregateOutcome}{B}  %
\newcommand{\heldoutRareOutcome}{B}  %
\newcommand{\leakepsAOutcome}{NEITHER-SUPPORTED}  %
\newcommand{\leakepsBDoseFive}{NO-DATA}  %
\newcommand{\leakepsBDoseFour}{SOFTMAX-HIGHER-BEYOND-BAND}  %
\newcommand{\leakepsBDoseOne}{SOFTMAX-HIGHER-BEYOND-BAND}  %
\newcommand{\leakepsBDoseThree}{WITHIN-BAND}  %
\newcommand{\leakepsBDoseTwo}{SOFTMAX-HIGHER-BEYOND-BAND}  %
\newcommand{\mechPOneTwoOutcome}{FAIL}  %
\newcommand{\qBandRepEOne}{the same-host floor holds}  %
\newcommand{\qBandRepEOneB}{the same-host floor holds}  %
\newcommand{\qCohSeriesOutcome}{NOT-COVERED}  %
\newcommand{\qGTwoRerunEFive}{the pairing gate passes: a new, clean size point}  %
\newcommand{\qGTwoRerunEFiveB}{PAIRED}  %
\newcommand{\qGTwoRerunSize}{same order}  %
\newcommand{\qNfExpFOutcome}{FLOOR-FAILS}  %
\newcommand{\qNfShareOutcome}{FLOOR-FAILS}  %
\newcommand{\qRowEpsHZeroSeedThree}{DENOMINATOR-DIRECT}  %
\newcommand{\qRowEpsSeedFive}{DENOMINATOR-DIRECT}  %
\newcommand{\qRowEpsSeedFour}{DENOMINATOR-DIRECT}  %
\newcommand{\qRowEpsSeedThree}{DENOMINATOR-DIRECT}  %
\newcommand{\qRowEpsSeedZero}{DENOMINATOR-DIRECT}  %
\newcommand{\qShadowXThreePBceCells}{12}  %
\newcommand{\qShadowXThreePBceCellsOut}{12}  %
\newcommand{\qShadowXThreePBceFifteenFourDecadeForwardSeedOne}{OUT-OF-RANGE}  %
\newcommand{\qShadowXThreePBceFifteenFourDecadeForwardSeedTwo}{OUT-OF-RANGE}  %
\newcommand{\qShadowXThreePBceFifteenFourDecadeForwardSeedZero}{OUT-OF-RANGE}  %
\newcommand{\qShadowXThreePBceFifteenFourDecadeReverseSeedOne}{OUT-OF-RANGE}  %
\newcommand{\qShadowXThreePBceFifteenFourDecadeReverseSeedTwo}{OUT-OF-RANGE}  %
\newcommand{\qShadowXThreePBceFifteenFourDecadeReverseSeedZero}{OUT-OF-RANGE}  %
\newcommand{\qShadowXThreePBceFifteenThreeDecadeForwardSeedOne}{OUT-OF-RANGE}  %
\newcommand{\qShadowXThreePBceFifteenThreeDecadeForwardSeedTwo}{OUT-OF-RANGE}  %
\newcommand{\qShadowXThreePBceFifteenThreeDecadeForwardSeedZero}{OUT-OF-RANGE}  %
\newcommand{\qShadowXThreePBceFifteenThreeDecadeReverseSeedOne}{OUT-OF-RANGE}  %
\newcommand{\qShadowXThreePBceFifteenThreeDecadeReverseSeedTwo}{OUT-OF-RANGE}  %
\newcommand{\qShadowXThreePBceFifteenThreeDecadeReverseSeedZero}{OUT-OF-RANGE}  %
\newcommand{\qShadowXThreePCvalBceFifteenFourDecadeForwardSeedOne}{1.2924}  %
\newcommand{\qShadowXThreePCvalBceFifteenFourDecadeForwardSeedTwo}{1.2598}  %
\newcommand{\qShadowXThreePCvalBceFifteenFourDecadeForwardSeedZero}{1.2966}  %
\newcommand{\qShadowXThreePCvalBceFifteenFourDecadeReverseSeedOne}{1.3461}  %
\newcommand{\qShadowXThreePCvalBceFifteenFourDecadeReverseSeedTwo}{1.3382}  %
\newcommand{\qShadowXThreePCvalBceFifteenFourDecadeReverseSeedZero}{1.3937}  %
\newcommand{\qShadowXThreePCvalBceFifteenThreeDecadeForwardSeedOne}{1.3431}  %
\newcommand{\qShadowXThreePCvalBceFifteenThreeDecadeForwardSeedTwo}{1.3258}  %
\newcommand{\qShadowXThreePCvalBceFifteenThreeDecadeForwardSeedZero}{1.4131}  %
\newcommand{\qShadowXThreePCvalBceFifteenThreeDecadeReverseSeedOne}{1.2737}  %
\newcommand{\qShadowXThreePCvalBceFifteenThreeDecadeReverseSeedTwo}{1.2687}  %
\newcommand{\qShadowXThreePCvalBceFifteenThreeDecadeReverseSeedZero}{1.4138}  %
\newcommand{\qShadowXThreePCvalSoftmaxFourDecadeForwardSeedOne}{1.3729}  %
\newcommand{\qShadowXThreePCvalSoftmaxFourDecadeForwardSeedTwo}{1.3333}  %
\newcommand{\qShadowXThreePCvalSoftmaxFourDecadeForwardSeedZero}{1.3200}  %
\newcommand{\qShadowXThreePCvalSoftmaxFourDecadeReverseSeedOne}{1.2047}  %
\newcommand{\qShadowXThreePCvalSoftmaxFourDecadeReverseSeedTwo}{1.2071}  %
\newcommand{\qShadowXThreePCvalSoftmaxFourDecadeReverseSeedZero}{1.1843}  %
\newcommand{\qShadowXThreePOutAbove}{1}  %
\newcommand{\qShadowXThreePRowCvalPrimeForwardSeedOne}{1.0110}  %
\newcommand{\qShadowXThreePRowCvalPrimeForwardSeedTwo}{1.0046}  %
\newcommand{\qShadowXThreePRowCvalPrimeForwardSeedZero}{1.0053}  %
\newcommand{\qShadowXThreePRowCvalPrimeReverseSeedOne}{0.9999}  %
\newcommand{\qShadowXThreePRowCvalPrimeReverseSeedTwo}{0.9997}  %
\newcommand{\qShadowXThreePRowCvalPrimeReverseSeedZero}{0.9998}  %
\newcommand{\qShadowXThreePRowForwardSeedOne}{FULL}  %
\newcommand{\qShadowXThreePRowForwardSeedTwo}{FULL}  %
\newcommand{\qShadowXThreePRowForwardSeedZero}{FULL}  %
\newcommand{\qShadowXThreePRowFullHi}{1.1}  %
\newcommand{\qShadowXThreePRowFullLo}{0.909}  %
\newcommand{\qShadowXThreePRowHiLinkComparedSeedOne}{214,359}  %
\newcommand{\qShadowXThreePRowHiLinkComparedSeedTwo}{214,359}  %
\newcommand{\qShadowXThreePRowHiLinkComparedSeedZero}{214,359}  %
\newcommand{\qShadowXThreePRowHiLinkDiffersSeedOne}{0}  %
\newcommand{\qShadowXThreePRowHiLinkDiffersSeedTwo}{0}  %
\newcommand{\qShadowXThreePRowHiLinkDiffersSeedZero}{0}  %
\newcommand{\qShadowXThreePRowHiLinkSeedOne}{IDENTICAL}  %
\newcommand{\qShadowXThreePRowHiLinkSeedTwo}{IDENTICAL}  %
\newcommand{\qShadowXThreePRowHiLinkSeedZero}{IDENTICAL}  %
\newcommand{\qShadowXThreePRowNoneLo}{0}  %
\newcommand{\qShadowXThreePRowOverAbove}{1.1}  %
\newcommand{\qShadowXThreePRowPartialLo}{0.25}  %
\newcommand{\qShadowXThreePRowReverseSeedOne}{FULL}  %
\newcommand{\qShadowXThreePRowReverseSeedTwo}{FULL}  %
\newcommand{\qShadowXThreePRowReverseSeedZero}{FULL}  %
\newcommand{\qShadowXThreePRowSetFull}{FULL}  %
\newcommand{\qShadowXThreePRowSetNone}{NONE}  %
\newcommand{\qShadowXThreePRowSetOver}{OVER}  %
\newcommand{\qShadowXThreePRowSetPartial}{PARTIAL}  %
\newcommand{\qShadowXThreePRowSetUndefined}{UNDEFINED}  %
\newcommand{\qShadowXThreePRowSetUnder}{UNDER}  %
\newcommand{\qShadowXThreePRowUndefFactor}{1.1}  %
\newcommand{\qShadowXThreePRowUnderBelow}{0}  %
\newcommand{\qShadowXThreePSetOutOfRange}{OUT-OF-RANGE}  %
\newcommand{\qShadowXThreePSoftmaxCells}{6}  %
\newcommand{\qShadowXThreePSoftmaxCellsOut}{6}  %
\newcommand{\qShadowXThreePSoftmaxFourDecadeForwardSeedOne}{OUT-OF-RANGE}  %
\newcommand{\qShadowXThreePSoftmaxFourDecadeForwardSeedTwo}{OUT-OF-RANGE}  %
\newcommand{\qShadowXThreePSoftmaxFourDecadeForwardSeedZero}{OUT-OF-RANGE}  %
\newcommand{\qShadowXThreePSoftmaxFourDecadeReverseSeedOne}{OUT-OF-RANGE}  %
\newcommand{\qShadowXThreePSoftmaxFourDecadeReverseSeedTwo}{OUT-OF-RANGE}  %
\newcommand{\qShadowXThreePSoftmaxFourDecadeReverseSeedZero}{OUT-OF-RANGE}  %
\newcommand{\qShadowXThreePUndefFactor}{1.1}  %
\newcommand{\qShadowXThreePXNineCvalForwardSeedOne}{1.0036}  %
\newcommand{\qShadowXThreePXNineCvalForwardSeedTwo}{1.0021}  %
\newcommand{\qShadowXThreePXNineCvalForwardSeedZero}{0.9894}  %
\newcommand{\qShadowXThreePXNineCvalReverseSeedOne}{1.0086}  %
\newcommand{\qShadowXThreePXNineCvalReverseSeedTwo}{1.0040}  %
\newcommand{\qShadowXThreePXNineCvalReverseSeedZero}{0.9939}  %
\newcommand{\qShadowXThreePXNineForwardSeedOne}{OUT-OF-RANGE}  %
\newcommand{\qShadowXThreePXNineForwardSeedTwo}{OUT-OF-RANGE}  %
\newcommand{\qShadowXThreePXNineForwardSeedZero}{LOCAL}  %
\newcommand{\qShadowXThreePXNineGlobalLe}{0.25}  %
\newcommand{\qShadowXThreePXNineLocalGe}{0.75}  %
\newcommand{\qShadowXThreePXNineOutAbove}{1}  %
\newcommand{\qShadowXThreePXNineOutBelow}{0}  %
\newcommand{\qShadowXThreePXNineReverseSeedOne}{OUT-OF-RANGE}  %
\newcommand{\qShadowXThreePXNineReverseSeedTwo}{OUT-OF-RANGE}  %
\newcommand{\qShadowXThreePXNineReverseSeedZero}{LOCAL}  %
\newcommand{\qShadowXThreePXNineSetGlobal}{GLOBAL}  %
\newcommand{\qShadowXThreePXNineSetLocal}{LOCAL}  %
\newcommand{\qShadowXThreePXNineSetMixed}{MIXED}  %
\newcommand{\qShadowXThreePXNineSetOutOfRange}{OUT-OF-RANGE}  %
\newcommand{\qShadowXThreePXNineSetUndefined}{UNDEFINED}  %
\newcommand{\qShareModeETwo}{under eager, rung 1's 0.10 criterion is wider than repeat noise}  %
\newcommand{\qStdcfgBatch}{16}  %
\newcommand{\qStdcfgCertLeaves}{220,925}  %
\newcommand{\qStdcfgCertUndeclared}{zero}  %
\newcommand{\qStdcfgDModel}{512}  %
\newcommand{\qStdcfgHeads}{8}  %
\newcommand{\qStdcfgLayers}{6}  %
\newcommand{\qStdcfgLr}{SELECTED}  %
\newcommand{\qStdcfgLrArgv}{0.0003}  %
\newcommand{\qStdcfgLrCandOne}{0.0001}  %
\newcommand{\qStdcfgLrCandThree}{0.0003}  %
\newcommand{\qStdcfgLrCandTwo}{0.0002}  %
\newcommand{\qStdcfgLrMinFrac}{0.1}  %
\newcommand{\qStdcfgLrSetNone}{LR-NONE}  %
\newcommand{\qStdcfgLrSetRefused}{REFUSED}  %
\newcommand{\qStdcfgLrSetSelected}{SELECTED}  %
\newcommand{\qStdcfgLrSteps}{20,000}  %
\newcommand{\qStdcfgNegatives}{15}  %
\newcommand{\qStdcfgProbeFreq}{4}  %
\newcommand{\qStdcfgSeqLen}{128}  %
\newcommand{\qStdcfgSevenSixEightAuditCommit}{1ec18042}  %
\newcommand{\qStdcfgSevenSixEightAuditCsOne}{68fab537e7e8}  %
\newcommand{\qStdcfgSevenSixEightAuditCsTwo}{3208fe07a072}  %
\newcommand{\qStdcfgSevenSixEightAuditCsZero}{e73f1f8aa6d9}  %
\newcommand{\qStdcfgSevenSixEightAuditLanded}{2026-09-30 08:17}  %
\newcommand{\qStdcfgSevenSixEightAuditReopenTFirst}{b55e8721}  %
\newcommand{\qStdcfgSevenSixEightAuditSizeBandHi}{2}  %
\newcommand{\qStdcfgSevenSixEightAuditSizeBandLo}{1/2}  %
\newcommand{\qStdcfgSevenSixEightAuditSizeBaseRead}{847ba5b02112}  %
\newcommand{\qStdcfgSevenSixEightAuditSizeBridgeOne}{d0b9e372}  %
\newcommand{\qStdcfgSevenSixEightAuditSizeBridgeOneTime}{14:55:40}  %
\newcommand{\qStdcfgSevenSixEightAuditSizeBridgeTwo}{14011807}  %
\newcommand{\qStdcfgSevenSixEightAuditSizeBridgeTwoTime}{14:55:50}  %
\newcommand{\qStdcfgSevenSixEightAuditSizeDefaultLine}{1110}  %
\newcommand{\qStdcfgSevenSixEightAuditSizeFormatLines}{five}  %
\newcommand{\qStdcfgSevenSixEightAuditSizeFreeze}{1bbf751b}  %
\newcommand{\qStdcfgSevenSixEightAuditSizeFreezeTime}{14:51:16}  %
\newcommand{\qStdcfgSevenSixEightAuditSizeJudgeLine}{78}  %
\newcommand{\qStdcfgSevenSixEightAuditSizeManifestAtP}{1785453bd243}  %
\newcommand{\qStdcfgSevenSixEightAuditSizeManifestHead}{fabce658c145}  %
\newcommand{\qStdcfgSevenSixEightAuditSizeTAfter}{d0b9e372}  %
\newcommand{\qStdcfgSevenSixEightAuditSizeTAfterTime}{09-29 14:55:40}  %
\newcommand{\qStdcfgSevenSixEightAuditXEightTDriver}{c621c554}  %
\newcommand{\qStdcfgSevenSixEightAuditXEightTFirst}{4703728b}  %
\newcommand{\qStdcfgSevenSixEightDModel}{768}  %
\newcommand{\qStdcfgSevenSixEightDevice}{NVIDIA GeForce RTX 3090 Ti}  %
\newcommand{\qStdcfgSevenSixEightDoseFour}{four-decade}  %
\newcommand{\qStdcfgSevenSixEightDoseThree}{three-decade}  %
\newcommand{\qStdcfgSevenSixEightHeads}{12}  %
\newcommand{\qStdcfgSevenSixEightHostMain}{e36408fdee61}  %
\newcommand{\qStdcfgSevenSixEightHostSoftmax}{db7315d9726b}  %
\newcommand{\qStdcfgSevenSixEightLayers}{8}  %
\newcommand{\qStdcfgSevenSixEightLr}{SELECTED}  %
\newcommand{\qStdcfgSevenSixEightLrArgv}{0.0001}  %
\newcommand{\qStdcfgSevenSixEightLrBaseLabel}{512 selection, side by side}  %
\newcommand{\qStdcfgSevenSixEightLrBaseLr}{0.0003}  %
\newcommand{\qStdcfgSevenSixEightLrBaseOutcome}{SELECTED}  %
\newcommand{\qStdcfgSevenSixEightLrRungOne}{0.0001}  %
\newcommand{\qStdcfgSevenSixEightLrRungOneERead}{0.2816}  %
\newcommand{\qStdcfgSevenSixEightLrRungOneEligible}{3}  %
\newcommand{\qStdcfgSevenSixEightLrRungOneSeeds}{3}  %
\newcommand{\qStdcfgSevenSixEightLrRungThree}{0.0003}  %
\newcommand{\qStdcfgSevenSixEightLrRungThreeERead}{0.2852}  %
\newcommand{\qStdcfgSevenSixEightLrRungThreeEligible}{3}  %
\newcommand{\qStdcfgSevenSixEightLrRungThreeSeeds}{3}  %
\newcommand{\qStdcfgSevenSixEightLrRungTwo}{0.0002}  %
\newcommand{\qStdcfgSevenSixEightLrRungTwoERead}{0.2843}  %
\newcommand{\qStdcfgSevenSixEightLrRungTwoEligible}{3}  %
\newcommand{\qStdcfgSevenSixEightLrRungTwoSeeds}{3}  %
\newcommand{\qStdcfgSevenSixEightLrRuns}{nine}  %
\newcommand{\qStdcfgSevenSixEightLrSetMachineLost}{VOID-MACHINE-LOST}  %
\newcommand{\qStdcfgSevenSixEightLrSetNone}{LR-NONE-768}  %
\newcommand{\qStdcfgSevenSixEightLrSetNotCovered}{NOT-COVERED}  %
\newcommand{\qStdcfgSevenSixEightLrSetSelected}{SELECTED}  %
\newcommand{\qStdcfgSevenSixEightMachineMain}{V6}  %
\newcommand{\qStdcfgSevenSixEightMachineSoftmax}{V3}  %
\newcommand{\qStdcfgSevenSixEightMissingCatCorrected}{4}  %
\newcommand{\qStdcfgSevenSixEightMissingCatFirst}{6}  %
\newcommand{\qStdcfgSevenSixEightReopenCert}{cert MISSING}  %
\newcommand{\qStdcfgSevenSixEightReopenHost}{e36408fdee61}  %
\newcommand{\qStdcfgSevenSixEightReopenLoGateSOne}{OK}  %
\newcommand{\qStdcfgSevenSixEightReopenLoGateSTwo}{OK}  %
\newcommand{\qStdcfgSevenSixEightReopenMachine}{V6}  %
\newcommand{\qStdcfgSevenSixEightReopenOutcome}{NOT-COVERED}  %
\newcommand{\qStdcfgSevenSixEightReopenToken}{20260929T082411Z}  %
\newcommand{\qStdcfgSevenSixEightReopenWordFourSOne}{FIRES}  %
\newcommand{\qStdcfgSevenSixEightReopenWordFourSTwo}{FIRES}  %
\newcommand{\qStdcfgSevenSixEightReopenWordFourSZero}{MISSING}  %
\newcommand{\qStdcfgSevenSixEightReopenWordThreeSOne}{NOT-FIRING}  %
\newcommand{\qStdcfgSevenSixEightReopenWordThreeSTwo}{NOT-FIRING}  %
\newcommand{\qStdcfgSevenSixEightReopenWordThreeSZero}{MISSING}  %
\newcommand{\qStdcfgSevenSixEightRestartTwoToken}{20260930T122042Z}  %
\newcommand{\qStdcfgSevenSixEightSize}{SAME-ORDER}  %
\newcommand{\qStdcfgSevenSixEightSizeBandHi}{2}  %
\newcommand{\qStdcfgSevenSixEightSizeBandLo}{1/2}  %
\newcommand{\qStdcfgSevenSixEightSizeBlockRatioBase}{11.3578}  %
\newcommand{\qStdcfgSevenSixEightSizeCeil}{4.0397}  %
\newcommand{\qStdcfgSevenSixEightSizeCellR}{0.7368}  %
\newcommand{\qStdcfgSevenSixEightSizeCellRatioBase}{11.3578}  %
\newcommand{\qStdcfgSevenSixEightSizeCellRatioWide}{8.3688}  %
\newcommand{\qStdcfgSevenSixEightSizeExpCeil}{56.8074}  %
\newcommand{\qStdcfgSevenSixEightSizeFlag}{reachable}  %
\newcommand{\qStdcfgSevenSixEightSizeLabel}{bce15, judged}  %
\newcommand{\qStdcfgSevenSixEightSizeLrBase}{0.0003}  %
\newcommand{\qStdcfgSevenSixEightSizeLrWide}{0.0001}  %
\newcommand{\qStdcfgSevenSixEightSizeMachineBase}{V6}  %
\newcommand{\qStdcfgSevenSixEightSizeMachineWide}{V6}  %
\newcommand{\qStdcfgSevenSixEightSizeR}{0.74}  %
\newcommand{\qStdcfgSevenSixEightSizeRUpper}{5.0016}  %
\newcommand{\qStdcfgSevenSixEightSizeRUpperShort}{5.00}  %
\newcommand{\qStdcfgSevenSixEightSizeSbsHostBaseSOne}{3cd66d135dc5}  %
\newcommand{\qStdcfgSevenSixEightSizeSbsHostBaseSTwo}{3cd66d135dc5}  %
\newcommand{\qStdcfgSevenSixEightSizeSbsHostBaseSZero}{3cd66d135dc5}  %
\newcommand{\qStdcfgSevenSixEightSizeSbsLabel}{side by side, not judged (machine differs)}  %
\newcommand{\qStdcfgSevenSixEightSizeSbsMachineBaseSOne}{V2}  %
\newcommand{\qStdcfgSevenSixEightSizeSbsMachineBaseSTwo}{V2}  %
\newcommand{\qStdcfgSevenSixEightSizeSbsMachineBaseSZero}{V2}  %
\newcommand{\qStdcfgSevenSixEightSizeSbsMachineWideSOne}{V6}  %
\newcommand{\qStdcfgSevenSixEightSizeSbsMachineWideSTwo}{V6}  %
\newcommand{\qStdcfgSevenSixEightSizeSbsMachineWideSZero}{V6}  %
\newcommand{\qStdcfgSevenSixEightSizeSbsRatioBaseSOne}{8.8996}  %
\newcommand{\qStdcfgSevenSixEightSizeSbsRatioBaseSTwo}{8.8982}  %
\newcommand{\qStdcfgSevenSixEightSizeSbsRatioBaseSZero}{11.3578}  %
\newcommand{\qStdcfgSevenSixEightSizeSbsStateBaseSOne}{FIRES}  %
\newcommand{\qStdcfgSevenSixEightSizeSbsStateBaseSTwo}{FIRES}  %
\newcommand{\qStdcfgSevenSixEightSizeSbsStateBaseSZero}{FIRES}  %
\newcommand{\qStdcfgSevenSixEightSizeSbsStateWideSOne}{MISSING}  %
\newcommand{\qStdcfgSevenSixEightSizeSbsStateWideSTwo}{MISSING}  %
\newcommand{\qStdcfgSevenSixEightSizeSbsStateWideSZero}{FIRES}  %
\newcommand{\qStdcfgSevenSixEightSizeSetChanges}{CHANGES-WITH-SIZE}  %
\newcommand{\qStdcfgSevenSixEightSizeSetNotCovered}{NOT-COVERED}  %
\newcommand{\qStdcfgSevenSixEightSizeSetNotRun}{NOT-RUN}  %
\newcommand{\qStdcfgSevenSixEightSizeSetSame}{SAME-ORDER}  %
\newcommand{\qStdcfgSevenSixEightSizeSmCeilSOne}{3.8354}  %
\newcommand{\qStdcfgSevenSixEightSizeSmCeilSTwo}{3.6299}  %
\newcommand{\qStdcfgSevenSixEightSizeSmCeilSZero}{3.7516}  %
\newcommand{\qStdcfgSevenSixEightSizeSmCellRSOne}{1.3195}  %
\newcommand{\qStdcfgSevenSixEightSizeSmCellRSTwo}{1.0390}  %
\newcommand{\qStdcfgSevenSixEightSizeSmCellRSZero}{0.7742}  %
\newcommand{\qStdcfgSevenSixEightSizeSmCellRatioBaseSOne}{4.9799}  %
\newcommand{\qStdcfgSevenSixEightSizeSmCellRatioBaseSTwo}{5.2718}  %
\newcommand{\qStdcfgSevenSixEightSizeSmCellRatioBaseSZero}{7.7063}  %
\newcommand{\qStdcfgSevenSixEightSizeSmCellRatioWideSOne}{6.5708}  %
\newcommand{\qStdcfgSevenSixEightSizeSmCellRatioWideSTwo}{5.4771}  %
\newcommand{\qStdcfgSevenSixEightSizeSmCellRatioWideSZero}{5.9666}  %
\newcommand{\qStdcfgSevenSixEightSizeSmExpCeilSOne}{46.3109}  %
\newcommand{\qStdcfgSevenSixEightSizeSmExpCeilSTwo}{37.7073}  %
\newcommand{\qStdcfgSevenSixEightSizeSmExpCeilSZero}{42.5908}  %
\newcommand{\qStdcfgSevenSixEightSizeSmFlagSOne}{reachable}  %
\newcommand{\qStdcfgSevenSixEightSizeSmFlagSTwo}{reachable}  %
\newcommand{\qStdcfgSevenSixEightSizeSmFlagSZero}{reachable}  %
\newcommand{\qStdcfgSevenSixEightSizeSmLabel}{softmax, second instrument}  %
\newcommand{\qStdcfgSevenSixEightSizeSmLrBaseSOne}{0.0003}  %
\newcommand{\qStdcfgSevenSixEightSizeSmLrBaseSTwo}{0.0003}  %
\newcommand{\qStdcfgSevenSixEightSizeSmLrBaseSZero}{0.0003}  %
\newcommand{\qStdcfgSevenSixEightSizeSmLrWideSOne}{0.0001}  %
\newcommand{\qStdcfgSevenSixEightSizeSmLrWideSTwo}{0.0001}  %
\newcommand{\qStdcfgSevenSixEightSizeSmLrWideSZero}{0.0001}  %
\newcommand{\qStdcfgSevenSixEightSizeSmMachineBaseSOne}{V3}  %
\newcommand{\qStdcfgSevenSixEightSizeSmMachineBaseSTwo}{V3}  %
\newcommand{\qStdcfgSevenSixEightSizeSmMachineBaseSZero}{V3}  %
\newcommand{\qStdcfgSevenSixEightSizeSmMachineWideSOne}{V3}  %
\newcommand{\qStdcfgSevenSixEightSizeSmMachineWideSTwo}{V3}  %
\newcommand{\qStdcfgSevenSixEightSizeSmMachineWideSZero}{V3}  %
\newcommand{\qStdcfgSevenSixEightSizeSmRSOne}{1.32}  %
\newcommand{\qStdcfgSevenSixEightSizeSmRSTwo}{1.04}  %
\newcommand{\qStdcfgSevenSixEightSizeSmRSZero}{0.77}  %
\newcommand{\qStdcfgSevenSixEightSizeSmRUpperSOne}{9.2995}  %
\newcommand{\qStdcfgSevenSixEightSizeSmRUpperSTwo}{7.1527}  %
\newcommand{\qStdcfgSevenSixEightSizeSmRUpperSZero}{5.5268}  %
\newcommand{\qStdcfgSevenSixEightSizeSmSetChanges}{CHANGES-WITH-SIZE}  %
\newcommand{\qStdcfgSevenSixEightSizeSmSetNotCovered}{NOT-COVERED}  %
\newcommand{\qStdcfgSevenSixEightSizeSmSetNotRun}{NOT-RUN}  %
\newcommand{\qStdcfgSevenSixEightSizeSmSetSame}{SAME-ORDER}  %
\newcommand{\qStdcfgSevenSixEightSizeSmStateBaseSOne}{FIRES}  %
\newcommand{\qStdcfgSevenSixEightSizeSmStateBaseSTwo}{FIRES}  %
\newcommand{\qStdcfgSevenSixEightSizeSmStateBaseSZero}{FIRES}  %
\newcommand{\qStdcfgSevenSixEightSizeSmStateWideSOne}{FIRES}  %
\newcommand{\qStdcfgSevenSixEightSizeSmStateWideSTwo}{FIRES}  %
\newcommand{\qStdcfgSevenSixEightSizeSmStateWideSZero}{FIRES}  %
\newcommand{\qStdcfgSevenSixEightSizeSmThresholdSOne}{1.3863}  %
\newcommand{\qStdcfgSevenSixEightSizeSmThresholdSTwo}{1.3863}  %
\newcommand{\qStdcfgSevenSixEightSizeSmThresholdSZero}{1.3863}  %
\newcommand{\qStdcfgSevenSixEightSizeSoftmaxOne}{SAME-ORDER}  %
\newcommand{\qStdcfgSevenSixEightSizeSoftmaxTwo}{SAME-ORDER}  %
\newcommand{\qStdcfgSevenSixEightSizeSoftmaxZero}{SAME-ORDER}  %
\newcommand{\qStdcfgSevenSixEightSizeStateBase}{FIRES}  %
\newcommand{\qStdcfgSevenSixEightSizeStateWide}{FIRES}  %
\newcommand{\qStdcfgSevenSixEightSizeThreshold}{1.7368}  %
\newcommand{\qStdcfgSevenSixEightXEightArmAllEightSTwo}{MISSING}  %
\newcommand{\qStdcfgSevenSixEightXEightBridgeSOne}{IDENTICAL}  %
\newcommand{\qStdcfgSevenSixEightXEightBridgeSTwo}{OTHER}  %
\newcommand{\qStdcfgSevenSixEightXEightBridgeSZero}{IDENTICAL}  %
\newcommand{\qStdcfgSevenSixEightXEightBridgeStateSTwo}{MISSING}  %
\newcommand{\qStdcfgSevenSixEightXEightCellRRestSOne}{0.6557}  %
\newcommand{\qStdcfgSevenSixEightXEightCellRRestSZero}{0.6824}  %
\newcommand{\qStdcfgSevenSixEightXEightCellRRowSOne}{0.8522}  %
\newcommand{\qStdcfgSevenSixEightXEightCellRRowSZero}{0.8352}  %
\newcommand{\qStdcfgSevenSixEightXEightClosed}{2}  %
\newcommand{\qStdcfgSevenSixEightXEightDEightSOne}{1.76250}  %
\newcommand{\qStdcfgSevenSixEightXEightDEightSZero}{1.83049}  %
\newcommand{\qStdcfgSevenSixEightXEightDFourSOne}{0.29486}  %
\newcommand{\qStdcfgSevenSixEightXEightDFourSZero}{0.26967}  %
\newcommand{\qStdcfgSevenSixEightXEightDRestSOne}{0.80010}  %
\newcommand{\qStdcfgSevenSixEightXEightDRestSZero}{0.76544}  %
\newcommand{\qStdcfgSevenSixEightXEightDRowSOne}{0.51180}  %
\newcommand{\qStdcfgSevenSixEightXEightDRowSZero}{0.52696}  %
\newcommand{\qStdcfgSevenSixEightXEightDenomSOne}{1.46764}  %
\newcommand{\qStdcfgSevenSixEightXEightDenomSZero}{1.56082}  %
\newcommand{\qStdcfgSevenSixEightXEightEntries}{15}  %
\newcommand{\qStdcfgSevenSixEightXEightLanded}{13}  %
\newcommand{\qStdcfgSevenSixEightXEightRRestSOne}{0.66}  %
\newcommand{\qStdcfgSevenSixEightXEightRRestSZero}{0.68}  %
\newcommand{\qStdcfgSevenSixEightXEightRRowSOne}{0.85}  %
\newcommand{\qStdcfgSevenSixEightXEightRRowSZero}{0.84}  %
\newcommand{\qStdcfgSevenSixEightXEightSOneRRest}{C-MIXED}  %
\newcommand{\qStdcfgSevenSixEightXEightSOneRRow}{A-DENOMINATOR}  %
\newcommand{\qStdcfgSevenSixEightXEightSTwoRRest}{NOT-COVERED}  %
\newcommand{\qStdcfgSevenSixEightXEightSTwoRRow}{NOT-COVERED}  %
\newcommand{\qStdcfgSevenSixEightXEightSZeroRRest}{C-MIXED}  %
\newcommand{\qStdcfgSevenSixEightXEightSZeroRRow}{A-DENOMINATOR}  %
\newcommand{\qStdcfgSevenSixEightXEightSeedOne}{1}  %
\newcommand{\qStdcfgSevenSixEightXEightSeedTwo}{2}  %
\newcommand{\qStdcfgSevenSixEightXEightSeedZero}{0}  %
\newcommand{\qStdcfgSevenSixEightXEightSetA}{A-DENOMINATOR}  %
\newcommand{\qStdcfgSevenSixEightXEightSetB}{B-TRAJECTORY}  %
\newcommand{\qStdcfgSevenSixEightXEightSetC}{C-MIXED}  %
\newcommand{\qStdcfgSevenSixEightXEightSetMissing}{MISSING}  %
\newcommand{\qStdcfgSevenSixEightXEightSetNotCovered}{NOT-COVERED}  %
\newcommand{\qStdcfgSevenSixEightXEightSetNotRun}{NOT-RUN}  %
\newcommand{\qStdcfgSevenSixEightXEightSetOut}{OUT-OF-RANGE}  %
\newcommand{\qStdcfgSevenSixEightXEightSetUndefined}{UNDEFINED}  %
\newcommand{\qStdcfgSevenSixEightXEightSetVoid}{VOID}  %
\newcommand{\qStdcfgSevenSixEightXEightSumSOne}{1.5079}  %
\newcommand{\qStdcfgSevenSixEightXEightSumSZero}{1.5175}  %
\newcommand{\qStdcfgSevenSixEightXNineCellCsSOne}{0.01840}  %
\newcommand{\qStdcfgSevenSixEightXNineCellCsSZero}{0.01760}  %
\newcommand{\qStdcfgSevenSixEightXNineCellRhoSOne}{0.17758}  %
\newcommand{\qStdcfgSevenSixEightXNineCellRhoSZero}{0.18128}  %
\newcommand{\qStdcfgSevenSixEightXNineCellShareRowSOne}{0.3095}  %
\newcommand{\qStdcfgSevenSixEightXNineCellShareRowSZero}{0.3082}  %
\newcommand{\qStdcfgSevenSixEightXNineCellSlrSOne}{0.4326}  %
\newcommand{\qStdcfgSevenSixEightXNineCellSlrSZero}{0.4227}  %
\newcommand{\qStdcfgSevenSixEightXNineOne}{LR-SAME-SIDE}  %
\newcommand{\qStdcfgSevenSixEightXNineSetMissing}{MISSING}  %
\newcommand{\qStdcfgSevenSixEightXNineSetNeutral}{LR-NEUTRAL}  %
\newcommand{\qStdcfgSevenSixEightXNineSetNotCovered}{NOT-COVERED}  %
\newcommand{\qStdcfgSevenSixEightXNineSetNotRun}{NOT-RUN}  %
\newcommand{\qStdcfgSevenSixEightXNineSetOpposite}{LR-OPPOSITE}  %
\newcommand{\qStdcfgSevenSixEightXNineSetOut}{OUT-OF-RANGE}  %
\newcommand{\qStdcfgSevenSixEightXNineSetSameSide}{LR-SAME-SIDE}  %
\newcommand{\qStdcfgSevenSixEightXNineSetVoid}{VOID}  %
\newcommand{\qStdcfgSevenSixEightXNineSlrSOne}{0.43}  %
\newcommand{\qStdcfgSevenSixEightXNineSlrSZero}{0.42}  %
\newcommand{\qStdcfgSevenSixEightXNineTwo}{NOT-COVERED}  %
\newcommand{\qStdcfgSevenSixEightXNineZero}{LR-SAME-SIDE}  %
\newcommand{\qStdcfgSevenSixEightXOne}{NOT-COVERED}  %
\newcommand{\qStdcfgSevenSixEightXOneCellCeilFour}{4.0397}  %
\newcommand{\qStdcfgSevenSixEightXOneCellCeilThree}{2.2828}  %
\newcommand{\qStdcfgSevenSixEightXOneCellEscHiFour}{10,000}  %
\newcommand{\qStdcfgSevenSixEightXOneCellEscHiThree}{8,000}  %
\newcommand{\qStdcfgSevenSixEightXOneCellEscLoFour}{8,000}  %
\newcommand{\qStdcfgSevenSixEightXOneCellEscLoThree}{8,000}  %
\newcommand{\qStdcfgSevenSixEightXOneCellExpFFour}{8.3688}  %
\newcommand{\qStdcfgSevenSixEightXOneCellExpFThree}{3.3249}  %
\newcommand{\qStdcfgSevenSixEightXOneCellFFour}{2.1245}  %
\newcommand{\qStdcfgSevenSixEightXOneCellFThree}{1.2014}  %
\newcommand{\qStdcfgSevenSixEightXOneCellHeadroomFour}{2.6534}  %
\newcommand{\qStdcfgSevenSixEightXOneCellHeadroomThree}{0.8965}  %
\newcommand{\qStdcfgSevenSixEightXOneCellKillFour}{0.1195}  %
\newcommand{\qStdcfgSevenSixEightXOneCellKillThree}{0.3008}  %
\newcommand{\qStdcfgSevenSixEightXOneCellShareFour}{0.4741}  %
\newcommand{\qStdcfgSevenSixEightXOneCellShareThree}{0.4737}  %
\newcommand{\qStdcfgSevenSixEightXOneCellSideFour}{low}  %
\newcommand{\qStdcfgSevenSixEightXOneCert}{cert MISSING}  %
\newcommand{\qStdcfgSevenSixEightXOneClosed}{7}  %
\newcommand{\qStdcfgSevenSixEightXOneEntries}{10}  %
\newcommand{\qStdcfgSevenSixEightXOneEscHi}{10,000}  %
\newcommand{\qStdcfgSevenSixEightXOneEscLo}{8,000}  %
\newcommand{\qStdcfgSevenSixEightXOneKillFour}{0.12}  %
\newcommand{\qStdcfgSevenSixEightXOneKillThree}{0.30}  %
\newcommand{\qStdcfgSevenSixEightXOneLanded}{3}  %
\newcommand{\qStdcfgSevenSixEightXOneMarginThree}{0.05}  %
\newcommand{\qStdcfgSevenSixEightXOneRatioFour}{8.4}  %
\newcommand{\qStdcfgSevenSixEightXOneSeedOne}{1}  %
\newcommand{\qStdcfgSevenSixEightXOneSeedTwo}{2}  %
\newcommand{\qStdcfgSevenSixEightXOneSeedZero}{0}  %
\newcommand{\qStdcfgSevenSixEightXOneSetGo}{GO}  %
\newcommand{\qStdcfgSevenSixEightXOneSetNoGo}{NO-GO}  %
\newcommand{\qStdcfgSevenSixEightXOneSetNotCovered}{NOT-COVERED}  %
\newcommand{\qStdcfgSevenSixEightXOneSetNotRun}{NOT-RUN}  %
\newcommand{\qStdcfgSevenSixEightXOneSoftmax}{S-GO}  %
\newcommand{\qStdcfgSevenSixEightXOneSoftmaxCellCeilSOne}{3.8354}  %
\newcommand{\qStdcfgSevenSixEightXOneSoftmaxCellCeilSTwo}{3.6299}  %
\newcommand{\qStdcfgSevenSixEightXOneSoftmaxCellCeilSZero}{3.7516}  %
\newcommand{\qStdcfgSevenSixEightXOneSoftmaxCellEscHiSOne}{10,000}  %
\newcommand{\qStdcfgSevenSixEightXOneSoftmaxCellEscHiSTwo}{10,000}  %
\newcommand{\qStdcfgSevenSixEightXOneSoftmaxCellEscHiSZero}{10,000}  %
\newcommand{\qStdcfgSevenSixEightXOneSoftmaxCellEscLoSOne}{9,000}  %
\newcommand{\qStdcfgSevenSixEightXOneSoftmaxCellEscLoSTwo}{10,000}  %
\newcommand{\qStdcfgSevenSixEightXOneSoftmaxCellEscLoSZero}{9,000}  %
\newcommand{\qStdcfgSevenSixEightXOneSoftmaxCellExpFSOne}{6.5708}  %
\newcommand{\qStdcfgSevenSixEightXOneSoftmaxCellExpFSTwo}{5.4771}  %
\newcommand{\qStdcfgSevenSixEightXOneSoftmaxCellExpFSZero}{5.9666}  %
\newcommand{\qStdcfgSevenSixEightXOneSoftmaxCellFSOne}{1.8826}  %
\newcommand{\qStdcfgSevenSixEightXOneSoftmaxCellFSTwo}{1.7006}  %
\newcommand{\qStdcfgSevenSixEightXOneSoftmaxCellFSZero}{1.7862}  %
\newcommand{\qStdcfgSevenSixEightXOneSoftmaxCellHeadroomSOne}{2.4491}  %
\newcommand{\qStdcfgSevenSixEightXOneSoftmaxCellHeadroomSTwo}{2.2436}  %
\newcommand{\qStdcfgSevenSixEightXOneSoftmaxCellHeadroomSZero}{2.3653}  %
\newcommand{\qStdcfgSevenSixEightXOneSoftmaxCellKillSOne}{0.1522}  %
\newcommand{\qStdcfgSevenSixEightXOneSoftmaxCellKillSTwo}{0.1826}  %
\newcommand{\qStdcfgSevenSixEightXOneSoftmaxCellKillSZero}{0.1676}  %
\newcommand{\qStdcfgSevenSixEightXOneSoftmaxCellShareSOne}{0.5091}  %
\newcommand{\qStdcfgSevenSixEightXOneSoftmaxCellShareSTwo}{0.5315}  %
\newcommand{\qStdcfgSevenSixEightXOneSoftmaxCellShareSZero}{0.5239}  %
\newcommand{\qStdcfgSevenSixEightXOneSoftmaxCellSideSOne}{low}  %
\newcommand{\qStdcfgSevenSixEightXOneSoftmaxCellSideSTwo}{low}  %
\newcommand{\qStdcfgSevenSixEightXOneSoftmaxCellSideSZero}{low}  %
\newcommand{\qStdcfgSevenSixEightXOneSoftmaxEscHi}{10,000}  %
\newcommand{\qStdcfgSevenSixEightXOneSoftmaxEscLo}{9,000}  %
\newcommand{\qStdcfgSevenSixEightXOneSoftmaxRatioSOne}{6.6}  %
\newcommand{\qStdcfgSevenSixEightXOneSoftmaxRatioSTwo}{5.5}  %
\newcommand{\qStdcfgSevenSixEightXOneSoftmaxRatioSZero}{6.0}  %
\newcommand{\qStdcfgSevenSixEightXOneSoftmaxSeedOne}{1}  %
\newcommand{\qStdcfgSevenSixEightXOneSoftmaxSeedTwo}{2}  %
\newcommand{\qStdcfgSevenSixEightXOneSoftmaxSeedZero}{0}  %
\newcommand{\qStdcfgSevenSixEightXOneSoftmaxSetGo}{S-GO}  %
\newcommand{\qStdcfgSevenSixEightXOneSoftmaxSetNoGo}{S-NO-GO}  %
\newcommand{\qStdcfgSevenSixEightXOneSoftmaxSetNotCovered}{NOT-COVERED}  %
\newcommand{\qStdcfgSevenSixEightXOneSoftmaxSetNotRun}{NOT-RUN}  %
\newcommand{\qStdcfgSevenSixEightXOneWordFourSOne}{MISSING}  %
\newcommand{\qStdcfgSevenSixEightXOneWordFourSTwo}{MISSING}  %
\newcommand{\qStdcfgSevenSixEightXOneWordFourSZero}{FIRES}  %
\newcommand{\qStdcfgSevenSixEightXOneWordThreeSOne}{MISSING}  %
\newcommand{\qStdcfgSevenSixEightXOneWordThreeSTwo}{MISSING}  %
\newcommand{\qStdcfgSevenSixEightXOneWordThreeSZero}{NOT-FIRING}  %
\newcommand{\qStdcfgSteps}{60,000}  %
\newcommand{\qStdcfgVocab}{32,000}  %
\newcommand{\qStdcfgWarmup}{2,000}  %
\newcommand{\qStdcfgXEightAGe}{0.75}  %
\newcommand{\qStdcfgXEightBLe}{0.25}  %
\newcommand{\qStdcfgXEightBridgeComparedSeedOne}{214,359}  %
\newcommand{\qStdcfgXEightBridgeComparedSeedTwo}{214,359}  %
\newcommand{\qStdcfgXEightBridgeComparedSeedZero}{214,359}  %
\newcommand{\qStdcfgXEightBridgeDiffersSeedOne}{0}  %
\newcommand{\qStdcfgXEightBridgeDiffersSeedTwo}{0}  %
\newcommand{\qStdcfgXEightBridgeDiffersSeedZero}{0}  %
\newcommand{\qStdcfgXEightBridgeSeedOne}{IDENTICAL}  %
\newcommand{\qStdcfgXEightBridgeSeedTwo}{IDENTICAL}  %
\newcommand{\qStdcfgXEightBridgeSeedZero}{IDENTICAL}  %
\newcommand{\qStdcfgXEightDenomSeedOne}{2.3654}  %
\newcommand{\qStdcfgXEightDenomSeedTwo}{2.2836}  %
\newcommand{\qStdcfgXEightDenomSeedZero}{2.5947}  %
\newcommand{\qStdcfgXEightExtBridgeComparedSeedFour}{214,359}  %
\newcommand{\qStdcfgXEightExtBridgeComparedSeedThree}{214,359}  %
\newcommand{\qStdcfgXEightExtBridgeDiffersSeedFour}{0}  %
\newcommand{\qStdcfgXEightExtBridgeDiffersSeedThree}{0}  %
\newcommand{\qStdcfgXEightExtBridgeSeedFour}{IDENTICAL}  %
\newcommand{\qStdcfgXEightExtBridgeSeedThree}{IDENTICAL}  %
\newcommand{\qStdcfgXEightExtDenomSeedFour}{2.3288}  %
\newcommand{\qStdcfgXEightExtDenomSeedThree}{2.1510}  %
\newcommand{\qStdcfgXEightExtRestRPrintedSeedFour}{0.6934}  %
\newcommand{\qStdcfgXEightExtRestRPrintedSeedThree}{0.6728}  %
\newcommand{\qStdcfgXEightExtRestRSeedFour}{0.69}  %
\newcommand{\qStdcfgXEightExtRestRSeedThree}{0.67}  %
\newcommand{\qStdcfgXEightExtRestSeedFour}{C-MIXED}  %
\newcommand{\qStdcfgXEightExtRestSeedThree}{C-MIXED}  %
\newcommand{\qStdcfgXEightExtRowRPrintedSeedFour}{0.8483}  %
\newcommand{\qStdcfgXEightExtRowRPrintedSeedThree}{0.8386}  %
\newcommand{\qStdcfgXEightExtRowRSeedFour}{0.85}  %
\newcommand{\qStdcfgXEightExtRowRSeedThree}{0.84}  %
\newcommand{\qStdcfgXEightExtRowSeedFour}{A-DENOMINATOR}  %
\newcommand{\qStdcfgXEightExtRowSeedThree}{A-DENOMINATOR}  %
\newcommand{\qStdcfgXEightExtSumSeedFour}{1.5416}  %
\newcommand{\qStdcfgXEightExtSumSeedThree}{1.5114}  %
\newcommand{\qStdcfgXEightOutAbove}{1}  %
\newcommand{\qStdcfgXEightOutBelow}{0}  %
\newcommand{\qStdcfgXEightRestRPrintedSeedOne}{0.7037}  %
\newcommand{\qStdcfgXEightRestRPrintedSeedTwo}{0.6938}  %
\newcommand{\qStdcfgXEightRestRPrintedSeedZero}{0.7249}  %
\newcommand{\qStdcfgXEightRestRSeedOne}{0.70}  %
\newcommand{\qStdcfgXEightRestRSeedTwo}{0.69}  %
\newcommand{\qStdcfgXEightRestRSeedZero}{0.72}  %
\newcommand{\qStdcfgXEightRestSeedOne}{C-MIXED}  %
\newcommand{\qStdcfgXEightRestSeedTwo}{C-MIXED}  %
\newcommand{\qStdcfgXEightRestSeedZero}{C-MIXED}  %
\newcommand{\qStdcfgXEightRowRPrintedSeedOne}{0.8391}  %
\newcommand{\qStdcfgXEightRowRPrintedSeedTwo}{0.8533}  %
\newcommand{\qStdcfgXEightRowRPrintedSeedZero}{0.8479}  %
\newcommand{\qStdcfgXEightRowRSeedOne}{0.84}  %
\newcommand{\qStdcfgXEightRowRSeedTwo}{0.85}  %
\newcommand{\qStdcfgXEightRowRSeedZero}{0.85}  %
\newcommand{\qStdcfgXEightRowSeedOne}{A-DENOMINATOR}  %
\newcommand{\qStdcfgXEightRowSeedTwo}{A-DENOMINATOR}  %
\newcommand{\qStdcfgXEightRowSeedZero}{A-DENOMINATOR}  %
\newcommand{\qStdcfgXEightSetADenominator}{A-DENOMINATOR}  %
\newcommand{\qStdcfgXEightSetBTrajectory}{B-TRAJECTORY}  %
\newcommand{\qStdcfgXEightSetCMixed}{C-MIXED}  %
\newcommand{\qStdcfgXEightSetOutOfRange}{OUT-OF-RANGE}  %
\newcommand{\qStdcfgXEightSetUndefined}{UNDEFINED}  %
\newcommand{\qStdcfgXEightSoftmaxDenomSeedOne}{2.7823}  %
\newcommand{\qStdcfgXEightSoftmaxDenomSeedTwo}{2.7102}  %
\newcommand{\qStdcfgXEightSoftmaxDenomSeedZero}{2.7932}  %
\newcommand{\qStdcfgXEightSoftmaxRestRPrintedSeedOne}{0.5527}  %
\newcommand{\qStdcfgXEightSoftmaxRestRPrintedSeedTwo}{0.5559}  %
\newcommand{\qStdcfgXEightSoftmaxRestRPrintedSeedZero}{0.5083}  %
\newcommand{\qStdcfgXEightSoftmaxRestRSeedOne}{0.55}  %
\newcommand{\qStdcfgXEightSoftmaxRestRSeedTwo}{0.56}  %
\newcommand{\qStdcfgXEightSoftmaxRestRSeedZero}{0.51}  %
\newcommand{\qStdcfgXEightSoftmaxRestSeedOne}{C-MIXED}  %
\newcommand{\qStdcfgXEightSoftmaxRestSeedTwo}{C-MIXED}  %
\newcommand{\qStdcfgXEightSoftmaxRestSeedZero}{C-MIXED}  %
\newcommand{\qStdcfgXEightSoftmaxRowRPrintedSeedOne}{0.8701}  %
\newcommand{\qStdcfgXEightSoftmaxRowRPrintedSeedTwo}{0.8789}  %
\newcommand{\qStdcfgXEightSoftmaxRowRPrintedSeedZero}{0.8733}  %
\newcommand{\qStdcfgXEightSoftmaxRowRSeedOne}{0.87}  %
\newcommand{\qStdcfgXEightSoftmaxRowRSeedTwo}{0.88}  %
\newcommand{\qStdcfgXEightSoftmaxRowRSeedZero}{0.87}  %
\newcommand{\qStdcfgXEightSoftmaxRowSeedOne}{A-DENOMINATOR}  %
\newcommand{\qStdcfgXEightSoftmaxRowSeedTwo}{A-DENOMINATOR}  %
\newcommand{\qStdcfgXEightSoftmaxRowSeedZero}{A-DENOMINATOR}  %
\newcommand{\qStdcfgXEightSoftmaxSumSeedOne}{1.4228}  %
\newcommand{\qStdcfgXEightSoftmaxSumSeedTwo}{1.4348}  %
\newcommand{\qStdcfgXEightSoftmaxSumSeedZero}{1.3817}  %
\newcommand{\qStdcfgXEightSumSeedOne}{1.5428}  %
\newcommand{\qStdcfgXEightSumSeedTwo}{1.5471}  %
\newcommand{\qStdcfgXEightSumSeedZero}{1.5728}  %
\newcommand{\qStdcfgXNineCsSeedOne}{0.01812}  %
\newcommand{\qStdcfgXNineCsSeedTwo}{0.01854}  %
\newcommand{\qStdcfgXNineCsSeedZero}{0.01141}  %
\newcommand{\qStdcfgXNineExtCsSeedFour}{0.01661}  %
\newcommand{\qStdcfgXNineExtCsSeedThree}{0.02056}  %
\newcommand{\qStdcfgXNineExtRhoMeasSeedFour}{0.1955}  %
\newcommand{\qStdcfgXNineExtRhoMeasSeedThree}{0.2387}  %
\newcommand{\qStdcfgXNineExtSeedFour}{LR-SAME-SIDE}  %
\newcommand{\qStdcfgXNineExtSeedThree}{LR-SAME-SIDE}  %
\newcommand{\qStdcfgXNineExtShareRowSeedFour}{0.3059}  %
\newcommand{\qStdcfgXNineExtShareRowSeedThree}{0.2991}  %
\newcommand{\qStdcfgXNineExtSlrPrintedSeedFour}{0.3983}  %
\newcommand{\qStdcfgXNineExtSlrPrintedSeedThree}{0.3688}  %
\newcommand{\qStdcfgXNineExtSlrSeedFour}{0.40}  %
\newcommand{\qStdcfgXNineExtSlrSeedThree}{0.37}  %
\newcommand{\qStdcfgXNineNeutralHi}{1.1}  %
\newcommand{\qStdcfgXNineNeutralLo}{0.9}  %
\newcommand{\qStdcfgXNineRhoMeasSeedOne}{0.2103}  %
\newcommand{\qStdcfgXNineRhoMeasSeedTwo}{0.2104}  %
\newcommand{\qStdcfgXNineRhoMeasSeedZero}{0.1345}  %
\newcommand{\qStdcfgXNineSeedOne}{LR-SAME-SIDE}  %
\newcommand{\qStdcfgXNineSeedTwo}{LR-SAME-SIDE}  %
\newcommand{\qStdcfgXNineSeedZero}{LR-SAME-SIDE}  %
\newcommand{\qStdcfgXNineSetNeutral}{LR-NEUTRAL}  %
\newcommand{\qStdcfgXNineSetOpposite}{LR-OPPOSITE}  %
\newcommand{\qStdcfgXNineSetOutOfRange}{OUT-OF-RANGE}  %
\newcommand{\qStdcfgXNineSetSameSide}{LR-SAME-SIDE}  %
\newcommand{\qStdcfgXNineShareRowSeedOne}{0.3033}  %
\newcommand{\qStdcfgXNineShareRowSeedTwo}{0.3126}  %
\newcommand{\qStdcfgXNineShareRowSeedZero}{0.2994}  %
\newcommand{\qStdcfgXNineSlrPrintedSeedOne}{0.3888}  %
\newcommand{\qStdcfgXNineSlrPrintedSeedTwo}{0.3908}  %
\newcommand{\qStdcfgXNineSlrPrintedSeedZero}{0.4484}  %
\newcommand{\qStdcfgXNineSlrSeedOne}{0.39}  %
\newcommand{\qStdcfgXNineSlrSeedTwo}{0.39}  %
\newcommand{\qStdcfgXNineSlrSeedZero}{0.45}  %
\newcommand{\qStdcfgXOne}{GO}  %
\newcommand{\qStdcfgXOneBceFourN}{three}  %
\newcommand{\qStdcfgXOneBceFourOf}{three}  %
\newcommand{\qStdcfgXOneBceFourRatioHi}{11.4}  %
\newcommand{\qStdcfgXOneBceFourRatioLo}{8.9}  %
\newcommand{\qStdcfgXOneBceThreeInN}{three}  %
\newcommand{\qStdcfgXOneBceThreeInOf}{three}  %
\newcommand{\qStdcfgXOneBceThreeKillHi}{0.31}  %
\newcommand{\qStdcfgXOneBceThreeKillLo}{0.27}  %
\newcommand{\qStdcfgXOneBceThreeMarginHi}{0.06}  %
\newcommand{\qStdcfgXOneBceThreeMarginLo}{0.02}  %
\newcommand{\qStdcfgXOneCellCeilFourSOne}{4.01}  %
\newcommand{\qStdcfgXOneCellCeilFourSTwo}{3.99}  %
\newcommand{\qStdcfgXOneCellCeilFourSZero}{4.47}  %
\newcommand{\qStdcfgXOneCellCeilThreeSOne}{2.29}  %
\newcommand{\qStdcfgXOneCellCeilThreeSTwo}{2.28}  %
\newcommand{\qStdcfgXOneCellCeilThreeSZero}{2.65}  %
\newcommand{\qStdcfgXOneCellEscHiFourSOne}{8,000}  %
\newcommand{\qStdcfgXOneCellEscHiFourSTwo}{8,000}  %
\newcommand{\qStdcfgXOneCellEscHiFourSZero}{8,000}  %
\newcommand{\qStdcfgXOneCellEscHiThreeSOne}{7,000}  %
\newcommand{\qStdcfgXOneCellEscHiThreeSTwo}{7,000}  %
\newcommand{\qStdcfgXOneCellEscHiThreeSZero}{7,000}  %
\newcommand{\qStdcfgXOneCellEscLoFourSOne}{8,000}  %
\newcommand{\qStdcfgXOneCellEscLoFourSTwo}{8,000}  %
\newcommand{\qStdcfgXOneCellEscLoFourSZero}{9,000}  %
\newcommand{\qStdcfgXOneCellEscLoThreeSOne}{8,000}  %
\newcommand{\qStdcfgXOneCellEscLoThreeSTwo}{8,000}  %
\newcommand{\qStdcfgXOneCellEscLoThreeSZero}{9,000}  %
\newcommand{\qStdcfgXOneCellFracFourSOne}{0.455}  %
\newcommand{\qStdcfgXOneCellFracFourSTwo}{0.452}  %
\newcommand{\qStdcfgXOneCellFracFourSZero}{0.457}  %
\newcommand{\qStdcfgXOneCellFracThreeSOne}{0.488}  %
\newcommand{\qStdcfgXOneCellFracThreeSTwo}{0.491}  %
\newcommand{\qStdcfgXOneCellFracThreeSZero}{0.504}  %
\newcommand{\qStdcfgXOneCellKillFourSOne}{0.112}  %
\newcommand{\qStdcfgXOneCellKillFourSTwo}{0.112}  %
\newcommand{\qStdcfgXOneCellKillFourSZero}{0.088}  %
\newcommand{\qStdcfgXOneCellKillThreeSOne}{0.310}  %
\newcommand{\qStdcfgXOneCellKillThreeSTwo}{0.313}  %
\newcommand{\qStdcfgXOneCellKillThreeSZero}{0.269}  %
\newcommand{\qStdcfgXOneCellRatioFourSOne}{8.90}  %
\newcommand{\qStdcfgXOneCellRatioFourSTwo}{8.90}  %
\newcommand{\qStdcfgXOneCellRatioFourSZero}{11.36}  %
\newcommand{\qStdcfgXOneCellRatioThreeSOne}{3.23}  %
\newcommand{\qStdcfgXOneCellRatioThreeSTwo}{3.19}  %
\newcommand{\qStdcfgXOneCellRatioThreeSZero}{3.72}  %
\newcommand{\qStdcfgXOneDoseFour}{four-decade}  %
\newcommand{\qStdcfgXOneDoseThree}{three-decade}  %
\newcommand{\qStdcfgXOneEpsFour}{10^{-4}}  %
\newcommand{\qStdcfgXOneEpsLo}{10^{-8}}  %
\newcommand{\qStdcfgXOneEpsThree}{10^{-5}}  %
\newcommand{\qStdcfgXOneEscBceHi}{9,000}  %
\newcommand{\qStdcfgXOneEscBceLo}{7,000}  %
\newcommand{\qStdcfgXOneEscSoftmaxHi}{10,000}  %
\newcommand{\qStdcfgXOneEscSoftmaxLo}{9,000}  %
\newcommand{\qStdcfgXOneEscWindow}{1,000}  %
\newcommand{\qStdcfgXOneExtBceFireSide}{low}  %
\newcommand{\qStdcfgXOneExtBceFourN}{two}  %
\newcommand{\qStdcfgXOneExtBceFourOf}{two}  %
\newcommand{\qStdcfgXOneExtBceFourRatioHi}{9.5}  %
\newcommand{\qStdcfgXOneExtBceFourRatioLo}{8.8}  %
\newcommand{\qStdcfgXOneExtBceThreeInN}{two}  %
\newcommand{\qStdcfgXOneExtBceThreeInOf}{two}  %
\newcommand{\qStdcfgXOneExtBceThreeKillHi}{0.32}  %
\newcommand{\qStdcfgXOneExtBceThreeKillLo}{0.29}  %
\newcommand{\qStdcfgXOneExtBceThreeMarginHi}{0.07}  %
\newcommand{\qStdcfgXOneExtBceThreeMarginLo}{0.04}  %
\newcommand{\qStdcfgXOneExtCellCeilBceFourSFour}{4.10}  %
\newcommand{\qStdcfgXOneExtCellCeilBceFourSThree}{3.88}  %
\newcommand{\qStdcfgXOneExtCellCeilBceThreeSFour}{2.36}  %
\newcommand{\qStdcfgXOneExtCellCeilBceThreeSThree}{2.17}  %
\newcommand{\qStdcfgXOneExtCellCeilSoftmaxFourSFour}{3.88}  %
\newcommand{\qStdcfgXOneExtCellCeilSoftmaxFourSThree}{3.76}  %
\newcommand{\qStdcfgXOneExtCellEscHiBceFourSFour}{8,000}  %
\newcommand{\qStdcfgXOneExtCellEscHiBceFourSThree}{8,000}  %
\newcommand{\qStdcfgXOneExtCellEscHiBceThreeSFour}{7,000}  %
\newcommand{\qStdcfgXOneExtCellEscHiBceThreeSThree}{7,000}  %
\newcommand{\qStdcfgXOneExtCellEscHiSoftmaxFourSFour}{10,000}  %
\newcommand{\qStdcfgXOneExtCellEscHiSoftmaxFourSThree}{9,000}  %
\newcommand{\qStdcfgXOneExtCellEscLoBceFourSFour}{8,000}  %
\newcommand{\qStdcfgXOneExtCellEscLoBceFourSThree}{7,000}  %
\newcommand{\qStdcfgXOneExtCellEscLoBceThreeSFour}{8,000}  %
\newcommand{\qStdcfgXOneExtCellEscLoBceThreeSThree}{7,000}  %
\newcommand{\qStdcfgXOneExtCellEscLoSoftmaxFourSFour}{10,000}  %
\newcommand{\qStdcfgXOneExtCellEscLoSoftmaxFourSThree}{10,000}  %
\newcommand{\qStdcfgXOneExtCellFracBceFourSFour}{0.451}  %
\newcommand{\qStdcfgXOneExtCellFracBceFourSThree}{0.441}  %
\newcommand{\qStdcfgXOneExtCellFracBceThreeSFour}{0.471}  %
\newcommand{\qStdcfgXOneExtCellFracBceThreeSThree}{0.468}  %
\newcommand{\qStdcfgXOneExtCellFracSoftmaxFourSFour}{0.512}  %
\newcommand{\qStdcfgXOneExtCellFracSoftmaxFourSThree}{0.526}  %
\newcommand{\qStdcfgXOneExtCellKillBceFourSFour}{0.105}  %
\newcommand{\qStdcfgXOneExtCellKillBceFourSThree}{0.114}  %
\newcommand{\qStdcfgXOneExtCellKillBceThreeSFour}{0.287}  %
\newcommand{\qStdcfgXOneExtCellKillBceThreeSThree}{0.316}  %
\newcommand{\qStdcfgXOneExtCellKillSoftmaxFourSFour}{0.151}  %
\newcommand{\qStdcfgXOneExtCellKillSoftmaxFourSThree}{0.168}  %
\newcommand{\qStdcfgXOneExtCellRatioBceFourSFour}{9.50}  %
\newcommand{\qStdcfgXOneExtCellRatioBceFourSThree}{8.78}  %
\newcommand{\qStdcfgXOneExtCellRatioBceThreeSFour}{3.49}  %
\newcommand{\qStdcfgXOneExtCellRatioBceThreeSThree}{3.17}  %
\newcommand{\qStdcfgXOneExtCellRatioSoftmaxFourSFour}{6.64}  %
\newcommand{\qStdcfgXOneExtCellRatioSoftmaxFourSThree}{5.95}  %
\newcommand{\qStdcfgXOneExtEscBceHi}{8,000}  %
\newcommand{\qStdcfgXOneExtEscBceLo}{7,000}  %
\newcommand{\qStdcfgXOneExtEscSoftmaxHi}{10,000}  %
\newcommand{\qStdcfgXOneExtEscSoftmaxLo}{9,000}  %
\newcommand{\qStdcfgXOneExtSeedHi}{4}  %
\newcommand{\qStdcfgXOneExtSeedLo}{3}  %
\newcommand{\qStdcfgXOneExtSetFires}{FIRES}  %
\newcommand{\qStdcfgXOneExtSetNotCovered}{NOT-COVERED}  %
\newcommand{\qStdcfgXOneExtSetNotFiring}{NOT-FIRING}  %
\newcommand{\qStdcfgXOneExtSoftmaxFireSide}{low}  %
\newcommand{\qStdcfgXOneExtSoftmaxFourN}{two}  %
\newcommand{\qStdcfgXOneExtSoftmaxFourOf}{two}  %
\newcommand{\qStdcfgXOneExtSoftmaxFourRatioHi}{6.6}  %
\newcommand{\qStdcfgXOneExtSoftmaxFourRatioLo}{6.0}  %
\newcommand{\qStdcfgXOneExtWordBceFourSFour}{FIRES}  %
\newcommand{\qStdcfgXOneExtWordBceFourSThree}{FIRES}  %
\newcommand{\qStdcfgXOneExtWordBceThreeSFour}{NOT-FIRING}  %
\newcommand{\qStdcfgXOneExtWordBceThreeSThree}{NOT-FIRING}  %
\newcommand{\qStdcfgXOneExtWordSoftmaxFourSFour}{FIRES}  %
\newcommand{\qStdcfgXOneExtWordSoftmaxFourSThree}{FIRES}  %
\newcommand{\qStdcfgXOneFireSide}{low}  %
\newcommand{\qStdcfgXOneKillHi}{4}  %
\newcommand{\qStdcfgXOneKillLo}{0.25}  %
\newcommand{\qStdcfgXOneLastCommonStep}{60,000}  %
\newcommand{\qStdcfgXOneLowSideCeil}{1.39}  %
\newcommand{\qStdcfgXOneReadEndStep}{59,000}  %
\newcommand{\qStdcfgXOneSeedHi}{2}  %
\newcommand{\qStdcfgXOneSeedLo}{0}  %
\newcommand{\qStdcfgXOneSetGo}{GO}  %
\newcommand{\qStdcfgXOneSetNoGo}{NO-GO}  %
\newcommand{\qStdcfgXOneSetNotCovered}{NOT-COVERED}  %
\newcommand{\qStdcfgXOneSoftmax}{S-GO}  %
\newcommand{\qStdcfgXOneSoftmaxCellCeilFourSOne}{3.58}  %
\newcommand{\qStdcfgXOneSoftmaxCellCeilFourSTwo}{3.62}  %
\newcommand{\qStdcfgXOneSoftmaxCellCeilFourSZero}{4.16}  %
\newcommand{\qStdcfgXOneSoftmaxCellEscHiFourSOne}{9,000}  %
\newcommand{\qStdcfgXOneSoftmaxCellEscHiFourSTwo}{10,000}  %
\newcommand{\qStdcfgXOneSoftmaxCellEscHiFourSZero}{9,000}  %
\newcommand{\qStdcfgXOneSoftmaxCellEscLoFourSOne}{10,000}  %
\newcommand{\qStdcfgXOneSoftmaxCellEscLoFourSTwo}{10,000}  %
\newcommand{\qStdcfgXOneSoftmaxCellEscLoFourSZero}{10,000}  %
\newcommand{\qStdcfgXOneSoftmaxCellFracFourSOne}{0.552}  %
\newcommand{\qStdcfgXOneSoftmaxCellFracFourSTwo}{0.541}  %
\newcommand{\qStdcfgXOneSoftmaxCellFracFourSZero}{0.509}  %
\newcommand{\qStdcfgXOneSoftmaxCellKillFourSOne}{0.201}  %
\newcommand{\qStdcfgXOneSoftmaxCellKillFourSTwo}{0.190}  %
\newcommand{\qStdcfgXOneSoftmaxCellKillFourSZero}{0.130}  %
\newcommand{\qStdcfgXOneSoftmaxCellRatioFourSOne}{4.98}  %
\newcommand{\qStdcfgXOneSoftmaxCellRatioFourSTwo}{5.27}  %
\newcommand{\qStdcfgXOneSoftmaxCellRatioFourSZero}{7.71}  %
\newcommand{\qStdcfgXOneSoftmaxFireSide}{low}  %
\newcommand{\qStdcfgXOneSoftmaxFourN}{three}  %
\newcommand{\qStdcfgXOneSoftmaxFourOf}{three}  %
\newcommand{\qStdcfgXOneSoftmaxFourRatioHi}{7.7}  %
\newcommand{\qStdcfgXOneSoftmaxFourRatioLo}{5.0}  %
\newcommand{\qStdcfgXOneSoftmaxSetGo}{S-GO}  %
\newcommand{\qStdcfgXOneSoftmaxSetNoGo}{S-NO-GO}  %
\newcommand{\qStdcfgXOneSoftmaxSetNotCovered}{NOT-COVERED}  %
\newcommand{\qStdcfgXOneSoftmaxWordFourSOne}{FIRES}  %
\newcommand{\qStdcfgXOneSoftmaxWordFourSTwo}{FIRES}  %
\newcommand{\qStdcfgXOneSoftmaxWordFourSZero}{FIRES}  %
\newcommand{\qStdcfgXOneWordFourSOne}{FIRES}  %
\newcommand{\qStdcfgXOneWordFourSTwo}{FIRES}  %
\newcommand{\qStdcfgXOneWordFourSZero}{FIRES}  %
\newcommand{\qStdcfgXOneWordThreeSOne}{NOT-FIRING}  %
\newcommand{\qStdcfgXOneWordThreeSTwo}{NOT-FIRING}  %
\newcommand{\qStdcfgXOneWordThreeSZero}{NOT-FIRING}  %
\newcommand{\qStdcfgXSeven}{four-decade fire holds on 8/8 panel rows (one trajectory, seed 0)}  %
\newcommand{\qStdcfgXSevenExpFHi}{12.54}  %
\newcommand{\qStdcfgXSevenExpFLo}{5.56}  %
\newcommand{\qStdcfgXSevenInert}{INERT}  %
\newcommand{\qStdcfgXSevenMaxOverMin}{2.26}  %
\newcommand{\qStdcfgXSevenOldPanelHi}{21.946}  %
\newcommand{\qStdcfgXSevenOldPanelLo}{6.879}  %
\newcommand{\qStdcfgXSevenProbeExpF}{11.36}  %
\newcommand{\qStdcfgXSevenReadItems}{960 / 960}  %
\newcommand{\qStdcfgXSixBce}{NEITHER}  %
\newcommand{\qStdcfgXSixBceControl}{NEITHER}  %
\newcommand{\qStdcfgXSixBceDbarFour}{-0.4227}  %
\newcommand{\qStdcfgXSixBceDbarThree}{-0.7935}  %
\newcommand{\qStdcfgXSixBceSigma}{0.1324}  %
\newcommand{\qStdcfgXSixRareFromTail}{482}  %
\newcommand{\qStdcfgXSixRarePositions}{500}  %
\newcommand{\qStdcfgXSixSetA}{A}  %
\newcommand{\qStdcfgXSixSetB}{B}  %
\newcommand{\qStdcfgXSixSetNeither}{NEITHER}  %
\newcommand{\qStdcfgXSixSoftmax}{B}  %
\newcommand{\qStdcfgXSixSoftmaxControl}{B}  %
\newcommand{\qStdcfgXSixSoftmaxControlDirection}{up}  %
\newcommand{\qStdcfgXSixSoftmaxDbarFour}{-3.4925}  %
\newcommand{\qStdcfgXSixSoftmaxDirection}{down}  %
\newcommand{\qStdcfgXSixSoftmaxSigma}{0.0576}  %
\newcommand{\qStdcfgXSixTailRows}{364}  %
\newcommand{\qXElevenFactor}{1.1}  %
\newcommand{\qXElevenLSeedOne}{STRONGER}  %
\newcommand{\qXElevenLSeedTwo}{STRONGER}  %
\newcommand{\qXElevenLSeedZero}{STRONGER}  %
\newcommand{\qXElevenLValSeedOne}{0.2017}  %
\newcommand{\qXElevenLValSeedTwo}{0.1785}  %
\newcommand{\qXElevenLValSeedZero}{0.2268}  %
\newcommand{\qXElevenNullScale}{0.0945}  %
\newcommand{\qXElevenPrediction}{WEAKER}  %
\newcommand{\qXElevenSetSame}{SAME}  %
\newcommand{\qXElevenSetStronger}{STRONGER}  %
\newcommand{\qXElevenSetWeaker}{WEAKER}  %
\newcommand{\stridePFiveOutcome}{NULL-MONOTONE}  %
\newcommand{\stridePFourOutcome}{PASS}  %
\newcommand{\wallNinetyOutcome}{NEITHER}  %

\newcommand{\artifacturl}{not linked in this version (code and data will be linked in a revision)}
\newcommand{\eps}{\varepsilon}
\newcommand{\slot}[2]{\par\noindent\fbox{\parbox{0.96\linewidth}{\footnotesize\textbf{[#1]}~#2}}\par} %
\newcommand{\figslot}[3]{\IfFileExists{figs/#1}{\includegraphics[width=\linewidth]{#1}}{\slot{#2}{#3}}}
\newsavebox{\protocolboxbin}
\newenvironment{protocolbox}{\begin{lrbox}{\protocolboxbin}\begin{minipage}{0.94\linewidth}\footnotesize}%
  {\end{minipage}\end{lrbox}\par\noindent\fbox{\usebox{\protocolboxbin}}\par}
\newcommand{\supported}[1]{}
\newcommand{\interpretation}[1]{#1}

\icmltitlerunning{Replaying the same gradients at a different epsilon}

\begin{document}
\raggedbottom

\twocolumn[
  \icmltitle{Replaying the Same Gradients at a Different \texorpdfstring{$\varepsilon$}{ε}:\\
  A Registered Test of the Fixed-Trajectory Account of Adam's \texorpdfstring{$\varepsilon$}{ε} for One Rare Row}
  \icmlsetsymbol{equal}{*}
  \begin{icmlauthorlist}
    \icmlauthor{Lngel Kyo}{}\\
    A2O Labs\\
    \mbox{lngel.kyo@a2o-labs.com}
  \end{icmlauthorlist}
  \icmlkeywords{Adam, epsilon, counterfactual replay, optimizer state, rare tokens}
  \vskip 0.3in %
]

\csname icml@noticeprintedtrue\endcsname

\begin{abstract}
Adam divides every update by the square root of a running average of squared gradients plus a small constant,
epsilon. For an output-embedding row whose token is rare, which we call the probe row, raising epsilon can be read
as shrinking that row's steps by arithmetic alone. We test that reading directly. While a training run proceeds, we
also divide the probe row's update by a different epsilon, using the run's own moment estimates; this replay is exact
arithmetic on the gradients the run actually saw, so it holds the run's trajectory fixed. A rule for when the replay
counts as wrong was written down before the runs.

We train a small causal transformer in a standard configuration: pre-LN with a final norm, a \qStdcfgWarmup-step
warmup and a cosine schedule, and a peak learning rate chosen by a rule registered in advance and applied to the
low-epsilon runs alone. When epsilon rises by four orders of magnitude, the replay counts as wrong on each of the
\qStdcfgXOneBceFourOf{} seeds registered for the test: the probe row keeps
\qStdcfgXOneBceFourRatioLo--\qStdcfgXOneBceFourRatioHi{} times more of its displacement than predicted with the main,
sampled-negative head, and \qStdcfgXOneSoftmaxFourRatioLo--\qStdcfgXOneSoftmaxFourRatioHi{} times more with a
full-softmax head scored as a second instrument. At three orders of magnitude, tested with the main head only, it does
not, by narrow margins; a further \qStdcfgXOneExtBceFourOf{} seeds, registered after this result had been committed
and scored on their own, follow the same pattern. Every one of these runs still falls through the midpoint of its loss
curve, at a step printed and never used to select a run; the learning-rate rule screens out a stall above that
midpoint before a late drop, not the fall itself. In separate real runs, raising epsilon on the probe row alone is enough to reproduce the row's change and
raising it on every other parameter alone also moves the row part of the way --- readings about the runs that do not
rescue the replay; scaling the row's learning rate instead, with the denominator untouched, is mispredicted by the
replay in the same direction.

This is one embedding row of one model family on one corpus; we make no claim about rates, about rows in general (other rows
were read only as a panel on one trajectory) or about how the effect changes with model size. The registrations and the scoring scripts are to ship with the paper.
\end{abstract}

\section{Introduction}
\label{sec:intro}

Adam divides each update by the square root of a running average of squared gradients plus a small constant, usually
written epsilon. The constant is conventionally a guard against division by zero, but wherever the squared-gradient
average is small next to it, it also shortens the step. What it does to one parameter row whose token is rare is less settled. The
natural explanation for any effect of epsilon on such a row is arithmetic: a larger constant makes the denominator
larger, so the row moves less.

That explanation can be tested without assuming anything about the network. While a run trains, divide one rare row's
update by a second constant as well, using the run's own moment estimates; since those estimates do not depend on the
constant, this equals integrating the run's gradients through a second Adam state that differs only in the constant
(Section~\ref{sec:setup}). If the arithmetic account were complete, this replay would predict how far the row moves
when the run itself uses the new constant. We test that prediction for one rare row, the probe row, in a standard
training configuration, against a rule registered before the runs for when the replay counts as wrong. We then take
the change apart in separate real runs, giving the new
constant to the probe row alone or to every other parameter alone, or giving the probe row a matched learning-rate
change instead.

\vspace{0pt plus 1fil}\begingroup\csname @beginparpenalty\endcsname=10000
\paragraph{Contributions.}
\begin{enumerate}\setlength{\itemsep}{1pt}\setlength{\parskip}{0pt}
\item A registered test of the replay on one rare row (Section~\ref{sec:x1}), in a standard configuration --- pre-LN
with a final norm, a \qStdcfgWarmup-step warmup, a cosine schedule, and a peak learning rate chosen by a registered rule
on the low-epsilon runs alone: at four orders of magnitude the replay over-predicts the row's suppression on
every seed registered for the test, with both output heads; at three, tested with the main head only, it is not counted
wrong. Every run still falls through the midpoint of its loss curve, at a step that is printed and never used to
select a run; the learning-rate rule screens out a stall above the midpoint before a late drop, not this fall.
\item A decomposition by per-row step size (Section~\ref{sec:decomp}): in separate real runs, raising the row's own
epsilon alone is enough to reproduce the row's change, raising every other parameter's epsilon alone also moves it part
of the way, and a per-row learning-rate change made without touching the denominator is mispredicted by the replay in
the same direction.
\item A second, wider model, registered on its own (Section~\ref{sec:w768}), whose tests are reported per seed or per
cell, each with the full set of outcomes its registration allowed and the one it reached, and never pooled with the
first.
\end{enumerate}\endgroup

\section{Setup and instrument}
\label{sec:setup}

\paragraph{Adam and its constant.}
For one coordinate with gradient $g_t$ at step $t$, Adam keeps two running moments and moves the weight by
\begin{equation}\label{eq:adam}
\begin{gathered}
m_t=\beta_1 m_{t-1}+(1-\beta_1)\,g_t,\;v_t=\beta_2 v_{t-1}+(1-\beta_2)\,g_t^2,\\
w_t=w_{t-1}-\alpha_t\,\frac{\hat m_t}{\sqrt{\hat v_t}+\eps},
\end{gathered}
\end{equation}
where $\hat m_t=m_t/(1-\beta_1^{\,t})$ and $\hat v_t=v_t/(1-\beta_2^{\,t})$ are the bias-corrected moments, $\alpha_t$ is
the scheduled learning rate and $\eps$ is the constant added to the denominator. Wherever $\sqrt{\hat v_t}$ is not large
next to $\eps$, raising $\eps$ shortens the step. The base runs use $\eps=\qStdcfgXOneEpsLo$; a run at a dose of $k$ decades
uses $10^{k}$ times that value, and the tests use the \qStdcfgXOneDoseThree{} dose ($\eps=\qStdcfgXOneEpsThree$) and the
\qStdcfgXOneDoseFour{} dose ($\eps=\qStdcfgXOneEpsFour$).

\paragraph{The standard configuration.}
The model is a causal transformer with LayerNorm before every sublayer and a final LayerNorm (pre-LN), of width
\qStdcfgDModel{}, with \qStdcfgLayers{} layers, \qStdcfgHeads{} attention heads and a vocabulary of \qStdcfgVocab{}
subword tokens (the TinyLlama tokenizer's). It is trained on WikiText-103 \citep{merity2017pointer} with next-token
targets and no dropout, on \qStdcfgBatch{} sequences of \qStdcfgSeqLen{} tokens per step, for \qStdcfgSteps{} steps. The
learning rate rises linearly to its peak over \qStdcfgWarmup{} steps and then follows a cosine down to
\qStdcfgLrMinFrac{} of the peak at the last step. Within a seed and head, the runs of a test differ only in $\eps$ (or,
in the arms of Section~\ref{sec:decomp}, in the probe row's $\eps$ alone, in every other parameter's $\eps$, or in the
probe row's learning rate), apart from side computations for another test (Section~\ref{sec:decomp}). That these leave
training unchanged was checked before the kill test, on each machine it used, with a shorter pair of runs with the
computations on and off, and at full length by the one repeat described below; the other runs, the high-$\eps$ runs
among them, carry the computations without a check of their own (Appendix~\ref{app:cert}).

\paragraph{Two output heads and the probe row.}
Every reading in the main text concerns one output-embedding row, the probe, whose token occurs \qStdcfgProbeFreq{}
times in the corpus; two controls in Appendix~\ref{app:decomp-controls} read other quantities, the cross-entropy at
held-out rare-target positions and a panel of other rows on one trajectory. The main instrument is a sampled-negative
binary cross-entropy (BCE) head: at each position the target's logit is trained up and the logits of
\qStdcfgNegatives{} uniformly drawn tokens are trained down, the draws being fixed by the seed and the step, so $\eps$
never enters them. The second instrument is the full softmax cross-entropy, whose gradient reaches every output row at
every step. The heads are read separately; the softmax readings are printed beside the main ones and never enter their
outcomes. A seed sets the initialization and the main head's negative draws; the batches arrive in the same fixed order
in every seed, so the probe's token occurs at the same steps in every run.

\paragraph{The learning-rate rule.}
The peak learning rate was chosen by a rule registered before any test run, not tuned. It read only the main head's
low-$\eps$ runs: each candidate peak (\qStdcfgLrCandOne, \qStdcfgLrCandTwo{} and \qStdcfgLrCandThree) was trained for the
first \qStdcfgLrSteps{} steps of the schedule on each of seeds \qStdcfgXOneSeedLo--\qStdcfgXOneSeedHi{} and qualified
if, on every seed, the training loss averaged over \qStdcfgXOneEscWindow-step windows fell from its early level by more
than a registered margin, did not hold nearly flat for three windows above the midpoint between its early and late
levels before crossing it, and after warmup never rose between windows by more than a small registered tolerance; the
lowest loss over the last ten windows, averaged over the seeds, then wins. Every candidate qualified on every seed, and
the rule took \qStdcfgLrArgv, the largest. The rule never reads a kill ratio, but it does read the seeds the tests then
use. Of the shapes it screens out, the one at issue is a stall above the midpoint followed by a late drop (it also
excludes a loss that never falls and one that rises after warmup); none of its runs showed any, and it does not screen
out the fall itself: every run of the kill test on seeds \qStdcfgXOneSeedLo--\qStdcfgXOneSeedHi{} still falls through its
loss midpoint, with the main head between steps \qStdcfgXOneEscBceLo{} and \qStdcfgXOneEscBceHi{} and with the softmax
head between \qStdcfgXOneEscSoftmaxLo{} and \qStdcfgXOneEscSoftmaxHi, the runs of a pair not always in the same window.
These crossings are printed beside every kill-test cell (Appendix~\ref{app:x1}) and never used to select a run or a
seed. The tests do not separate ``holds in a standard recipe'' from ``holds at this learning rate, warmup and budget''
(Section~\ref{sec:limits}).

\paragraph{The counterfactual accumulator.}
While a run trains, an accumulator divides the probe row's update a second time, using the run's own moment estimates
and learning rate, with only the denominator constant replaced by $\eps'$; because the moments do not depend on $\eps$,
this equals integrating the run's gradients through a second Adam state that differs only in that constant:
\begin{equation}\label{eq:cf}
W_{\mathrm{cf}}[\eps']=-\sum_{t\le T}\alpha_t\,\frac{\hat m_t}{\sqrt{\hat v_t}+\eps'}\qquad\text{(probe row only),}
\end{equation}
where $T$ is the last step at which the accumulators add. At $\eps'=\eps$ the sum is the row's actual displacement
$\Delta w=w_T-w_0$. That identity is the instrument's self-calibration: every run checks the norm of the sum against the
norm of the row's actual displacement to a registered relative tolerance and is void if it fails, and every run of the
kill test, with both output heads, passed it. Because the moments are the run's own, $W_{\mathrm{cf}}[\eps']$ is exact
arithmetic, not a linearisation: it is how far the row would have moved had the same gradients been divided by a
different constant. It holds the trajectory fixed, and that is the assumption under test. We call it the replay.

\paragraph{What is estimated.}
For a seed and head, a \emph{pair} is the run at the base $\eps_{\mathrm{lo}}$ (lo) and a run at a dose
$\eps_{\mathrm{hi}}$ (hi), on one machine, with accumulators $W^{\mathrm{lo}}_{\mathrm{cf}}$ and
$W^{\mathrm{hi}}_{\mathrm{cf}}$; a pair's reading is a \emph{cell}. Then
\begin{equation}\label{eq:estimands}
\begin{gathered}
\rho_{\mathrm{cf}}=\frac{\lVert W^{\mathrm{lo}}_{\mathrm{cf}}[\eps_{\mathrm{hi}}]\rVert}
                        {\lVert W^{\mathrm{lo}}_{\mathrm{cf}}[\eps_{\mathrm{lo}}]\rVert},\qquad
\rho_{\mathrm{meas}}=\frac{\lVert W^{\mathrm{hi}}_{\mathrm{cf}}[\eps_{\mathrm{hi}}]\rVert}
                          {\lVert W^{\mathrm{lo}}_{\mathrm{cf}}[\eps_{\mathrm{lo}}]\rVert},\\
F=\ln\rho_{\mathrm{meas}}-\ln\rho_{\mathrm{cf}},\qquad \rho_{\mathrm{meas}}=\rho_{\mathrm{cf}}^{\,s}.
\end{gathered}
\end{equation}
$\rho_{\mathrm{cf}}$ is the replay's prediction: the displacement ratio that the low run's own gradients give
when divided by the higher constant. $\rho_{\mathrm{meas}}$ is the ratio that the two real runs produce, each
accumulator at its run's own $\eps$ being that run's displacement. $F$, in nats, is zero exactly when the prediction is
right. The exponent $s=\ln\rho_{\mathrm{meas}}/\ln\rho_{\mathrm{cf}}$ is the realised fraction, the fraction of the
predicted log-suppression that the runs realise ($s=1$ when the prediction is right); with $c=-\ln\rho_{\mathrm{cf}}$,
$F=(1-s)\,c$. The kill ratio is $\rho_{\mathrm{cf}}/\rho_{\mathrm{meas}}=e^{-F}$, and a cell \emph{fires}, that is,
the replay counts as wrong for it, when the ratio falls outside $[\qStdcfgXOneKillLo,\qStdcfgXOneKillHi]$. Whenever
the high run moves the row no more than the low run, $F\le c$, so a fire below the interval needs
$c>\ln\qStdcfgXOneKillHi$; $c$ depends on the low run alone and is printed for every cell before the high run is read. Ratios are taken at the last read point both runs report (Appendix~\ref{app:x1}).

\paragraph{Determinism is a certificate, not a noise floor.}
Runs are compiled and use deterministic algorithms; the softmax head's runs set the check to warn rather than refuse,
a registered mode. The two runs of a pair share mode, build and machine. Two inputs of Section~\ref{sec:decomp} come
from runs on another machine, the learning-rate arm's factor and a later comparison's shadows. The factor, and the
shadows carried by the kill test's low-$\eps$ runs, are admitted through a per-seed whole-run comparison of that run
with the arms' own low-$\eps$ run; the shadows carried by the high-$\eps$ runs are set against arms whose whole-run
comparison with those runs was registered to be printed, not as a gate (Appendix~\ref{app:decomp-reg}). No value is
set equal across machines beyond these comparisons. A main-head run of the kill test was repeated on its machine with the same training command, minus the side computations its twin carries: of
\qStdcfgCertLeaves{} compared entries, \qStdcfgCertUndeclared{} differed outside the declared differences. A single
repeat says only that this repeat did not flip; it is neither a rate nor a floor. The softmax head has no such repeat;
its certificate is its machine's instrument gates (Appendix~\ref{app:cert}); the second width's certificate is
reported with its reads (Appendix~\ref{app:w768}). The noise that matters is between seeds: the kill test and the per-row tests are read per seed (only the
learning-rate rule and the held-out control of Appendix~\ref{app:decomp-controls} average over seeds, as registered),
and no run or seed is selected by when its loss fell through its midpoint.

\begin{protocolbox}
\textbf{Rate rule} (\path{PREREG_STDCFG.md} \S1.2 and \S1.3 as amended by AMENDMENT-1 A1.1; re-run on another machine
under AMENDMENT-9 after the first read's machine was lost, and frozen by AMENDMENT-12; row \texttt{Q-STDCFG-LR}): a
candidate qualifies when none of its runs is \texttt{P-plateau} and every one is \texttt{P-monotone}; among qualifying
candidates the lowest mean loss over the last ten windows, averaged over the seeds, wins, ties to the larger rate. Outcome set
\{\qStdcfgLrSetSelected, \qStdcfgLrSetNone, \qStdcfgLrSetRefused\}; read \textbf{\qStdcfgLr} (\qStdcfgLrArgv).
\textbf{Rate rule at width \qStdcfgSevenSixEightDModel} (\path{PREREG_STDCFG_768.md} \S1.3, \S1.5; row
\texttt{Q-STDCFG768-LR}), the same predicates and choice run again at that width: outcome set
\{\qStdcfgSevenSixEightLrSetSelected, \qStdcfgSevenSixEightLrSetNone, \qStdcfgSevenSixEightLrSetNotCovered,
\qStdcfgSevenSixEightLrSetMachineLost\}; read \textbf{\qStdcfgSevenSixEightLr} (\qStdcfgSevenSixEightLrArgv), printed
beside the base width's and not compared with it. Rows are in \path{scripts/claims.jsonl}.
\end{protocolbox}

\section{Results}
\label{sec:results}

Each subsection reports one registered test or one family of them, per seed and per population; the outcome words the
registrations allowed, and the ones they reached, are in the framed box of each subsection; for
Section~\ref{sec:decomp} the words per seed or cell and the outcome sets are in Appendix~\ref{app:decomp}.

\subsection{The kill test in the standard configuration}
\label{sec:x1}

\begin{figure}[t]
\figslot{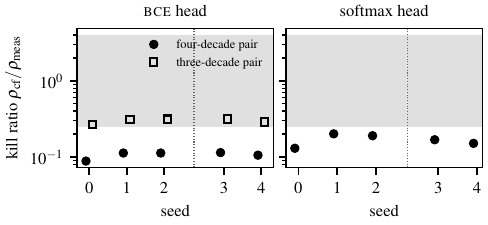}{FIG-1}{The kill-ratio figure is drawn from its rows by \texttt{icml\_figs.py}; this
frame shows only while the figure file is missing.}
\caption{Kill ratio $\rho_{\mathrm{cf}}/\rho_{\mathrm{meas}}$ of every pair, per seed, on a log scale. Left, the BCE
head: filled, the \qStdcfgXOneDoseFour{} pairs; open, the \qStdcfgXOneDoseThree{} pairs. Right, the softmax head,
\qStdcfgXOneDoseFour{} pairs only. Shaded: the interval $[\qStdcfgXOneKillLo,\qStdcfgXOneKillHi]$ fixed before the
runs; below it, the replay predicts more suppression than the runs show by more than the rule tolerates. Seeds
\qStdcfgXOneExtSeedLo{} and \qStdcfgXOneExtSeedHi{}, right of the dotted rule, were registered after the result for
seeds \qStdcfgXOneSeedLo--\qStdcfgXOneSeedHi{} had been committed and are scored on their own; the two groups stand
side by side and are never pooled.\supported{Q-STDCFG-X1}\supported{Q-STDCFG-X1-SOFTMAX}\supported{Q-STDCFG-X1EXT}}
\label{fig:x1}
\end{figure}

The first test asks whether the replay of Section~\ref{sec:setup} survives a large step in $\eps$. For each seed and
head it pairs a run at $\eps=\qStdcfgXOneEpsLo$ with a run that differs from it only in $\eps$ and in the side
computations of Section~\ref{sec:setup}: $\qStdcfgXOneEpsFour$ with both heads (the \qStdcfgXOneDoseFour{} pair) and
$\qStdcfgXOneEpsThree$ with the BCE head (the \qStdcfgXOneDoseThree{} pair). At the last step the two runs share, it
compares the replayed displacement ratio $\rho_{\mathrm{cf}}$ with the measured one $\rho_{\mathrm{meas}}$ through the
kill ratio $\rho_{\mathrm{cf}}/\rho_{\mathrm{meas}}$. The rule, fixed before any of these runs, says that the replay is
wrong for a pair whose kill ratio lies outside $[\qStdcfgXOneKillLo,\qStdcfgXOneKillHi]$. The BCE head is the main
instrument and passes only if its \qStdcfgXOneDoseFour{} pair leaves the interval on every one of seeds
\qStdcfgXOneSeedLo--\qStdcfgXOneSeedHi{} while its \qStdcfgXOneDoseThree{} pair leaves it on none; the softmax head is a
second instrument, scored on its \qStdcfgXOneDoseFour{} pairs only, beside the first and never inside its pass condition.
The pass condition is therefore a dose pattern fixed before the runs --- a failure beyond the tolerance at the larger
step and none at the smaller one --- and either a missing fire at four decades or a fire at three would have failed it.
Neither half of that condition was settled in advance: in every pair the low-$\eps$ run, scored before its partner,
already left room for a kill ratio below the interval, at three decades as at four (Appendix~\ref{app:x1}).

In the standard configuration (pre-LN with a final norm, a \qStdcfgWarmup-step warmup, a cosine schedule, and
a peak learning rate chosen by the pre-registered rule on the $\eps=\qStdcfgXOneEpsLo$ arm alone), the replay
predicts far more suppression of the probe row than the runs show once $\eps$ moves by four decades, on each head
and in each seed set, while at three decades the BCE pairs stay inside the rule. Per head and seed set:
\begin{enumerate}
\item \emph{BCE head, seeds \qStdcfgXOneSeedLo--\qStdcfgXOneSeedHi.} The \qStdcfgXOneDoseFour{} pair leaves the
interval on its \qStdcfgXOneFireSide{} side on \qStdcfgXOneBceFourN{} of \qStdcfgXOneBceFourOf{} seeds: the measured
displacement ratio is \qStdcfgXOneBceFourRatioLo--\qStdcfgXOneBceFourRatioHi{} times the replayed one. The
\qStdcfgXOneDoseThree{} pair stays inside on \qStdcfgXOneBceThreeInN{} of \qStdcfgXOneBceThreeInOf{} seeds, with kill
ratios \qStdcfgXOneBceThreeKillLo--\qStdcfgXOneBceThreeKillHi, only
\qStdcfgXOneBceThreeMarginLo--\qStdcfgXOneBceThreeMarginHi{} above the lower edge. These results meet the pass
condition written before the runs.\supported{Q-STDCFG-X1}\label{res:xone-bce}
\item \emph{Softmax head, seeds \qStdcfgXOneSeedLo--\qStdcfgXOneSeedHi.} The \qStdcfgXOneDoseFour{} pair leaves the
interval on its \qStdcfgXOneSoftmaxFireSide{} side on \qStdcfgXOneSoftmaxFourN{} of \qStdcfgXOneSoftmaxFourOf{} seeds,
the measured ratio \qStdcfgXOneSoftmaxFourRatioLo--\qStdcfgXOneSoftmaxFourRatioHi{} times the replayed
one.\supported{Q-STDCFG-X1-SOFTMAX}\label{res:xone-softmax}
\item \emph{Seeds \qStdcfgXOneExtSeedLo{} and \qStdcfgXOneExtSeedHi}, registered after the result above had been
committed and scored on their own with the same statistic and interval; they have no pass condition and are never
pooled with seeds \qStdcfgXOneSeedLo--\qStdcfgXOneSeedHi. On the BCE head the \qStdcfgXOneDoseFour{} pair leaves the
interval on its \qStdcfgXOneExtBceFireSide{} side on \qStdcfgXOneExtBceFourN{} of \qStdcfgXOneExtBceFourOf{} seeds
(\qStdcfgXOneExtBceFourRatioLo--\qStdcfgXOneExtBceFourRatioHi{} times the replayed ratio) and the
\qStdcfgXOneDoseThree{} pair stays inside on \qStdcfgXOneExtBceThreeInN{} of \qStdcfgXOneExtBceThreeInOf{} (kill
ratios \qStdcfgXOneExtBceThreeKillLo--\qStdcfgXOneExtBceThreeKillHi,
\qStdcfgXOneExtBceThreeMarginLo--\qStdcfgXOneExtBceThreeMarginHi{} above the edge); on the softmax head the
\qStdcfgXOneDoseFour{} pair leaves it on its \qStdcfgXOneExtSoftmaxFireSide{} side on \qStdcfgXOneExtSoftmaxFourN{}
of \qStdcfgXOneExtSoftmaxFourOf{} seeds
(\qStdcfgXOneExtSoftmaxFourRatioLo--\qStdcfgXOneExtSoftmaxFourRatioHi{} times).\supported{Q-STDCFG-X1EXT}\label{res:xone-ext}
\end{enumerate}

\paragraph{What goes with these numbers.} The \qStdcfgXOneDoseThree{} pairs stay inside the interval by small
margins; that half of the test is reported with its margins and nothing further is drawn from it. Every run crosses
the midpoint of its loss curve, and the learning-rate rule screened out a stall above the midpoint followed by a late
drop, not this crossing: seeds \qStdcfgXOneSeedLo--\qStdcfgXOneSeedHi{} cross as Section~\ref{sec:setup} gives, and
seeds \qStdcfgXOneExtSeedLo--\qStdcfgXOneExtSeedHi, registered after the result for seeds
\qStdcfgXOneSeedLo--\qStdcfgXOneSeedHi{} had been committed, between steps \qStdcfgXOneExtEscBceLo{} and
\qStdcfgXOneExtEscBceHi{} with the BCE head and \qStdcfgXOneExtEscSoftmaxLo{} and \qStdcfgXOneExtEscSoftmaxHi{} with
the softmax head. The steps are listed per pair in Appendix~\ref{app:x1}, with the realised fraction $s$ of every cell
(a description, not a law); no pair was kept or dropped because of them.

\paragraph{What the test does not separate.} It separates ``the \qStdcfgXOneDoseFour{} pair leaves the interval and
the \qStdcfgXOneDoseThree{} pair does not'' from every other combination, and nothing more: not the standard
configuration from this learning rate, warmup and budget, and no rate (Section~\ref{sec:limits}).

\begin{protocolbox}
\textbf{X1, the kill test} (\path{PREREG_STDCFG.md} \S2; softmax head AMENDMENT-4; seeds
\qStdcfgXOneExtSeedLo{} and \qStdcfgXOneExtSeedHi{} AMENDMENT-23 and AMENDMENT-25). Per pair, $\eps$ low
$\qStdcfgXOneEpsLo$, high $\qStdcfgXOneEpsFour$ (\qStdcfgXOneDoseFour) or $\qStdcfgXOneEpsThree$
(\qStdcfgXOneDoseThree), same seed, head, machine and mode: kill $=\rho_{\mathrm{cf}}/\rho_{\mathrm{meas}}$ at the
last common step; a cell fires when kill lies outside $[\qStdcfgXOneKillLo,\qStdcfgXOneKillHi]$, endpoints not
firing; whenever the high run moves the row no more than the low run, a low-side fire needs
$c=-\ln\rho_{\mathrm{cf}}>\ln\qStdcfgXOneKillHi\approx\qStdcfgXOneLowSideCeil$, printed from the low-$\eps$ runs before
any high-$\eps$ run was read; a cell that fires low without that headroom contradicts the ceiling.
\emph{BCE, seeds \qStdcfgXOneSeedLo--\qStdcfgXOneSeedHi}, outcome set \{\qStdcfgXOneSetGo, \qStdcfgXOneSetNoGo,
\qStdcfgXOneSetNotCovered\}: \qStdcfgXOneSetGo{} iff the \qStdcfgXOneDoseFour{} cell fires on every seed and the
\qStdcfgXOneDoseThree{} cell on none; a seed whose pair fails a run gate, or whose cell contradicts the ceiling, makes
it \qStdcfgXOneSetNotCovered. Outcome: \textbf{\qStdcfgXOne} (cell words in Appendix~\ref{app:x1}).
\emph{Softmax, seeds \qStdcfgXOneSeedLo--\qStdcfgXOneSeedHi, \qStdcfgXOneDoseFour{} only, beside
\qStdcfgXOneSetGo{} and not part of it}, outcome set \{\qStdcfgXOneSoftmaxSetGo, \qStdcfgXOneSoftmaxSetNoGo,
\qStdcfgXOneSoftmaxSetNotCovered\}: \qStdcfgXOneSoftmaxSetGo{} iff the cell fires on every seed. Outcome:
\textbf{\qStdcfgXOneSoftmax}.
\emph{Seeds \qStdcfgXOneExtSeedLo{} and \qStdcfgXOneExtSeedHi}, registered after the two outcomes above were
committed; scored alone, counted per head and dose, no outcome word; cell words \{\qStdcfgXOneExtSetFires,
\qStdcfgXOneExtSetNotFiring, \qStdcfgXOneExtSetNotCovered\}. BCE \qStdcfgXOneDoseFour: \qStdcfgXOneExtWordBceFourSThree,
\qStdcfgXOneExtWordBceFourSFour; BCE \qStdcfgXOneDoseThree: \qStdcfgXOneExtWordBceThreeSThree,
\qStdcfgXOneExtWordBceThreeSFour; softmax \qStdcfgXOneDoseFour: \qStdcfgXOneExtWordSoftmaxFourSThree,
\qStdcfgXOneExtWordSoftmaxFourSFour.
Two declared deviations of these reads are in Appendix~\ref{app:x1}.
Rows \texttt{Q-STDCFG-X1}, \texttt{Q-STDCFG-X1-SOFTMAX}, \texttt{Q-STDCFG-X1EXT} in \path{scripts/claims.jsonl}.
\end{protocolbox}

\subsection{Decomposition: per-row step size}
\label{sec:decomp}

\begin{figure}[t]
\figslot{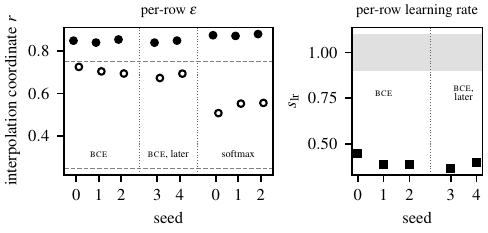}{FIG-2}{Left: $r_{\mathrm{row}}$ (filled) and $r_{\mathrm{rest}}$ (open) per seed, three
populations side by side with a gap between them, the registered thresholds dashed. Right: $s_{\mathrm{lr}}$ per seed, two
populations side by side, the registered near-exact band shaded. Generator: \texttt{scripts/icml\_figs.py x8x9} (contract
in Appendix~\ref{app:decomp}).}
\caption{Per-row step size, per seed; populations side by side, never pooled. \emph{Left:} the interpolation coordinate of
the row-only run ($r_{\mathrm{row}}$, filled) and of the rest-only run ($r_{\mathrm{rest}}$, open) for the sampled-negative
head on seeds 0--2, the same head on seeds 3--4 (a later registration, read on its own) and the softmax head on seeds 0--2
(read on its own); dashed, the registered thresholds $\qStdcfgXEightBLe$ and $\qStdcfgXEightAGe$. $r$ places a run between
the two endpoints; it is not a share. \emph{Right:} the realised fraction $s_{\mathrm{lr}}$ of the learning-rate arm for
the sampled-negative head on seeds 0--2 and 3--4; shaded, the near-exact band
$[\qStdcfgXNineNeutralLo,\qStdcfgXNineNeutralHi]$, in which or above which the prediction would have lost.\supported{Q-STDCFG-X8}\supported{Q-STDCFG-X9}}
\label{fig:decomp}
\end{figure}

Section~\ref{sec:x1} raised $\eps$ for every parameter at once. Two pre-registered experiments take that change apart,
and a third instrument tests the replay itself: is raising the probe row's own $\eps$ alone enough to reproduce the row's
change, does the replay also miss a step-size change made through the learning rate, and does the miss close when the
row alone feels its own replayed position?

\paragraph{Per-row $\eps$.} Write $\eps_{\mathrm{lo}}$ and $\eps_{\mathrm{hi}}$ for the two settings of the four-decade
pair of Section~\ref{sec:x1}, and $D$ for the probe row's displacement at the last accumulation step
(Section~\ref{sec:setup}), whose ratio between two runs is $\rho_{\mathrm{meas}}$. Each seed has four runs: $\eps_{\mathrm{lo}}$ everywhere ($D_{\mathrm{lo}}$),
$\eps_{\mathrm{hi}}$ everywhere ($D_{\mathrm{hi}}$), $\eps_{\mathrm{hi}}$ on the probe row alone ($D_{\mathrm{row}}$), and
$\eps_{\mathrm{hi}}$ on every parameter except the probe row ($D_{\mathrm{rest}}$). The coordinates
\[
  r_{\mathrm{row}} = \frac{D_{\mathrm{lo}} - D_{\mathrm{row}}}{D_{\mathrm{lo}} - D_{\mathrm{hi}}},\qquad
  r_{\mathrm{rest}} = \frac{D_{\mathrm{lo}} - D_{\mathrm{rest}}}{D_{\mathrm{lo}} - D_{\mathrm{hi}}}
\]
say how far each one-sided run moves the row's displacement from the low endpoint toward the high one, in units of the
distance between them. They are \emph{interpolation coordinates}, not shares of an effect: the two need not add up to the
whole, and a value below $\qStdcfgXEightOutBelow$ or above $\qStdcfgXEightOutAbove$ is recorded as a defect of the
registration. The registration reads $r_{\mathrm{row}} \ge \qStdcfgXEightAGe$ as ``the probe row's own $\eps$ is enough
to reproduce that row's change in displacement; \ldots'', $r \le \qStdcfgXEightBLe$ as ``changing only this $\eps$
hardly moves the row; \ldots'', and a value in between as ``both pathways have a part; the design does not subdivide
further''. The words cut at each \ldots{} name a pathway; like the rest, they read these arms alone and are never cited
for or against the replay of Section~\ref{sec:x1}. Each seed is judged alone, and the two one-sided runs are read
together.

\paragraph{Per-row learning rate.} Let $c_s$ be the replay's prediction $\rho_{\mathrm{cf}}$ for the row-only run,
computed for each seed from its low-$\eps$ run of the four-decade pair and frozen before the learning-rate arm ran. The
new arm keeps $\eps_{\mathrm{lo}}$ everywhere and multiplies the probe row's learning rate by $c_s$. On a frozen gradient
stream a learning-rate factor scales the displacement exactly, since Adam's moment estimates do not depend on it, so the
replay predicts the same ratio, $c_s$, for this arm as for the row-only run. With $\rho_{\mathrm{meas}}$ the arm's
displacement over $D_{\mathrm{lo}}$, its realised fraction (Section~\ref{sec:setup}) is
$s_{\mathrm{lr}} = \ln\rho_{\mathrm{meas}}/\ln c_s$.
The registered prediction was a non-negative $s_{\mathrm{lr}}$ below $\qStdcfgXNineNeutralLo$: the replay under-predicts
this arm's displacement, in the same direction as for $\eps$. It loses if $s_{\mathrm{lr}}$ falls in the near-exact band
$[\qStdcfgXNineNeutralLo,\qStdcfgXNineNeutralHi]$, or above it, where the replay would over-predict the displacement.

\paragraph{The shadow.} A real run carries a copy of the probe row with its own Adam state, stepped at a replayed
setting; its gradient uses each step's real hidden state and the real logits of every other output row, recomputes only
the probe row's own logit from the copy's position, and touches no real parameter, optimizer state or loss. It differs
from the replay in one respect only: the row feels its own replayed position. With $\rho_{\mathrm{sh}}$ the copy's
displacement ratio, and $\rho_{\mathrm{cf}}$ and $F$ the replay's prediction and gap in the shadow's own direction (for
the shadow carried by the high run, that run's accumulators re-divided at $\eps_{\mathrm{lo}}$), the closure coordinate
$c_{\mathrm{sh}} = (\ln\rho_{\mathrm{sh}} - \ln\rho_{\mathrm{cf}})/F$ places the shadow between the replay and the
measurement in units of the gap. The registration reads $c_{\mathrm{sh}} \ge \qShadowXThreePXNineLocalGe$ as ``the row's
own position dependence closes the gap'' and $c_{\mathrm{sh}} \le \qShadowXThreePXNineGlobalLe$ as ``the gap lies in the
rest of the network''; below $\qShadowXThreePXNineOutBelow$ or above $\qShadowXThreePXNineOutAbove$ is out of range, a
defect of a criterion that assumed the shadow lands between replay and measurement, and no band surrounds
$\qShadowXThreePXNineOutAbove$. Each seed and direction (the shadow carried by the low run, or by the other run of the
pair) is judged alone; $c_{\mathrm{sh}}$ and $r$ are never set against each other.

\begin{enumerate}
\item \textbf{Per-row $\eps$, population by population.} Sampled-negative head, seeds 0--2:
$r_{\mathrm{row}} = \qStdcfgXEightRowRSeedZero/\qStdcfgXEightRowRSeedOne/\qStdcfgXEightRowRSeedTwo$ and
$r_{\mathrm{rest}} = \qStdcfgXEightRestRSeedZero/\qStdcfgXEightRestRSeedOne/\qStdcfgXEightRestRSeedTwo$; each seed has
$r_{\mathrm{row}} \ge \qStdcfgXEightAGe$ and $r_{\mathrm{rest}}$ between the two thresholds. Seeds 3 and 4, a later
registration read on its own: $r_{\mathrm{row}} = \qStdcfgXEightExtRowRSeedThree/\qStdcfgXEightExtRowRSeedFour$ and
$r_{\mathrm{rest}} = \qStdcfgXEightExtRestRSeedThree/\qStdcfgXEightExtRestRSeedFour$; each seed again has
$r_{\mathrm{row}} \ge \qStdcfgXEightAGe$ and $r_{\mathrm{rest}}$ between the thresholds. Softmax head, seeds 0--2, read
on its own: $r_{\mathrm{row}} = \qStdcfgXEightSoftmaxRowRSeedZero/\qStdcfgXEightSoftmaxRowRSeedOne/\qStdcfgXEightSoftmaxRowRSeedTwo$
and $r_{\mathrm{rest}} = \qStdcfgXEightSoftmaxRestRSeedZero/\qStdcfgXEightSoftmaxRestRSeedOne/\qStdcfgXEightSoftmaxRestRSeedTwo$;
each seed again has $r_{\mathrm{row}} \ge \qStdcfgXEightAGe$ and $r_{\mathrm{rest}}$ between the thresholds. Read as
registered, seed by seed, the probe row's own $\eps$ is enough to reproduce its change in displacement, and for the
rest-only run both pathways have a part. This does not say that the replay is rescued: its miss in
Section~\ref{sec:x1} stands, and neither reading is cited for the other.%
\supported{Q-STDCFG-X8}\supported{Q-STDCFG-X8EXT}\supported{Q-STDCFG-X8SM}\label{find:decomp-eps}

\item \textbf{The replay misses a per-row learning-rate change in the same direction.} Sampled-negative head, seeds
0--2: $s_{\mathrm{lr}} = \qStdcfgXNineSlrSeedZero/\qStdcfgXNineSlrSeedOne/\qStdcfgXNineSlrSeedTwo$, each non-negative and
below $\qStdcfgXNineNeutralLo$, the predicted branch: on each of these seeds the replay under-predicts this arm's
displacement, as it does for $\eps$. Seeds 3 and 4, read on their own:
$s_{\mathrm{lr}} = \qStdcfgXNineExtSlrSeedThree/\qStdcfgXNineExtSlrSeedFour$, each non-negative and below
$\qStdcfgXNineNeutralLo$, the predicted branch.%
\supported{Q-STDCFG-X9}\supported{Q-STDCFG-X9EXT}\label{find:decomp-lr}

\item \textbf{For the learning-rate arm, every cell's coordinate is close to one}, the value at which the shadow
reproduces the measurement (each seed, both directions). Seed 0: $c_{\mathrm{sh}} = \qShadowXThreePXNineCvalForwardSeedZero/\qShadowXThreePXNineCvalReverseSeedZero$
(forward/reverse), in the closure branch. Seeds 1 and 2:
$\qShadowXThreePXNineCvalForwardSeedOne/\qShadowXThreePXNineCvalReverseSeedOne$ and
$\qShadowXThreePXNineCvalForwardSeedTwo/\qShadowXThreePXNineCvalReverseSeedTwo$, above $\qShadowXThreePXNineOutAbove$ and
therefore out of range by registration; no attribution is attached to them.%
\supported{Q-SHADOW-X3PX9}\label{find:decomp-shlr}

\item \textbf{When every parameter's $\eps$ changes, the shadow overshoots.} On the seeds 0--2 pairs of
Section~\ref{sec:x1} --- both directions; both doses with the sampled-negative head, the four-decade dose with softmax ---
the shadow predicts less suppression than the runs show ($\rho_{\mathrm{sh}} > \rho_{\mathrm{meas}}$, so $c_{\mathrm{sh}}$ lies above
$\qShadowXThreePOutAbove$): with the sampled-negative head in \qShadowXThreePBceCellsOut{} of \qShadowXThreePBceCells{}
cells, with the softmax head in \qShadowXThreePSoftmaxCellsOut{} of \qShadowXThreePSoftmaxCells. By registration this is
a defect of the criterion, not an attribution. The shadows carried by the runs of seeds 3 and 4 have not been
read.\supported{Q-SHADOW-X3P}\label{find:decomp-shall}

\item \textbf{A comparison registered after these readings.} After the shadow and per-row readings had been printed, a
further comparison was registered that sets the shadow carried by the low run against the row-only run, and the shadow
carried by the high run against the rest-only run; its coordinate $c'$ is $c_{\mathrm{sh}}$ with that one-sided run in
place of the measurement. With the sampled-negative head, seeds 0--2, it is
$\qShadowXThreePRowCvalPrimeForwardSeedZero/\qShadowXThreePRowCvalPrimeForwardSeedOne/\qShadowXThreePRowCvalPrimeForwardSeedTwo$
forward and
$\qShadowXThreePRowCvalPrimeReverseSeedZero/\qShadowXThreePRowCvalPrimeReverseSeedOne/\qShadowXThreePRowCvalPrimeReverseSeedTwo$
reverse, each inside the registered band $[\qShadowXThreePRowFullLo,\qShadowXThreePRowFullHi]$. These cells are
descriptive, not a test, because their inputs had been printed before the comparison was registered. Its registered
test, on populations whose one-sided arms had not been read when it was registered (seeds 3 and 4 of the
sampled-negative head, whose replay ratios but not shadows had been printed, and the softmax head, whose shadows and
replay ratios had been printed), is not reported here.\supported{Q-SHADOW-X3PROW}\label{find:decomp-shrow}

\item \textbf{The registered prediction for the size of the probe row's gradient did not hold.} The registration
predicted that raising every other parameter's $\eps$ would shrink it: $L$, the log ratio (rest-only over low-$\eps$ run)
of the row's root second-moment estimate summed over the read points, at or below $-\ln\qXElevenFactor$ on each seed.
Sampled-negative head, seeds 0--2: $L = \qXElevenLValSeedZero/\qXElevenLValSeedOne/\qXElevenLValSeedTwo$, each above
$\ln\qXElevenFactor$ and outside the seed spread ($\qXElevenNullScale$). The registered prediction did not hold on any of
the three seeds. The registration wrote a consequence only for an unchanged size; after this outcome was read, a ruling
gave a larger size the same consequence, so the route through the gradient's size is rejected, and the gradient's
coherence, the question registered next, is not reported here.%
\supported{Q-X11}\label{find:decomp-grad}
\end{enumerate}

\begin{protocolbox}
\textbf{Registrations and rows} (every outcome set, and every word per seed or cell, in Appendix~\ref{app:decomp}).
Per-row $\eps$ and learning rate: \path{PREREG_STDCFG.md} \S4--5, with AMENDMENT-35, AMENDMENT-37 and AMENDMENT-38
for the later per-row $\eps$ populations and AMENDMENT-37, AMENDMENT-39 and AMENDMENT-41 (factor frozen by
AMENDMENT-38) for the later learning-rate population; rows \texttt{Q-STDCFG-X8}, \texttt{-X8EXT}, \texttt{-X8SM}, \texttt{-X9}, \texttt{-X9EXT}; the
learning-rate arm's predicted branch \qStdcfgXNineSetSameSide. Shadow: \path{PREREG_SHADOW.md} \S3 and AMENDMENT-1;
rows \texttt{Q-SHADOW-X3P}, \texttt{-X3PX9}, \texttt{-X3PROW}; the all-rows cells' word \qShadowXThreePSetOutOfRange.
Gradient size: \path{PREREG_X11.md}; row \texttt{Q-X11}; prediction \qXElevenPrediction. Rows are in
\path{scripts/claims.jsonl}.
\end{protocolbox}

Two controls of the kill test, not of the decomposition --- the held-out rare-target cross-entropy and a panel of rows on
one trajectory --- are in Appendix~\ref{app:decomp-controls}; every per-seed and per-cell value of this section is in
Appendix~\ref{app:decomp}.

\subsection{Width \qStdcfgSevenSixEightDModel}
\label{sec:w768}

A second batch, registered on its own before any of its runs, repeats the kill test and the two per-row arms at width
\qStdcfgSevenSixEightDModel{} (\qStdcfgSevenSixEightLayers{} layers, \qStdcfgSevenSixEightHeads{} heads; the corpus,
schedule and budget of the base width; peak rate \qStdcfgSevenSixEightLrArgv, chosen again by the rule) and compares
the two sizes. Every test is read per seed or per cell and never pooled with the base width; the widths meet only in
the size comparison. The box at the end gives each test's full set of registered outcomes and the one it reached.

\begin{enumerate}
\item \emph{Kill test, main head.} Only seed~\qStdcfgSevenSixEightXOneSeedZero{} has runs in this test. Its
\qStdcfgSevenSixEightDoseFour{} pair leaves the interval of Section~\ref{sec:x1} on the
\qStdcfgSevenSixEightXOneCellSideFour{} side, the measured
displacement ratio being \qStdcfgSevenSixEightXOneRatioFour{} times the replayed one (kill ratio
\qStdcfgSevenSixEightXOneKillFour); its \qStdcfgSevenSixEightDoseThree{} pair stays inside, with kill ratio
\qStdcfgSevenSixEightXOneKillThree, \qStdcfgSevenSixEightXOneMarginThree{} above the lower edge. The host restarted
before any run of seeds~\qStdcfgSevenSixEightXOneSeedOne{} and~\qStdcfgSevenSixEightXOneSeedTwo{} finished; the test
needs every seed, so it could not be decided. Those seeds were run again after the restart as a separate read,
printed side by side only: it enters neither this test's outcome, nor the softmax head's, nor the size comparison. As
registered, its low-$\eps$ runs were the source of the per-row arms' factor and identity check on
seeds~\qStdcfgSevenSixEightXOneSeedOne{} and~\qStdcfgSevenSixEightXOneSeedTwo; on
seed~\qStdcfgSevenSixEightXOneSeedTwo{} neither was used, because the per-row run they pair with did not finish
(items 5--6; Appendix~\ref{app:w768}).\supported{Q-STDCFG768-X1}\label{res:w768-xone}

\item \emph{Kill test, softmax head}, the second instrument, read beside the main head and never part of its outcome.
Seed~\qStdcfgSevenSixEightXOneSoftmaxSeedZero: the \qStdcfgSevenSixEightDoseFour{} pair leaves the interval on the
\qStdcfgSevenSixEightXOneSoftmaxCellSideSZero{} side, the measured ratio \qStdcfgSevenSixEightXOneSoftmaxRatioSZero{}
times the replayed one. Seed~\qStdcfgSevenSixEightXOneSoftmaxSeedOne: the pair leaves the interval on the
\qStdcfgSevenSixEightXOneSoftmaxCellSideSOne{} side, \qStdcfgSevenSixEightXOneSoftmaxRatioSOne{} times.
Seed~\qStdcfgSevenSixEightXOneSoftmaxSeedTwo: the pair leaves the interval on the
\qStdcfgSevenSixEightXOneSoftmaxCellSideSTwo{} side, \qStdcfgSevenSixEightXOneSoftmaxRatioSTwo{}
times.\supported{Q-STDCFG768-X1-SOFTMAX}\label{res:w768-xone-softmax}

\item \emph{Size comparison, main head.} Let $R$ be a cell's $e^{F}$ (Section~\ref{sec:setup}) in the wider model over
its $e^{F}$ at the base width. A cell meets the registered criterion when the two widths' \qStdcfgSevenSixEightDoseFour{}
pairs agree in whether they leave the interval and $R$ lies in
$[\qStdcfgSevenSixEightSizeBandLo,\qStdcfgSevenSixEightSizeBandHi]$; a cell is judged
only where the runs of both widths shared one machine. For the main head that is
seed~\qStdcfgSevenSixEightXOneSeedZero, whose base-width pair was trained again on the wider model's machine for this
comparison. Inside the size read, before the reader took any wider-model high-$\eps$ value (the wider model's kill
test had already read those runs), the wider model's low run and the base-width pair bounded $R$ above by
\qStdcfgSevenSixEightSizeRUpperShort{} whenever the high run moves the row no more than the low run, so the criterion
could be met. Both pairs leave the interval, and
$R=\qStdcfgSevenSixEightSizeR$, with the base width at peak rate \qStdcfgSevenSixEightSizeLrBase{} and the wider model
at \qStdcfgSevenSixEightSizeLrWide, each at the rate its own registered rule chose.\supported{Q-STDCFG768-SIZE}\label{res:w768-size}

\item \emph{Size comparison, softmax head}, the second instrument: each cell is judged alone and printed beside the main
cell. Seed~\qStdcfgSevenSixEightXOneSoftmaxSeedZero, both widths on one machine: both pairs leave the interval,
$R=\qStdcfgSevenSixEightSizeSmRSZero$ (peak rates \qStdcfgSevenSixEightSizeSmLrBaseSZero{} /
\qStdcfgSevenSixEightSizeSmLrWideSZero). Seed~\qStdcfgSevenSixEightXOneSoftmaxSeedOne, both widths on one machine: both
pairs leave the interval, $R=\qStdcfgSevenSixEightSizeSmRSOne$ (\qStdcfgSevenSixEightSizeSmLrBaseSOne{} /
\qStdcfgSevenSixEightSizeSmLrWideSOne). Seed~\qStdcfgSevenSixEightXOneSoftmaxSeedTwo, both widths on one machine: both
pairs leave the interval, $R=\qStdcfgSevenSixEightSizeSmRSTwo$ (\qStdcfgSevenSixEightSizeSmLrBaseSTwo{} /
\qStdcfgSevenSixEightSizeSmLrWideSTwo). Rates are base width first.%
\supported{Q-STDCFG768-SIZE-SOFTMAX}\label{res:w768-size-softmax}

\item \emph{Per-row $\eps$, main head} (Section~\ref{sec:decomp}). Seed~\qStdcfgSevenSixEightXEightSeedZero:
$r_{\mathrm{row}}=\qStdcfgSevenSixEightXEightRRowSZero$, at or above the upper threshold;
$r_{\mathrm{rest}}=\qStdcfgSevenSixEightXEightRRestSZero$, between the two thresholds.
Seed~\qStdcfgSevenSixEightXEightSeedOne: $r_{\mathrm{row}}=\qStdcfgSevenSixEightXEightRRowSOne$, at or above the upper
threshold; $r_{\mathrm{rest}}=\qStdcfgSevenSixEightXEightRRestSOne$, between the two thresholds.
Seed~\qStdcfgSevenSixEightXEightSeedTwo{} has no coordinate: a second host restart cut off its run with the low $\eps$
everywhere, the endpoint both coordinates need.\supported{Q-STDCFG768-X8}\label{res:w768-rowe}

\item \emph{Per-row learning rate, main head.} Seed~\qStdcfgSevenSixEightXEightSeedZero:
$s_{\mathrm{lr}}=\qStdcfgSevenSixEightXNineSlrSZero$, non-negative and below the near-exact band, on the side
registered as the prediction before any wider-model run (copied from the base width's registration, whose own reading
of this arm was known by then). Seed~\qStdcfgSevenSixEightXEightSeedOne:
$s_{\mathrm{lr}}=\qStdcfgSevenSixEightXNineSlrSOne$, non-negative and below the near-exact band, on the same registered
side. Seed~\qStdcfgSevenSixEightXEightSeedTwo{} has no value: its arm, paired with the run that did not finish, never
started.\supported{Q-STDCFG768-X9}\label{res:w768-rowlr}
\end{enumerate}

Two boundaries go with these readings. Every run of the two kill tests' own pairs crosses its loss midpoint between
steps \qStdcfgSevenSixEightXOneEscLo{} and \qStdcfgSevenSixEightXOneEscHi{} with the main head and
\qStdcfgSevenSixEightXOneSoftmaxEscLo{} and \qStdcfgSevenSixEightXOneSoftmaxEscHi{} with the softmax head; the rate rule
screened out a stall followed by a late drop, not this crossing, and the steps, listed per pair in
Appendix~\ref{app:w768}, select nothing. The size criterion concerns one step in size, in which width, depth and heads
change together and each width trains at its own selected rate, so it does not separate size from learning rate. It was
registered after the base-width kill test had been read, and the registration declares what that means: the base-width
values of the softmax cells, and that of the main head's seed-\qStdcfgSevenSixEightXOneSeedZero{} pair on another
machine, were known then, so a cell's verdict turned on where one wider-model value fell (the main cell's own base-width
pair was trained afterwards, on the wider model's machine). A cell that meets the criterion may show only that the wider
model, at its own rate, stays in the same regime as the base width, not that the effect does not change with size.

\begin{protocolbox}
\textbf{Width \qStdcfgSevenSixEightDModel} (\path{PREREG_STDCFG_768.md}, read per seed or per cell, never pooled with
the base width; rows in \path{scripts/claims.jsonl}). For each test named in Section~\ref{sec:w768}: its full registered outcome set (\S6.10; each set as
its row records it; the re-run's is A9.1's, with NOT-RUN by the letter of \S6.10) and the outcome it reached; seeds in order
\qStdcfgSevenSixEightXOneSeedZero, \qStdcfgSevenSixEightXOneSeedOne, \qStdcfgSevenSixEightXOneSeedTwo.
\emph{LR-768}, the rate rule run again at this width (\S1.3, \S1.5; \texttt{Q-STDCFG768-LR}): set
\{\qStdcfgSevenSixEightLrSetSelected, \qStdcfgSevenSixEightLrSetNone, \qStdcfgSevenSixEightLrSetNotCovered,
\qStdcfgSevenSixEightLrSetMachineLost\}; outcome \textbf{\qStdcfgSevenSixEightLr} (\qStdcfgSevenSixEightLrArgv).
\emph{X1-768}, main head (\S2.4, \S2.5; \texttt{Q-STDCFG768-X1}): set \{\qStdcfgSevenSixEightXOneSetGo,
\qStdcfgSevenSixEightXOneSetNoGo, \qStdcfgSevenSixEightXOneSetNotCovered, \qStdcfgSevenSixEightXOneSetNotRun\};
outcome \textbf{\qStdcfgSevenSixEightXOne} (cells in Appendix~\ref{app:w768}).
\emph{X1-768-S}, softmax head, beside X1-768 and never part of it (\S2.4; \texttt{Q-STDCFG768-X1-SOFTMAX}): set
\{\qStdcfgSevenSixEightXOneSoftmaxSetGo, \qStdcfgSevenSixEightXOneSoftmaxSetNoGo,
\qStdcfgSevenSixEightXOneSoftmaxSetNotCovered, \qStdcfgSevenSixEightXOneSoftmaxSetNotRun\}; outcome
\textbf{\qStdcfgSevenSixEightXOneSoftmax} (cells in Appendix~\ref{app:w768}).
\emph{Re-run of X1-768 seeds \qStdcfgSevenSixEightXOneSeedOne{} and \qStdcfgSevenSixEightXOneSeedTwo} (machine
\qStdcfgSevenSixEightReopenMachine, host \qStdcfgSevenSixEightReopenHost, token \qStdcfgSevenSixEightReopenToken), side
by side only, never in \qStdcfgSevenSixEightXOneSetGo, \qStdcfgSevenSixEightXOneSoftmaxSetGo{} or SIZE-768 (A9.1; cells
in Appendix~\ref{app:w768}): set \{\qStdcfgSevenSixEightXOneSetGo, \qStdcfgSevenSixEightXOneSetNoGo,
\qStdcfgSevenSixEightXOneSetNotCovered\} (A9.1), with \qStdcfgSevenSixEightXOneSetNotRun{} by the letter of \S6.10 only;
outcome \textbf{\qStdcfgSevenSixEightReopenOutcome}; as registered, its low runs were the source of the factor and
identity check of X8-768 and X9-768 on seeds \qStdcfgSevenSixEightXOneSeedOne{} and \qStdcfgSevenSixEightXOneSeedTwo{}
(on seed \qStdcfgSevenSixEightXOneSeedTwo{} neither was used).
\emph{SIZE-768}, per judged cell (\S3.2, \S3.5; \texttt{Q-STDCFG768-SIZE}, \texttt{-SIZE-SOFTMAX}): set
\{\qStdcfgSevenSixEightSizeSetSame, \qStdcfgSevenSixEightSizeSetChanges, \qStdcfgSevenSixEightSizeSetNotCovered,
\qStdcfgSevenSixEightSizeSetNotRun\}; main head, seed \qStdcfgSevenSixEightXOneSeedZero: outcome
\textbf{\qStdcfgSevenSixEightSize}. Softmax cells, second instrument, each alone and beside the main cell: set
\{\qStdcfgSevenSixEightSizeSmSetSame, \qStdcfgSevenSixEightSizeSmSetChanges, \qStdcfgSevenSixEightSizeSmSetNotCovered,
\qStdcfgSevenSixEightSizeSmSetNotRun\}; outcomes in seed order \textbf{\qStdcfgSevenSixEightSizeSoftmaxZero},
\textbf{\qStdcfgSevenSixEightSizeSoftmaxOne}, \textbf{\qStdcfgSevenSixEightSizeSoftmaxTwo}.
\emph{X8-768}, main head, per seed, $r_{\mathrm{row}}$\,/\,$r_{\mathrm{rest}}$ (\S4.2; \texttt{Q-STDCFG768-X8}): set
\{\qStdcfgSevenSixEightXEightSetA, \qStdcfgSevenSixEightXEightSetB, \qStdcfgSevenSixEightXEightSetC,
\qStdcfgSevenSixEightXEightSetOut, \qStdcfgSevenSixEightXEightSetUndefined, \qStdcfgSevenSixEightXEightSetVoid,
\qStdcfgSevenSixEightXEightSetMissing, \qStdcfgSevenSixEightXEightSetNotCovered, \qStdcfgSevenSixEightXEightSetNotRun\};
outcomes in seed order \textbf{\qStdcfgSevenSixEightXEightSZeroRRow}\,/\,\textbf{\qStdcfgSevenSixEightXEightSZeroRRest},
\textbf{\qStdcfgSevenSixEightXEightSOneRRow}\,/\,\textbf{\qStdcfgSevenSixEightXEightSOneRRest},
\textbf{\qStdcfgSevenSixEightXEightSTwoRRow}\,/\,\textbf{\qStdcfgSevenSixEightXEightSTwoRRest}.
\emph{X9-768}, main head, per seed; registered prediction \qStdcfgSevenSixEightXNineSetSameSide{} on every seed (\S5.4;
\texttt{Q-STDCFG768-X9}): set \{\qStdcfgSevenSixEightXNineSetSameSide, \qStdcfgSevenSixEightXNineSetNeutral,
\qStdcfgSevenSixEightXNineSetOpposite, \qStdcfgSevenSixEightXNineSetOut, \qStdcfgSevenSixEightXNineSetVoid,
\qStdcfgSevenSixEightXNineSetMissing, \qStdcfgSevenSixEightXNineSetNotCovered, \qStdcfgSevenSixEightXNineSetNotRun\};
outcomes in seed order \textbf{\qStdcfgSevenSixEightXNineZero}, \textbf{\qStdcfgSevenSixEightXNineOne},
\textbf{\qStdcfgSevenSixEightXNineTwo}.
\end{protocolbox}

\section{Related work}
\label{sec:related}

That the constant is a lever on the step size is established \citep{choi2019empirical,wortsman2024smallscale,
everett2024scaling}, and that rare tokens fare differently under Adam is measured \citep{kunstner2024heavy,
stollenwerk2025coupled,zhao2024deconstructing}; none of these measures a per-row displacement or has a counterfactual.
Read through stochastic differential equations, the constant moves with the batch size, to $\eps/\sqrt{\kappa}$ when the
batch size $B$ becomes $\kappa B$ \citep{malladi2022sdes}, and a study that holds it fixed finds that the best optimizer
for language-model pretraining changes with the batch size \citep{dang2026bestoptimizer}; the standard configuration, at
both widths, trains at a single batch size (Section~\ref{sec:setup}), and the constant's effect is read at that batch
size only. Recomputing a recorded Adam update at another epsilon from the run's own moment estimates has been done
before, at epsilon zero, as a within-run diagnostic never compared with a real run at that epsilon
\citep{molybog2023theory}; here that re-division is the prediction under test. Replaying a recorded trajectory for a
what-if is trajectory-based data attribution \citep{hara2019sgdinfluence}, whose error has been measured and broken into
its sources \citep{deng2026faithful}; there the perturbation is data and the counterfactual a first-order linearisation,
so the miss reported here cannot be charged to a Taylor error. Machine unlearning for online L-BFGS has been formulated
with, as its target, the optimizer state that would have arisen had the deleted samples never been processed
\citep{stewart2026form}. Truncating a trajectory in time biases hyperparameter conclusions \citep{wu2018shorthorizon},
and a frozen Adam preconditioner is valid only over a short horizon \citep{cohen2022adaptive}; post-hoc averaging,
reversible learning and grafting reconstruct or share optimizer states without a prediction checked against a run
\citep{sanyal2023early,karras2024analyzing,maclaurin2015gradient,agarwal2020disentangling}. Readings of a per-row step
\citep{balles2018dissecting,chatterjee2020coherent} are not tested here, and a sweep of the constant that left
pretraining loss nearly unchanged \citep{semenov2025benchmarking} reads a corpus-level loss that cannot single out one
row. Appendix~\ref{app:related} gives each comparison in full.

\section{Limitations}
\label{sec:limits}

One probe row, one model family and one corpus. At the base width, three seeds per registered population of the kill
test and the per-row tests, two more registered after the result had been committed and read on their own; the panel of
rows is one seed. The kill test and the per-row tests are read per seed, and we claim no rate. The kill test does not
separate the standard configuration from this learning rate, warmup and budget; a check at another number of tokens per
step was named before the runs and has not been done, and since the best optimizer has been found to change with the
batch size \citep{dang2026bestoptimizer}, no reading here is carried to another batch size. The crossings of the base
width's kill test (Section~\ref{sec:setup}) are printed beside their cells and never used to select a run. The
interpolation coordinates of Section~\ref{sec:decomp} are not shares of an effect, and most shadow cells fall outside
the registered range, so the decomposition says which per-row change reproduces the row's displacement, not why. What
each test at the second width reached, with its full set of registered outcomes, is given only in
Section~\ref{sec:w768}. Which part of the trajectory the replay misses is not settled: the shadow, which asked whether
the row's own position closes the gap, fell outside its range whenever every parameter's $\eps$ changed; of the two
routes through the rest of the network that were registered and read, the gradient's size was rejected, by a ruling made
after its read, and the gradient's coherence is not reported here. The corrections ruled into the record while this
version was prepared are listed in Appendix~\ref{app:withdrawn}; none of them changes a number printed here.

\section{Conclusion and interpretation}
\label{sec:conclusion}

In the standard configuration of Section~\ref{sec:setup} (pre-LN with a final norm, a \qStdcfgWarmup-step warmup, a
cosine schedule, and a peak learning rate chosen by the registered rule on the $\eps=\qStdcfgXOneEpsLo$ runs alone),
holding a run's gradients fixed and changing only epsilon by four orders of magnitude predicted more suppression of
the probe row than the runs showed: the row kept \qStdcfgXOneBceFourRatioLo--\qStdcfgXOneBceFourRatioHi{} times more of
its displacement than predicted with the main head and \qStdcfgXOneSoftmaxFourRatioLo--\qStdcfgXOneSoftmaxFourRatioHi{}
times more with the softmax head, on each of seeds \qStdcfgXOneSeedLo--\qStdcfgXOneSeedHi. Every one of these runs still
falls through the midpoint of its loss curve (main head between steps \qStdcfgXOneEscBceLo{} and \qStdcfgXOneEscBceHi,
softmax head between \qStdcfgXOneEscSoftmaxLo{} and \qStdcfgXOneEscSoftmaxHi, the runs of a pair not always in the same
window); the steps are printed and never used to select a run, and the learning-rate rule screened out a stall above the
midpoint before a late drop, not that fall. In separate real runs, raising the row's own epsilon alone is enough to
reproduce its change, raising every other parameter's epsilon alone also moves it part of the way, and a matched
learning-rate change is mispredicted by the replay in the same direction.
\interpretation{We read the two misses together as a reason to doubt, for this row, any per-row step-size argument that
holds the trajectory fixed; that the miss keeps its sign with the denominator untouched suggests it is not specific to
epsilon, though one learning-rate factor per seed was tested and no other row.}\label{endofmaintext}

\section*{Impact Statement}
This paper presents work whose goal is to advance the field of Machine Learning. There are many potential societal
consequences of our work, none which we feel must be specifically highlighted here.

\bibliography{refs}
\bibliographystyle{icml2026}

\newpage
\appendix
\onecolumn
\raggedbottom

\section{Registrations, criteria and reading rules}
\label{app:reg}
For every row whose values this paper prints, the tables give the row's own record of its registration: the field
\texttt{prior\_registration} of \path{scripts/claims.jsonl}, split at the read commit, the commit whose tree a read was
taken from ((b) in the width-768 registration), chosen as each registration defines it. P marks the commit at which a
read's reader was pinned and T the commit that transcribed the auditors' structural audit of a read; the audit file is
that structural audit, and its \S16 is printed in Appendix~\ref{app:w768-audit}. The tables are generated from the
registry by \path{scripts/v40_records.py}. The field is printed as recorded, except that the auditors are named as in the
rest of this paper and a note on how one ruling reached the authors is cut.

\begin{small}
\medskip\noindent\begin{minipage}{\linewidth}\textbf{\texttt{PREREG\_STDCFG.md}}\\[2pt]
\begin{tabular}{P{0.27\linewidth}P{0.31\linewidth}P{0.14\linewidth}P{0.18\linewidth}} %
\toprule
row & sections, amendments and rulings & read commit & recorded after the read commit \\
\midrule
\texttt{Q-STDCFG-LR} & \S{}1.2--1.3 with AMENDMENT-1 A1.1 (three seeds), AMENDMENT-9 (rerun on V3, V1 VOID-MACHINE-LOST), AMENDMENT-11 (the per-machine reader) & \texttt{1095bf98d13f} & frozen by AMENDMENT-12 \\
\texttt{Q-STDCFG-X1} & \S{}2 with AMENDMENT-4 A4.1 and every amendment and ruling in it & \texttt{6e5afbbf424a} & unlocked by RULING-5 (the auditors' structural audit) \\
\texttt{Q-STDCFG-X1-SOFTMAX} & \S{}2 with AMENDMENT-4 A4.1 and every amendment and ruling in it & \texttt{6e5afbbf424a} & unlocked by RULING-5 (the auditors' structural audit) \\
\texttt{Q-STDCFG-X1EXT} & AMENDMENT-23 (registered after X1's read, A23.2) with \S{}2.2--2.5 and AMENDMENT-25 (the reader) & \texttt{740cb06fd685} & -- \\
\texttt{Q-STDCFG-X6} & \S{}3 with AMENDMENT-4 A4.1 (softmax one dose) and A6.4 item 4 & \texttt{6e5afbbf424a} & unlocked by RULING-6 item 5 after the auditors' X6 audit (audit file \S{}4) \\
\texttt{Q-STDCFG-X7} & AMENDMENT-27 (27.1--27.4) with AMENDMENT-29 (the reader and its supplementary registrations) & \texttt{eaae13fa4400} & audited after the read (A33 item 3, RULING-7 item 2) \\
\texttt{Q-STDCFG-X8} & \S{}4 (PREREG\_ROWEPS \S{}1.2 verbatim, the population changed) with RULING-1's bridge and AMENDMENT-34 (the read commit) & \texttt{3a0fde1b62b7} & audited before the read by the auditors (audit file \S{}5, RULING-7 item 6) \\
\texttt{Q-STDCFG-X8EXT} & \S{}4.2 / \S{}4.3 (PREREG\_ROWEPS \S{}1.2 verbatim) on AMENDMENT-35 35.2 / -37 / -38's population with RULING-1's bridge (AMENDMENT-39 / -41, A40.4); read by AMENDMENT-42's definition with AMENDMENT-44's pinned reader & \texttt{39a4ff48a551} & (= T, the auditors' audit transcription, audit file \S{}11); A40.1 / A42.8 declare the read commit after the data cut-off \\
\texttt{Q-STDCFG-X8SM} & \S{}4.2 / \S{}4.3 (PREREG\_ROWEPS \S{}1.2 verbatim) on AMENDMENT-35 35.1's population; units per seed (A40.5); read by AMENDMENT-43's reader set & \texttt{2f79221334d1} & that the reader derived from the harvest log and git (43.6); audited before the read by the auditors (audit file \S{}12) \\
\texttt{Q-STDCFG-X9} & \S{}5 with AMENDMENT-4 A4.4 (c\_s from the X1 bce15 lo), RULING-1's bridge, AMENDMENT-21 (c\_s frozen), AMENDMENT-26 / -28 / -30 (the c-arms) and AMENDMENT-34 (the read commit) & \texttt{4efa4942067d} & audited before the read (audit file \S{}6, RULING-7 item 6) \\
\texttt{Q-STDCFG-X9EXT} & \S{}5.2 / \S{}5.3 on AMENDMENT-37 / -39 / -41's population (A40.4: the c-arm lines of the population constant) with c\_s frozen by AMENDMENT-38; read by AMENDMENT-42's definition with AMENDMENT-44's pinned reader & \texttt{39a4ff48a551} & (= T, audit file \S{}11); A40.1 / A42.8 declare the read commit after the data cut-off \\
\bottomrule
\end{tabular}\end{minipage}

\medskip\noindent\begin{minipage}{\linewidth}\textbf{\texttt{PREREG\_STDCFG\_768.md}}\\[2pt]
\begin{tabular}{P{0.27\linewidth}P{0.31\linewidth}P{0.14\linewidth}P{0.18\linewidth}} %
\toprule
row & sections, amendments and rulings & read commit & recorded after the read commit \\
\midrule
\texttt{Q-STDCFG768-LR} & \S{}1 (\S{}1.3--\S{}1.6) with AMENDMENT-1 (the reader pinned at P fd7d8d79) & \texttt{32de1f1fb73d} & frozen by AMENDMENT-3 A3.2 after the read; T not required (\S{}9 item 7) \\
\texttt{Q-STDCFG768-X1} & \S{}2 (\S{}2.2--\S{}2.5) with AMENDMENT-3 (the frozen lr) and AMENDMENT-4 (the reader pinned at P dccd7c1d) & \texttt{8045e0b45544} & T 0ba280ba \\
\texttt{Q-STDCFG768-X1-SOFTMAX} & \S{}2 (\S{}2.2--\S{}2.5) with AMENDMENT-3 (the frozen lr) and AMENDMENT-4 (the reader pinned at P dccd7c1d) & \texttt{283597f0ed46} & T 0ba280ba \\
\texttt{Q-STDCFG768-SIZE} & \S{}3 (\S{}3.2--\S{}3.8) with AMENDMENT-6 (the stage-3 reader pinned at P 1bbf751b) & \texttt{cfc1942a03b2} & T absent at (b): the structural audit \S{}16.3 (1ec180422c84) is late and printed beside \\
\texttt{Q-STDCFG768-SIZE-SOFTMAX} & \S{}3 (\S{}3.2--\S{}3.8) with AMENDMENT-6 (the stage-3 reader pinned at P 1bbf751b) & \texttt{cfc1942a03b2} & T absent at (b): the structural audit \S{}16.3 (1ec180422c84) is late and printed beside \\
\texttt{Q-STDCFG768-X8} & \S{}4 / \S{}5 with AMENDMENT-6 (the stage-3 reader pinned at P 1bbf751b), AMENDMENT-10 (x8-rest reopened), AMENDMENT-11--14 (c\_s per seed), AMENDMENT-16 / -17 / -21 (the bridges) and AMENDMENT-15 A15.2 (the T correction) & \texttt{333e480c3d65} & T misattributed by the driver, the structural audit \S{}16.2 (1ec180422c84) late \\
\texttt{Q-STDCFG768-X9} & \S{}4 / \S{}5 with AMENDMENT-6 (the stage-3 reader pinned at P 1bbf751b), AMENDMENT-10 (x8-rest reopened), AMENDMENT-11--14 (c\_s per seed), AMENDMENT-16 / -17 / -21 (the bridges) and AMENDMENT-15 A15.2 (the T correction) & \texttt{333e480c3d65} & T misattributed by the driver, the structural audit \S{}16.2 (1ec180422c84) late \\
\bottomrule
\end{tabular}\end{minipage}

\medskip\noindent\begin{minipage}{\linewidth}\textbf{\texttt{PREREG\_SHADOW.md}}\\[2pt]
\begin{tabular}{P{0.27\linewidth}P{0.31\linewidth}P{0.14\linewidth}P{0.18\linewidth}} %
\toprule
row & sections, amendments and rulings & read commit & recorded after the read commit \\
\midrule
\texttt{Q-SHADOW-X3P} & \S{}3--\S{}4 and its RULING-1 & \texttt{6e5afbbf424a} & (= X1's read commit, PREREG\_STDCFG AMENDMENT-36); audited before the read by the auditors (audit file \S{}7); the READ's wording by RULING-8 item 1, RULING-9 item 1 and RULING-13 (2026-09-29; its paragraph on the three READ drafts, NOTES \S{}A items 1--10 and \S{}B items 1, 2, 4, 5, 6, 9) \\
\texttt{Q-SHADOW-X3PX9} & \S{}3 (the X9 shadows) and RULING-1 & \texttt{4efa4942067d} & (= X9's read commit, PREREG\_STDCFG AMENDMENT-36); audited before the read (audit file \S{}7); the READ's wording by RULING-8 item 1, RULING-9 item 1 and RULING-13 (2026-09-29; its paragraph on the three READ drafts, NOTES \S{}A items 1--10 and \S{}B items 1, 2, 4, 5, 6, 9) \\
\texttt{Q-SHADOW-X3PROW} & AMENDMENT-1 (A1.1--A1.7) and AMENDMENT-2 (the reader and its read) & \texttt{70616fe67386} & (the commit that transcribed the auditors' structural audit, audit file \S{}8; RULING-9 item 6); the READ's wording by RULING-13 (2026-09-29; its paragraph on the three READ drafts, NOTES \S{}A items 1--10 and \S{}B items 1, 2, 4, 5, 6, 9) \\
\bottomrule
\end{tabular}\end{minipage}

\medskip\noindent\begin{minipage}{\linewidth}\textbf{\texttt{PREREG\_X11.md}}\\[2pt]
\begin{tabular}{P{0.27\linewidth}P{0.31\linewidth}P{0.14\linewidth}P{0.18\linewidth}} %
\toprule
row & sections, amendments and rulings & read commit & recorded after the read commit \\
\midrule
\texttt{Q-X11} & \S{}1--\S{}9 and AMENDMENT-1 (the reader and its read) & \texttt{70616fe67386} & (the commit that transcribed the auditors' structural audit, audit file \S{}9; RULING-9 item 6); the READ's wording by RULING-13 (2026-09-29; its paragraph on the three READ drafts, NOTES \S{}A items 1--10 and \S{}B items 1, 2, 4, 5, 6, 9) \\
\bottomrule
\end{tabular}\end{minipage}
\end{small}

\section{The kill test, cell by cell}
\label{app:x1}

Every cell of the kill test (Section~\ref{sec:x1}), per head, seed and dose. $c=-\ln\rho_{\mathrm{cf}}$ is fixed by the
low-$\eps$ run alone and was printed before the high run was read; whenever the high run moves the row no more than
the low run, a cell can fire below the interval only if $c>\ln\qStdcfgXOneKillHi\approx\qStdcfgXOneLowSideCeil$ nats (a
cell that fires low without that headroom contradicts the ceiling and makes its seed not covered), and every cell, at
three decades as at four, cleared it, so both halves of the pass condition could have gone either way. The kill ratio is $\rho_{\mathrm{cf}}/\rho_{\mathrm{meas}}$,
$e^{F}$ its inverse, and $s$ the realised fraction (Section~\ref{sec:setup}); a cell fires when the kill ratio lies
outside $[\qStdcfgXOneKillLo,\qStdcfgXOneKillHi]$. The crossing steps are the steps at which each run of the pair
first crosses the midpoint of its loss curve, read at a resolution of \qStdcfgXOneEscWindow{} steps; they are printed,
never used to select a cell. Ratios are read at step \qStdcfgXOneLastCommonStep{}, the last step both runs report; the
accumulators stop adding after step \qStdcfgXOneReadEndStep. The first block is the main test (seeds
\qStdcfgXOneSeedLo--\qStdcfgXOneSeedHi), the second the softmax head beside it, the third the two seeds registered
afterwards and read on their own; no block is pooled with another.

\paragraph{Two declared deviations.} Before the formal read a dry run showed the auditors every cell word and both outcome words, the $F\le c$ flags and
sides, the reachability labels, the crossing steps and the certificate's count of differing entries; every
floating-point value (kill ratio, $e^{F}$, $\rho$, $c$, $F$, $s$) was masked. The outcome rule had been fixed before,
and the read is mechanical. Seeds \qStdcfgXOneExtSeedLo{} and \qStdcfgXOneExtSeedHi{} were read before
their structural audit, which is therefore a post-read audit.

\begin{table}[h]
\centering\small
\begin{tabular}{lllllllll}
\toprule
head & seed & pair & $c$ (nats) & kill ratio & $e^{F}$ & $s$ & crossing step (low / high run) & cell word \\
\midrule
main & 0 & \qStdcfgXOneDoseFour & $\qStdcfgXOneCellCeilFourSZero$ & $\qStdcfgXOneCellKillFourSZero$ & $\qStdcfgXOneCellRatioFourSZero$ & $\qStdcfgXOneCellFracFourSZero$ & \qStdcfgXOneCellEscLoFourSZero / \qStdcfgXOneCellEscHiFourSZero & \qStdcfgXOneWordFourSZero \\
main & 0 & \qStdcfgXOneDoseThree & $\qStdcfgXOneCellCeilThreeSZero$ & $\qStdcfgXOneCellKillThreeSZero$ & $\qStdcfgXOneCellRatioThreeSZero$ & $\qStdcfgXOneCellFracThreeSZero$ & \qStdcfgXOneCellEscLoThreeSZero / \qStdcfgXOneCellEscHiThreeSZero & \qStdcfgXOneWordThreeSZero \\
main & 1 & \qStdcfgXOneDoseFour & $\qStdcfgXOneCellCeilFourSOne$ & $\qStdcfgXOneCellKillFourSOne$ & $\qStdcfgXOneCellRatioFourSOne$ & $\qStdcfgXOneCellFracFourSOne$ & \qStdcfgXOneCellEscLoFourSOne / \qStdcfgXOneCellEscHiFourSOne & \qStdcfgXOneWordFourSOne \\
main & 1 & \qStdcfgXOneDoseThree & $\qStdcfgXOneCellCeilThreeSOne$ & $\qStdcfgXOneCellKillThreeSOne$ & $\qStdcfgXOneCellRatioThreeSOne$ & $\qStdcfgXOneCellFracThreeSOne$ & \qStdcfgXOneCellEscLoThreeSOne / \qStdcfgXOneCellEscHiThreeSOne & \qStdcfgXOneWordThreeSOne \\
main & 2 & \qStdcfgXOneDoseFour & $\qStdcfgXOneCellCeilFourSTwo$ & $\qStdcfgXOneCellKillFourSTwo$ & $\qStdcfgXOneCellRatioFourSTwo$ & $\qStdcfgXOneCellFracFourSTwo$ & \qStdcfgXOneCellEscLoFourSTwo / \qStdcfgXOneCellEscHiFourSTwo & \qStdcfgXOneWordFourSTwo \\
main & 2 & \qStdcfgXOneDoseThree & $\qStdcfgXOneCellCeilThreeSTwo$ & $\qStdcfgXOneCellKillThreeSTwo$ & $\qStdcfgXOneCellRatioThreeSTwo$ & $\qStdcfgXOneCellFracThreeSTwo$ & \qStdcfgXOneCellEscLoThreeSTwo / \qStdcfgXOneCellEscHiThreeSTwo & \qStdcfgXOneWordThreeSTwo \\
\midrule
softmax & 0 & \qStdcfgXOneDoseFour & $\qStdcfgXOneSoftmaxCellCeilFourSZero$ & $\qStdcfgXOneSoftmaxCellKillFourSZero$ & $\qStdcfgXOneSoftmaxCellRatioFourSZero$ & $\qStdcfgXOneSoftmaxCellFracFourSZero$ & \qStdcfgXOneSoftmaxCellEscLoFourSZero / \qStdcfgXOneSoftmaxCellEscHiFourSZero & \qStdcfgXOneSoftmaxWordFourSZero \\
softmax & 1 & \qStdcfgXOneDoseFour & $\qStdcfgXOneSoftmaxCellCeilFourSOne$ & $\qStdcfgXOneSoftmaxCellKillFourSOne$ & $\qStdcfgXOneSoftmaxCellRatioFourSOne$ & $\qStdcfgXOneSoftmaxCellFracFourSOne$ & \qStdcfgXOneSoftmaxCellEscLoFourSOne / \qStdcfgXOneSoftmaxCellEscHiFourSOne & \qStdcfgXOneSoftmaxWordFourSOne \\
softmax & 2 & \qStdcfgXOneDoseFour & $\qStdcfgXOneSoftmaxCellCeilFourSTwo$ & $\qStdcfgXOneSoftmaxCellKillFourSTwo$ & $\qStdcfgXOneSoftmaxCellRatioFourSTwo$ & $\qStdcfgXOneSoftmaxCellFracFourSTwo$ & \qStdcfgXOneSoftmaxCellEscLoFourSTwo / \qStdcfgXOneSoftmaxCellEscHiFourSTwo & \qStdcfgXOneSoftmaxWordFourSTwo \\
\midrule
main & 3 & \qStdcfgXOneDoseFour & $\qStdcfgXOneExtCellCeilBceFourSThree$ & $\qStdcfgXOneExtCellKillBceFourSThree$ & $\qStdcfgXOneExtCellRatioBceFourSThree$ & $\qStdcfgXOneExtCellFracBceFourSThree$ & \qStdcfgXOneExtCellEscLoBceFourSThree / \qStdcfgXOneExtCellEscHiBceFourSThree & \qStdcfgXOneExtWordBceFourSThree \\
main & 3 & \qStdcfgXOneDoseThree & $\qStdcfgXOneExtCellCeilBceThreeSThree$ & $\qStdcfgXOneExtCellKillBceThreeSThree$ & $\qStdcfgXOneExtCellRatioBceThreeSThree$ & $\qStdcfgXOneExtCellFracBceThreeSThree$ & \qStdcfgXOneExtCellEscLoBceThreeSThree / \qStdcfgXOneExtCellEscHiBceThreeSThree & \qStdcfgXOneExtWordBceThreeSThree \\
softmax & 3 & \qStdcfgXOneDoseFour & $\qStdcfgXOneExtCellCeilSoftmaxFourSThree$ & $\qStdcfgXOneExtCellKillSoftmaxFourSThree$ & $\qStdcfgXOneExtCellRatioSoftmaxFourSThree$ & $\qStdcfgXOneExtCellFracSoftmaxFourSThree$ & \qStdcfgXOneExtCellEscLoSoftmaxFourSThree / \qStdcfgXOneExtCellEscHiSoftmaxFourSThree & \qStdcfgXOneExtWordSoftmaxFourSThree \\
main & 4 & \qStdcfgXOneDoseFour & $\qStdcfgXOneExtCellCeilBceFourSFour$ & $\qStdcfgXOneExtCellKillBceFourSFour$ & $\qStdcfgXOneExtCellRatioBceFourSFour$ & $\qStdcfgXOneExtCellFracBceFourSFour$ & \qStdcfgXOneExtCellEscLoBceFourSFour / \qStdcfgXOneExtCellEscHiBceFourSFour & \qStdcfgXOneExtWordBceFourSFour \\
main & 4 & \qStdcfgXOneDoseThree & $\qStdcfgXOneExtCellCeilBceThreeSFour$ & $\qStdcfgXOneExtCellKillBceThreeSFour$ & $\qStdcfgXOneExtCellRatioBceThreeSFour$ & $\qStdcfgXOneExtCellFracBceThreeSFour$ & \qStdcfgXOneExtCellEscLoBceThreeSFour / \qStdcfgXOneExtCellEscHiBceThreeSFour & \qStdcfgXOneExtWordBceThreeSFour \\
softmax & 4 & \qStdcfgXOneDoseFour & $\qStdcfgXOneExtCellCeilSoftmaxFourSFour$ & $\qStdcfgXOneExtCellKillSoftmaxFourSFour$ & $\qStdcfgXOneExtCellRatioSoftmaxFourSFour$ & $\qStdcfgXOneExtCellFracSoftmaxFourSFour$ & \qStdcfgXOneExtCellEscLoSoftmaxFourSFour / \qStdcfgXOneExtCellEscHiSoftmaxFourSFour & \qStdcfgXOneExtWordSoftmaxFourSFour \\
\bottomrule
\end{tabular}
\caption{The kill test, every cell. Rows: \texttt{Q-STDCFG-X1}, \texttt{Q-STDCFG-X1-SOFTMAX}, \texttt{Q-STDCFG-X1EXT}.}
\label{tab:app-x1}
\end{table}

\section{Determinism certificates}
\label{app:cert}

At width \qStdcfgDModel{} one run of the main head's kill test (seed \qStdcfgXOneSeedLo, $\eps=\qStdcfgXOneEpsLo$)
was repeated on the same machine with the same training command, minus the side computations its twin carries for
another test. Of \qStdcfgCertLeaves{} compared entries, \qStdcfgCertUndeclared{} differed outside the declared
differences. A single repeat says only that this repeat did not flip; it is not a rate and is not used as a floor for
anything. The softmax head's kill-test runs ran on another machine, where no such repeat was made; their certificate is
that machine's instrument gates, and a same-command softmax pair on the first machine is printed beside it in the read,
not claimed for these runs.

\paragraph{The side computations.} The shadow of Section~\ref{sec:decomp} rides on the kill test's runs. Before the
kill test, each machine it used passed a gate in which a shorter run with the shadow on was compared with the same run
with it off: entry by entry for the main head, by the digests of the parameters and the optimizer state for the softmax
head. At full length the one check is the repeat above, whose twin carries the shadow and which does not. No other run
has a check of its own; the high-$\eps$ runs carry the shadow at other settings than their low-$\eps$ partners.

At width \qStdcfgSevenSixEightDModel{} the certificate and the identity checks are reported with the reads they
belong to, each beside its full set of registered outcomes (Appendix~\ref{app:w768}).

\section{Decomposition and its controls: the per-seed record}
\label{app:decomp}

This appendix carries every value behind Section~\ref{sec:decomp} and two controls of the kill test. Each table block
is one population; populations are set side by side, never pooled, and no statistic spans two of them. Seed numbers are
labels.

\begin{protocolbox}
\textbf{Registered words per seed or cell} (populations side by side, never pooled; outcome sets in
Table~\ref{tab:app-decomp-sets}).
\emph{Per-row $\eps$}, $r_{\mathrm{row}}$\,/\,$r_{\mathrm{rest}}$ (\path{PREREG_STDCFG.md} \S4.2).
BCE, seeds 0, 1, 2 (Q-STDCFG-X8): \qStdcfgXEightRowSeedZero\,/\,\qStdcfgXEightRestSeedZero;
\qStdcfgXEightRowSeedOne\,/\,\qStdcfgXEightRestSeedOne; \qStdcfgXEightRowSeedTwo\,/\,\qStdcfgXEightRestSeedTwo.
BCE, seeds 3, 4, read on their own (AMENDMENT-35/-37/-38; Q-STDCFG-X8EXT):
\qStdcfgXEightExtRowSeedThree\,/\,\qStdcfgXEightExtRestSeedThree; \qStdcfgXEightExtRowSeedFour\,/\,\qStdcfgXEightExtRestSeedFour.
Softmax, seeds 0, 1, 2, read on its own (AMENDMENT-35; Q-STDCFG-X8SM):
\qStdcfgXEightSoftmaxRowSeedZero\,/\,\qStdcfgXEightSoftmaxRestSeedZero;
\qStdcfgXEightSoftmaxRowSeedOne\,/\,\qStdcfgXEightSoftmaxRestSeedOne;
\qStdcfgXEightSoftmaxRowSeedTwo\,/\,\qStdcfgXEightSoftmaxRestSeedTwo.\\
\emph{Per-row learning rate} (\S5.3; predicted branch \qStdcfgXNineSetSameSide): BCE, seeds 0, 1, 2 (Q-STDCFG-X9):
\qStdcfgXNineSeedZero, \qStdcfgXNineSeedOne, \qStdcfgXNineSeedTwo; seeds 3, 4, read on their own (Q-STDCFG-X9EXT):
\qStdcfgXNineExtSeedThree, \qStdcfgXNineExtSeedFour.\\
\emph{Shadow}, forward\,/\,reverse (\path{PREREG_SHADOW.md} \S3). Learning-rate arm (Q-SHADOW-X3PX9): seed 0
\qShadowXThreePXNineForwardSeedZero\,/\,\qShadowXThreePXNineReverseSeedZero; seed 1
\qShadowXThreePXNineForwardSeedOne\,/\,\qShadowXThreePXNineReverseSeedOne; seed 2
\qShadowXThreePXNineForwardSeedTwo\,/\,\qShadowXThreePXNineReverseSeedTwo. Every parameter's $\eps$ changed, the seeds 0--2 pairs
of Section~\ref{sec:x1} (Q-SHADOW-X3P): BCE \qShadowXThreePSetOutOfRange{} in \qShadowXThreePBceCellsOut{} of
\qShadowXThreePBceCells{} cells; softmax \qShadowXThreePSetOutOfRange{} in \qShadowXThreePSoftmaxCellsOut{} of
\qShadowXThreePSoftmaxCells{} cells (Table~\ref{tab:app-x3p}). Same-world comparison, BCE seeds 0--2, descriptive
(AMENDMENT-1 A1.5; Q-SHADOW-X3PROW): seed 0 \qShadowXThreePRowForwardSeedZero\,/\,\qShadowXThreePRowReverseSeedZero;
seed 1 \qShadowXThreePRowForwardSeedOne\,/\,\qShadowXThreePRowReverseSeedOne; seed 2
\qShadowXThreePRowForwardSeedTwo\,/\,\qShadowXThreePRowReverseSeedTwo.\\
\emph{Gradient size} (\path{PREREG_X11.md} \S2--3; prediction \qXElevenPrediction; Q-X11): BCE, seeds 0, 1, 2
\qXElevenLSeedZero, \qXElevenLSeedOne, \qXElevenLSeedTwo; consequence of \qXElevenSetStronger{} ruled after the read
(RULING-13 \S A item 2).
\end{protocolbox}

\subsection{What was registered}
\label{app:decomp-reg}

\paragraph{Per-row $\eps$} (\path{PREREG_STDCFG.md} \S4; statistic and outcome set copied from \path{PREREG_ROWEPS.md}
\S1.2). The four arms of a seed run on one machine. The denominator $D_{\mathrm{lo}} - D_{\mathrm{hi}}$ is
printed before any coordinate; a denominator that is not positive leaves the seed's $r$ undefined. The sum
$r_{\mathrm{row}} + r_{\mathrm{rest}}$ is printed per seed and is not a criterion. The registration forbids reading the
denominator-direct branch as the replay being rescued, or citing either for the other; reading the
trajectory-mediated branch as ``$\eps$ has no effect''; calling $r$ a share or a percentage attribution; and merging the
seeds' $r$ or reporting only one of them. The seeds-3--4 population (AMENDMENT-35, -37, -38) and the softmax population
(AMENDMENT-35) were decided before the output of the first read of this test was committed, and their runs and commands
were registered after it; each is read on its own. For both, the reader was pinned only after runs of the population
had landed, a declared deviation (AMENDMENT-40).

\paragraph{Per-row learning rate} (\path{PREREG_STDCFG.md} \S5). $c_s = \lVert W_{\mathrm{cf}}[\eps_{\mathrm{hi}}]\rVert /
\lVert W_{\mathrm{cf}}[\eps_{\mathrm{lo}}]\rVert$ at the last accumulation step of the seed's low-$\eps$ run of the
four-decade pair, frozen before the arm ran (AMENDMENT-4 A4.4 and AMENDMENT-21; AMENDMENT-38 for seeds 3--4). With it the
replay predicts the same displacement ratio for the learning-rate arm and for the row-only $\eps$ arm. That low-$\eps$
run belongs to the kill test and ran on another machine than the arms; per seed, a whole-run comparison of it with the
arms' own low-$\eps$ run reads, for seeds 0, 1, 2, \qStdcfgXEightBridgeSeedZero{} (\qStdcfgXEightBridgeDiffersSeedZero{}
of \qStdcfgXEightBridgeComparedSeedZero{} entries differ), \qStdcfgXEightBridgeSeedOne{}
(\qStdcfgXEightBridgeDiffersSeedOne{} of \qStdcfgXEightBridgeComparedSeedOne) and \qStdcfgXEightBridgeSeedTwo{}
(\qStdcfgXEightBridgeDiffersSeedTwo{} of \qStdcfgXEightBridgeComparedSeedTwo), and for seeds 3 and 4
\qStdcfgXEightExtBridgeSeedThree{} (\qStdcfgXEightExtBridgeDiffersSeedThree{} of
\qStdcfgXEightExtBridgeComparedSeedThree) and \qStdcfgXEightExtBridgeSeedFour{} (\qStdcfgXEightExtBridgeDiffersSeedFour{}
of \qStdcfgXEightExtBridgeComparedSeedFour). The same comparison admits the forward shadows of the same-world
comparison below. Its reverse cells rest on a second comparison, of the arms' run with $\eps_{\mathrm{hi}}$ everywhere
against the kill test's high-$\eps$ run, registered to be printed and never a gate (PREREG\_SHADOW A1.4); for seeds 0, 1
and 2 it reads \qShadowXThreePRowHiLinkSeedZero{} (\qShadowXThreePRowHiLinkDiffersSeedZero{} of
\qShadowXThreePRowHiLinkComparedSeedZero{} entries differ), \qShadowXThreePRowHiLinkSeedOne{}
(\qShadowXThreePRowHiLinkDiffersSeedOne{} of \qShadowXThreePRowHiLinkComparedSeedOne) and
\qShadowXThreePRowHiLinkSeedTwo{} (\qShadowXThreePRowHiLinkDiffersSeedTwo{} of
\qShadowXThreePRowHiLinkComparedSeedTwo). Each is a check of these runs, not a statement that the two machines agree in
general. A drafted
branch on the side of a reverse replay was removed before any run (\S5.4, ratified by RULING-1): for a learning-rate
factor the forward and reverse replays coincide by construction, so that branch could not lose. The row-only arm's
fraction $s_{\mathrm{row}} = \ln(D_{\mathrm{row}}/D_{\mathrm{lo}})/\ln c_s$ is printed beside $s_{\mathrm{lr}}$ and is
never subtracted from it, divided into it or combined with it.

\paragraph{The shadow} (\path{PREREG_SHADOW.md} \S0--\S5; same-world comparison in its AMENDMENT-1 and -2). Each cell's
gap $F$ is printed before $c_{\mathrm{sh}}$; $|F| < \ln \qShadowXThreePUndefFactor$ would leave the cell undefined, and no
cell was. Out-of-range cells carry their coordinate and word only, with no attribution sentence; the in-range cells of the
learning-rate arm are printed the same way. The outcome set has no band around $\qShadowXThreePOutAbove$, so a coordinate
just above it and one far above it receive the same word; this is a limit of the registration and was not changed after
the read. The same-world comparison was registered after the all-rows read had come out out of range; for seeds 0--2 of
the sampled-negative head ``the $c'$ of seeds 0--2 was already fixed at registration by numbers that had been read; it is
recorded as descriptive and is not a test of the prediction'' (A1.5). Its prediction, both directions in the full band, is
registered to be scored on seeds 3--4 and on the softmax head (A1.5); those reads are not reported here, and no test of
the prediction is.

\paragraph{Gradient size} (\path{PREREG_X11.md}). $L_s = \ln\bigl(\sum_t g_{\mathrm{rest}}(t) / \sum_t
g_{\mathrm{lo}}(t)\bigr)$ over the read points common to the two runs, where $g(t)$ is the mean over the probe row's
coordinates of the square root of Adam's (uncorrected) second-moment accumulator. It sees the gradient's size, not its
direction or step-to-step coherence. The seed spread, the range over seeds of the low-$\eps$ runs' log sums, is a flag
printed beside the word, not a criterion.

\begin{table}[h]
\centering\small
\caption{Registered outcome sets of the decomposition and the shadow (thresholds as registered; the per-run, per-seed
and per-unit states that are never a branch, such as VOID, MISSING and NOT-COVERED, are omitted).}
\label{tab:app-decomp-sets}
\resizebox{\linewidth}{!}{%
\begin{tabular}{@{}lll@{}}
\toprule
Test (row) & branches & predicted \\
\midrule
Per-row $\eps$, $r$ (Q-STDCFG-X8, -X8EXT, -X8SM)
  & \qStdcfgXEightSetADenominator{} $r \ge \qStdcfgXEightAGe$; \qStdcfgXEightSetBTrajectory{} $r \le \qStdcfgXEightBLe$;
    \qStdcfgXEightSetCMixed{} between; & none \\
  & \qStdcfgXEightSetOutOfRange{} $r < \qStdcfgXEightOutBelow$ or $r > \qStdcfgXEightOutAbove$;
    \qStdcfgXEightSetUndefined{} denominator not positive & \\
Per-row lr, $s_{\mathrm{lr}}$ (Q-STDCFG-X9, -X9EXT)
  & \qStdcfgXNineSetSameSide{} non-negative, $< \qStdcfgXNineNeutralLo$; \qStdcfgXNineSetNeutral{}
    $[\qStdcfgXNineNeutralLo, \qStdcfgXNineNeutralHi]$; & \qStdcfgXNineSetSameSide \\
  & \qStdcfgXNineSetOpposite{} $> \qStdcfgXNineNeutralHi$; \qStdcfgXNineSetOutOfRange{} negative & \\
Shadow, $c_{\mathrm{sh}}$ (Q-SHADOW-X3P, -X3PX9)
  & \qShadowXThreePXNineSetLocal{} $\ge \qShadowXThreePXNineLocalGe$; \qShadowXThreePXNineSetGlobal{}
    $\le \qShadowXThreePXNineGlobalLe$; \qShadowXThreePXNineSetMixed{} between; & none \\
  & \qShadowXThreePXNineSetOutOfRange{} $< \qShadowXThreePXNineOutBelow$ or $> \qShadowXThreePXNineOutAbove$;
    \qShadowXThreePXNineSetUndefined{} $|F| < \ln\qShadowXThreePUndefFactor$ & \\
Same world, $c'$ (Q-SHADOW-X3PROW)
  & \qShadowXThreePRowSetFull{} $[\qShadowXThreePRowFullLo, \qShadowXThreePRowFullHi]$; \qShadowXThreePRowSetPartial{}
    $[\qShadowXThreePRowPartialLo, \qShadowXThreePRowFullLo)$; & \qShadowXThreePRowSetFull{} \\
  & \qShadowXThreePRowSetNone{} $[\qShadowXThreePRowNoneLo, \qShadowXThreePRowPartialLo)$;
    \qShadowXThreePRowSetOver{} $> \qShadowXThreePRowOverAbove$; \qShadowXThreePRowSetUnder{} $< \qShadowXThreePRowUnderBelow$;
    & (both directions) \\
  & \qShadowXThreePRowSetUndefined{} row-arm gap below $\ln\qShadowXThreePRowUndefFactor$ & \\
Gradient size, $L$ (Q-X11)
  & \qXElevenSetWeaker{} $L \le -\ln\qXElevenFactor$; \qXElevenSetSame{} $|L| < \ln\qXElevenFactor$;
    \qXElevenSetStronger{} $L \ge \ln\qXElevenFactor$ & \qXElevenPrediction \\
\bottomrule
\end{tabular}}
\end{table}

\subsection{Per-seed and per-cell values}
\label{app:decomp-values}

\begin{table}[h]
\centering\small
\caption{Per-row $\eps$, per seed, at the readers' printed precision. The last column is printed, not a criterion.}
\label{tab:app-x8}
\resizebox{\linewidth}{!}{%
\begin{tabular}{@{}llrrlrlr@{}}
\toprule
Population (row) & seed & $D_{\mathrm{lo}}-D_{\mathrm{hi}}$ & $r_{\mathrm{row}}$ & word & $r_{\mathrm{rest}}$ & word
  & $r_{\mathrm{row}}+r_{\mathrm{rest}}$ \\
\midrule
sampled-negative (Q-STDCFG-X8)
  & 0 & \qStdcfgXEightDenomSeedZero & \qStdcfgXEightRowRPrintedSeedZero & \qStdcfgXEightRowSeedZero
      & \qStdcfgXEightRestRPrintedSeedZero & \qStdcfgXEightRestSeedZero & \qStdcfgXEightSumSeedZero \\
  & 1 & \qStdcfgXEightDenomSeedOne & \qStdcfgXEightRowRPrintedSeedOne & \qStdcfgXEightRowSeedOne
      & \qStdcfgXEightRestRPrintedSeedOne & \qStdcfgXEightRestSeedOne & \qStdcfgXEightSumSeedOne \\
  & 2 & \qStdcfgXEightDenomSeedTwo & \qStdcfgXEightRowRPrintedSeedTwo & \qStdcfgXEightRowSeedTwo
      & \qStdcfgXEightRestRPrintedSeedTwo & \qStdcfgXEightRestSeedTwo & \qStdcfgXEightSumSeedTwo \\
\midrule
sampled-negative, read on its own (Q-STDCFG-X8EXT)
  & 3 & \qStdcfgXEightExtDenomSeedThree & \qStdcfgXEightExtRowRPrintedSeedThree & \qStdcfgXEightExtRowSeedThree
      & \qStdcfgXEightExtRestRPrintedSeedThree & \qStdcfgXEightExtRestSeedThree & \qStdcfgXEightExtSumSeedThree \\
  & 4 & \qStdcfgXEightExtDenomSeedFour & \qStdcfgXEightExtRowRPrintedSeedFour & \qStdcfgXEightExtRowSeedFour
      & \qStdcfgXEightExtRestRPrintedSeedFour & \qStdcfgXEightExtRestSeedFour & \qStdcfgXEightExtSumSeedFour \\
\midrule
softmax, read on its own (Q-STDCFG-X8SM)
  & 0 & \qStdcfgXEightSoftmaxDenomSeedZero & \qStdcfgXEightSoftmaxRowRPrintedSeedZero & \qStdcfgXEightSoftmaxRowSeedZero
      & \qStdcfgXEightSoftmaxRestRPrintedSeedZero & \qStdcfgXEightSoftmaxRestSeedZero & \qStdcfgXEightSoftmaxSumSeedZero \\
  & 1 & \qStdcfgXEightSoftmaxDenomSeedOne & \qStdcfgXEightSoftmaxRowRPrintedSeedOne & \qStdcfgXEightSoftmaxRowSeedOne
      & \qStdcfgXEightSoftmaxRestRPrintedSeedOne & \qStdcfgXEightSoftmaxRestSeedOne & \qStdcfgXEightSoftmaxSumSeedOne \\
  & 2 & \qStdcfgXEightSoftmaxDenomSeedTwo & \qStdcfgXEightSoftmaxRowRPrintedSeedTwo & \qStdcfgXEightSoftmaxRowSeedTwo
      & \qStdcfgXEightSoftmaxRestRPrintedSeedTwo & \qStdcfgXEightSoftmaxRestSeedTwo & \qStdcfgXEightSoftmaxSumSeedTwo \\
\bottomrule
\end{tabular}}
\end{table}

\begin{table}[h]
\centering\small
\caption{Per-row learning rate, per seed. $s_{\mathrm{row}}$ is the row-only $\eps$ arm's fraction against the same $c_s$,
printed beside $s_{\mathrm{lr}}$ and never combined with it.}
\label{tab:app-x9}
\begin{tabular}{@{}llrrrlr@{}}
\toprule
Population (row) & seed & $c_s$ & $\rho_{\mathrm{meas}}$ & $s_{\mathrm{lr}}$ & word & $s_{\mathrm{row}}$ (beside) \\
\midrule
sampled-negative (Q-STDCFG-X9)
  & 0 & \qStdcfgXNineCsSeedZero & \qStdcfgXNineRhoMeasSeedZero & \qStdcfgXNineSlrPrintedSeedZero & \qStdcfgXNineSeedZero
      & \qStdcfgXNineShareRowSeedZero \\
  & 1 & \qStdcfgXNineCsSeedOne & \qStdcfgXNineRhoMeasSeedOne & \qStdcfgXNineSlrPrintedSeedOne & \qStdcfgXNineSeedOne
      & \qStdcfgXNineShareRowSeedOne \\
  & 2 & \qStdcfgXNineCsSeedTwo & \qStdcfgXNineRhoMeasSeedTwo & \qStdcfgXNineSlrPrintedSeedTwo & \qStdcfgXNineSeedTwo
      & \qStdcfgXNineShareRowSeedTwo \\
\midrule
sampled-negative, read on its own (Q-STDCFG-X9EXT)
  & 3 & \qStdcfgXNineExtCsSeedThree & \qStdcfgXNineExtRhoMeasSeedThree & \qStdcfgXNineExtSlrPrintedSeedThree
      & \qStdcfgXNineExtSeedThree & \qStdcfgXNineExtShareRowSeedThree \\
  & 4 & \qStdcfgXNineExtCsSeedFour & \qStdcfgXNineExtRhoMeasSeedFour & \qStdcfgXNineExtSlrPrintedSeedFour
      & \qStdcfgXNineExtSeedFour & \qStdcfgXNineExtShareRowSeedFour \\
\bottomrule
\end{tabular}
\end{table}

\begin{table}[h]
\centering\small
\caption{The shadow, per cell: coordinate $c_{\mathrm{sh}}$ and word as the readers printed them. Top: the learning-rate
arm (Q-SHADOW-X3PX9). Bottom: the pairs of Section~\ref{sec:x1}, every parameter's $\eps$ changed (Q-SHADOW-X3P).}
\label{tab:app-x3p}
\resizebox{\linewidth}{!}{%
\begin{tabular}{@{}lrlrlrl@{}}
\toprule
Cell & \multicolumn{2}{c}{seed 0} & \multicolumn{2}{c}{seed 1} & \multicolumn{2}{c}{seed 2} \\
\midrule
learning-rate arm, forward
  & \qShadowXThreePXNineCvalForwardSeedZero & \qShadowXThreePXNineForwardSeedZero
  & \qShadowXThreePXNineCvalForwardSeedOne & \qShadowXThreePXNineForwardSeedOne
  & \qShadowXThreePXNineCvalForwardSeedTwo & \qShadowXThreePXNineForwardSeedTwo \\
learning-rate arm, reverse
  & \qShadowXThreePXNineCvalReverseSeedZero & \qShadowXThreePXNineReverseSeedZero
  & \qShadowXThreePXNineCvalReverseSeedOne & \qShadowXThreePXNineReverseSeedOne
  & \qShadowXThreePXNineCvalReverseSeedTwo & \qShadowXThreePXNineReverseSeedTwo \\
\midrule
sampled-negative, three-decade, forward
  & \qShadowXThreePCvalBceFifteenThreeDecadeForwardSeedZero & \qShadowXThreePBceFifteenThreeDecadeForwardSeedZero
  & \qShadowXThreePCvalBceFifteenThreeDecadeForwardSeedOne & \qShadowXThreePBceFifteenThreeDecadeForwardSeedOne
  & \qShadowXThreePCvalBceFifteenThreeDecadeForwardSeedTwo & \qShadowXThreePBceFifteenThreeDecadeForwardSeedTwo \\
sampled-negative, three-decade, reverse
  & \qShadowXThreePCvalBceFifteenThreeDecadeReverseSeedZero & \qShadowXThreePBceFifteenThreeDecadeReverseSeedZero
  & \qShadowXThreePCvalBceFifteenThreeDecadeReverseSeedOne & \qShadowXThreePBceFifteenThreeDecadeReverseSeedOne
  & \qShadowXThreePCvalBceFifteenThreeDecadeReverseSeedTwo & \qShadowXThreePBceFifteenThreeDecadeReverseSeedTwo \\
sampled-negative, four-decade, forward
  & \qShadowXThreePCvalBceFifteenFourDecadeForwardSeedZero & \qShadowXThreePBceFifteenFourDecadeForwardSeedZero
  & \qShadowXThreePCvalBceFifteenFourDecadeForwardSeedOne & \qShadowXThreePBceFifteenFourDecadeForwardSeedOne
  & \qShadowXThreePCvalBceFifteenFourDecadeForwardSeedTwo & \qShadowXThreePBceFifteenFourDecadeForwardSeedTwo \\
sampled-negative, four-decade, reverse
  & \qShadowXThreePCvalBceFifteenFourDecadeReverseSeedZero & \qShadowXThreePBceFifteenFourDecadeReverseSeedZero
  & \qShadowXThreePCvalBceFifteenFourDecadeReverseSeedOne & \qShadowXThreePBceFifteenFourDecadeReverseSeedOne
  & \qShadowXThreePCvalBceFifteenFourDecadeReverseSeedTwo & \qShadowXThreePBceFifteenFourDecadeReverseSeedTwo \\
\midrule
softmax, four-decade, forward
  & \qShadowXThreePCvalSoftmaxFourDecadeForwardSeedZero & \qShadowXThreePSoftmaxFourDecadeForwardSeedZero
  & \qShadowXThreePCvalSoftmaxFourDecadeForwardSeedOne & \qShadowXThreePSoftmaxFourDecadeForwardSeedOne
  & \qShadowXThreePCvalSoftmaxFourDecadeForwardSeedTwo & \qShadowXThreePSoftmaxFourDecadeForwardSeedTwo \\
softmax, four-decade, reverse
  & \qShadowXThreePCvalSoftmaxFourDecadeReverseSeedZero & \qShadowXThreePSoftmaxFourDecadeReverseSeedZero
  & \qShadowXThreePCvalSoftmaxFourDecadeReverseSeedOne & \qShadowXThreePSoftmaxFourDecadeReverseSeedOne
  & \qShadowXThreePCvalSoftmaxFourDecadeReverseSeedTwo & \qShadowXThreePSoftmaxFourDecadeReverseSeedTwo \\
\bottomrule
\end{tabular}}
\end{table}

\begin{table}[h]
\centering\small
\caption{The same-world comparison, sampled-negative head, seeds 0--2: $c'$ and word as printed. Descriptive (A1.5):
the inputs were printed before the comparison was registered; not a test of its prediction (Q-SHADOW-X3PROW).}
\label{tab:app-x3prow}
\begin{tabular}{@{}lrlrlrl@{}}
\toprule
Direction & \multicolumn{2}{c}{seed 0} & \multicolumn{2}{c}{seed 1} & \multicolumn{2}{c}{seed 2} \\
\midrule
forward (row-only run)
  & \qShadowXThreePRowCvalPrimeForwardSeedZero & \qShadowXThreePRowForwardSeedZero
  & \qShadowXThreePRowCvalPrimeForwardSeedOne & \qShadowXThreePRowForwardSeedOne
  & \qShadowXThreePRowCvalPrimeForwardSeedTwo & \qShadowXThreePRowForwardSeedTwo \\
reverse (rest-only run)
  & \qShadowXThreePRowCvalPrimeReverseSeedZero & \qShadowXThreePRowReverseSeedZero
  & \qShadowXThreePRowCvalPrimeReverseSeedOne & \qShadowXThreePRowReverseSeedOne
  & \qShadowXThreePRowCvalPrimeReverseSeedTwo & \qShadowXThreePRowReverseSeedTwo \\
\bottomrule
\end{tabular}
\end{table}

\paragraph{Gradient size, per seed} (Q-X11; registered prediction \qXElevenPrediction{} on every seed). Seed 0: $L =
\qXElevenLValSeedZero$, \qXElevenLSeedZero; seed 1: $\qXElevenLValSeedOne$, \qXElevenLSeedOne; seed 2:
$\qXElevenLValSeedTwo$, \qXElevenLSeedTwo. Seed spread $\qXElevenNullScale$; no seed lies within it. The registration
wrote a losing consequence only for \qXElevenSetSame{}; a ruling made after the read gives \qXElevenSetStronger{} the same
consequence, so the route through the gradient's size is rejected. A side print the reader made (a momentum-norm ratio)
is not cited, by ruling.

\subsection{Controls of the kill test}
\label{app:decomp-controls}

\paragraph{Held-out rare targets} (\path{PREREG_STDCFG.md} \S3; Q-STDCFG-X6). After training, once, the next-token
cross-entropy on $\qStdcfgXSixRarePositions$ held-out rare-target positions; per head, $\bar\Delta(d)$ is the mean over
seeds of each seed's change from its low-$\eps$ run at dose $d$, and $\sigma$ the standard deviation over the seeds'
low-$\eps$ values, itself unstable with so few seeds. \qStdcfgXSixSetA{}: flat at both doses ($|\bar\Delta| \le \sigma$);
\qStdcfgXSixSetB{}: moves with dose (same sign, larger at the four-decade dose, beyond twice $\sigma$ at one dose at
least; the softmax head has the four-decade dose only); \qStdcfgXSixSetNeither{}: otherwise. Sampled-negative head:
\qStdcfgXSixBce{} ($\sigma = \qStdcfgXSixBceSigma$, unstable; $\bar\Delta = \qStdcfgXSixBceDbarThree$ at three decades and
$\qStdcfgXSixBceDbarFour$ at four). Softmax head: \qStdcfgXSixSoftmax{}, direction \qStdcfgXSixSoftmaxDirection{}
($\sigma = \qStdcfgXSixSoftmaxSigma$, unstable; $\bar\Delta = \qStdcfgXSixSoftmaxDbarFour$); the direction is printed, not
interpreted. The all-position control, judged by the same outcome set with its own $\sigma$, reads \qStdcfgXSixBceControl{}
and \qStdcfgXSixSoftmaxControl{} (\qStdcfgXSixSoftmaxControlDirection), printed beside. The heads are never merged. Blind
spot: $\qStdcfgXSixRareFromTail$ of the $\qStdcfgXSixRarePositions$ positions come from $\qStdcfgXSixTailRows$ rows at the
corpus tail, one topical stretch, so the test does not separate a consequence of $\eps$ from that text's response to it;
the strata are printed, not scored. Nothing here is a statement about generalization, and the held-out values are never
set against the training-side displacement.

\paragraph{A panel of rows on one trajectory} (\path{PREREG_STDCFG.md} AMENDMENT-27; Q-STDCFG-X7). The one sentence the
test licenses: ``\qStdcfgXSeven''. The panel's two runs repeat the seed-0 four-decade pair of Section~\ref{sec:x1} on its
machine and are bit-identical to it in trajectory and read items (\qStdcfgXSevenInert; \qStdcfgXSevenReadItems{} read items
per run), so the panel rows are coordinates of one trajectory, not replications. Per-row $e^F$ runs from
$\qStdcfgXSevenExpFLo$ to $\qStdcfgXSevenExpFHi$ (largest over smallest $\qStdcfgXSevenMaxOverMin$), set beside the earlier
configuration's panel ($\qStdcfgXSevenOldPanelLo$--$\qStdcfgXSevenOldPanelHi$, a different configuration, never pooled)
and the seed-0 probe row's $\qStdcfgXSevenProbeExpF$. The read preceded its structural audit, which is therefore a
post-read audit. Nothing is claimed for all rows or for the model as a whole.

\section{The width-768 batch}
\label{app:w768}

Every value of Section~\ref{sec:w768}, per seed and per cell, never pooled with the base width. Rows
\texttt{Q-STDCFG768-LR}, \texttt{-X1}, \texttt{-X1-SOFTMAX}, \texttt{-SIZE}, \texttt{-SIZE-SOFTMAX}, \texttt{-X8},
\texttt{-X9}; reading rules \texttt{PREREG\_STDCFG\_768.md} \S6.10. Machine and host labels are the registry's; every unit
ran on one machine, and each size cell compares runs from one machine; device \qStdcfgSevenSixEightDevice.

\subsection{The learning-rate rule at this width}
The rule of Section~\ref{sec:setup}, run again at width \qStdcfgSevenSixEightDModel{} rather than carried over, read
\qStdcfgSevenSixEightLrRuns{} preflight runs.
\begin{center}\small
\begin{tabular}{llll}
\toprule
candidate peak rate & seeds judged & seeds eligible & mean over seeds of $E_{\mathrm{read}}$ (last ten windows) \\
\midrule
\qStdcfgSevenSixEightLrRungOne & \qStdcfgSevenSixEightLrRungOneSeeds & \qStdcfgSevenSixEightLrRungOneEligible & \qStdcfgSevenSixEightLrRungOneERead \\
\qStdcfgSevenSixEightLrRungTwo & \qStdcfgSevenSixEightLrRungTwoSeeds & \qStdcfgSevenSixEightLrRungTwoEligible & \qStdcfgSevenSixEightLrRungTwoERead \\
\qStdcfgSevenSixEightLrRungThree & \qStdcfgSevenSixEightLrRungThreeSeeds & \qStdcfgSevenSixEightLrRungThreeEligible & \qStdcfgSevenSixEightLrRungThreeERead \\
\bottomrule
\end{tabular}
\end{center}
Registered outcome set \{\qStdcfgSevenSixEightLrSetSelected, \qStdcfgSevenSixEightLrSetNone,
\qStdcfgSevenSixEightLrSetNotCovered, \qStdcfgSevenSixEightLrSetMachineLost\}; outcome \textbf{\qStdcfgSevenSixEightLr}, rate
\qStdcfgSevenSixEightLrArgv. Beside it, under the row's own label
(``\qStdcfgSevenSixEightLrBaseLabel''), the base width's selection: \qStdcfgSevenSixEightLrBaseOutcome,
rate \qStdcfgSevenSixEightLrBaseLr. The two selections are printed side by side and never compared.

\subsection{The kill test at this width}
\textbf{Main head.} The read registered \qStdcfgSevenSixEightXOneEntries{} entries (the runs and the certificate);
\qStdcfgSevenSixEightXOneLanded{} landed and \qStdcfgSevenSixEightXOneClosed{} were closed by the record, after the host
(machine \qStdcfgSevenSixEightMachineMain, host \qStdcfgSevenSixEightHostMain) restarted before any run of seeds
\qStdcfgSevenSixEightXOneSeedOne{} and \qStdcfgSevenSixEightXOneSeedTwo{} finished (first recorded as a lost connection,
category \qStdcfgSevenSixEightMissingCatFirst, and corrected after the read to a host restart, category
\qStdcfgSevenSixEightMissingCatCorrected, restart token
\qStdcfgSevenSixEightReopenToken). The certificate word is ``\qStdcfgSevenSixEightXOneCert''. Cell words, seeds \qStdcfgSevenSixEightXOneSeedZero,
\qStdcfgSevenSixEightXOneSeedOne{} and \qStdcfgSevenSixEightXOneSeedTwo{} in order: \qStdcfgSevenSixEightDoseFour{}
\qStdcfgSevenSixEightXOneWordFourSZero, \qStdcfgSevenSixEightXOneWordFourSOne, \qStdcfgSevenSixEightXOneWordFourSTwo;
\qStdcfgSevenSixEightDoseThree{} \qStdcfgSevenSixEightXOneWordThreeSZero, \qStdcfgSevenSixEightXOneWordThreeSOne,
\qStdcfgSevenSixEightXOneWordThreeSTwo. Seed
\qStdcfgSevenSixEightXOneSeedZero{}, every value from the reader's output, rounded as printed here:
\begin{center}\small
\begin{tabular}{lllllllll}
\toprule
pair & $c$ (nats) & $c-\ln 4$ & kill ratio & $e^{F}$ & $F$ (nats) & $s$ & side & crossing step (low / high run) \\
\midrule
\qStdcfgSevenSixEightDoseFour & \qStdcfgSevenSixEightXOneCellCeilFour & \qStdcfgSevenSixEightXOneCellHeadroomFour & \qStdcfgSevenSixEightXOneCellKillFour & \qStdcfgSevenSixEightXOneCellExpFFour & \qStdcfgSevenSixEightXOneCellFFour & \qStdcfgSevenSixEightXOneCellShareFour & \qStdcfgSevenSixEightXOneCellSideFour & \qStdcfgSevenSixEightXOneCellEscLoFour{} / \qStdcfgSevenSixEightXOneCellEscHiFour \\
\qStdcfgSevenSixEightDoseThree & \qStdcfgSevenSixEightXOneCellCeilThree & \qStdcfgSevenSixEightXOneCellHeadroomThree & \qStdcfgSevenSixEightXOneCellKillThree & \qStdcfgSevenSixEightXOneCellExpFThree & \qStdcfgSevenSixEightXOneCellFThree & \qStdcfgSevenSixEightXOneCellShareThree & --- & \qStdcfgSevenSixEightXOneCellEscLoThree{} / \qStdcfgSevenSixEightXOneCellEscHiThree \\
\bottomrule
\end{tabular}
\end{center}
Registered outcome set \{\qStdcfgSevenSixEightXOneSetGo, \qStdcfgSevenSixEightXOneSetNoGo, \qStdcfgSevenSixEightXOneSetNotCovered, \qStdcfgSevenSixEightXOneSetNotRun\}; outcome \textbf{\qStdcfgSevenSixEightXOne}.

\textbf{The reopened read, side by side only.} Seeds \qStdcfgSevenSixEightXOneSeedOne{} and
\qStdcfgSevenSixEightXOneSeedTwo{} were run again on machine \qStdcfgSevenSixEightReopenMachine{} (host
\qStdcfgSevenSixEightReopenHost; read token \qStdcfgSevenSixEightReopenToken) as a read of their own: the low runs after
the first restart, the higher-$\eps$ runs after a second restart (token \qStdcfgSevenSixEightRestartTwoToken), the
machine's gates run again and passed after each. Registered outcome set \{\qStdcfgSevenSixEightXOneSetGo,
\qStdcfgSevenSixEightXOneSetNoGo, \qStdcfgSevenSixEightXOneSetNotCovered\} (A9.1), with
\qStdcfgSevenSixEightXOneSetNotRun{} by the letter of \S6.10 only; its outcome is \textbf{\qStdcfgSevenSixEightReopenOutcome} by construction (seed \qStdcfgSevenSixEightXOneSeedZero{} has no run in it), and it
enters neither the main kill test, nor the softmax test, nor the size comparison. Cell words (four / three decades):
seed \qStdcfgSevenSixEightXOneSeedZero{} \qStdcfgSevenSixEightReopenWordFourSZero{} / \qStdcfgSevenSixEightReopenWordThreeSZero;
seed \qStdcfgSevenSixEightXOneSeedOne{} \qStdcfgSevenSixEightReopenWordFourSOne{} / \qStdcfgSevenSixEightReopenWordThreeSOne;
seed \qStdcfgSevenSixEightXOneSeedTwo{} \qStdcfgSevenSixEightReopenWordFourSTwo{} / \qStdcfgSevenSixEightReopenWordThreeSTwo.
Its own gate on the two low runs: \qStdcfgSevenSixEightReopenLoGateSOne{} and \qStdcfgSevenSixEightReopenLoGateSTwo;
certificate ``\qStdcfgSevenSixEightReopenCert'' (the reader's word; A9.1 and the audit call this certificate not covered
by construction).

\textbf{Softmax head} (second instrument; machine \qStdcfgSevenSixEightMachineSoftmax, host
\qStdcfgSevenSixEightHostSoftmax), \qStdcfgSevenSixEightDoseFour{} pairs:
\begin{center}\small
\begin{tabular}{lllllllll}
\toprule
seed & $c$ (nats) & $c-\ln 4$ & kill ratio & $e^{F}$ & $F$ (nats) & $s$ & side & crossing step (low / high run) \\
\midrule
\qStdcfgSevenSixEightXOneSoftmaxSeedZero & \qStdcfgSevenSixEightXOneSoftmaxCellCeilSZero & \qStdcfgSevenSixEightXOneSoftmaxCellHeadroomSZero & \qStdcfgSevenSixEightXOneSoftmaxCellKillSZero & \qStdcfgSevenSixEightXOneSoftmaxCellExpFSZero & \qStdcfgSevenSixEightXOneSoftmaxCellFSZero & \qStdcfgSevenSixEightXOneSoftmaxCellShareSZero & \qStdcfgSevenSixEightXOneSoftmaxCellSideSZero & \qStdcfgSevenSixEightXOneSoftmaxCellEscLoSZero{} / \qStdcfgSevenSixEightXOneSoftmaxCellEscHiSZero \\
\qStdcfgSevenSixEightXOneSoftmaxSeedOne & \qStdcfgSevenSixEightXOneSoftmaxCellCeilSOne & \qStdcfgSevenSixEightXOneSoftmaxCellHeadroomSOne & \qStdcfgSevenSixEightXOneSoftmaxCellKillSOne & \qStdcfgSevenSixEightXOneSoftmaxCellExpFSOne & \qStdcfgSevenSixEightXOneSoftmaxCellFSOne & \qStdcfgSevenSixEightXOneSoftmaxCellShareSOne & \qStdcfgSevenSixEightXOneSoftmaxCellSideSOne & \qStdcfgSevenSixEightXOneSoftmaxCellEscLoSOne{} / \qStdcfgSevenSixEightXOneSoftmaxCellEscHiSOne \\
\qStdcfgSevenSixEightXOneSoftmaxSeedTwo & \qStdcfgSevenSixEightXOneSoftmaxCellCeilSTwo & \qStdcfgSevenSixEightXOneSoftmaxCellHeadroomSTwo & \qStdcfgSevenSixEightXOneSoftmaxCellKillSTwo & \qStdcfgSevenSixEightXOneSoftmaxCellExpFSTwo & \qStdcfgSevenSixEightXOneSoftmaxCellFSTwo & \qStdcfgSevenSixEightXOneSoftmaxCellShareSTwo & \qStdcfgSevenSixEightXOneSoftmaxCellSideSTwo & \qStdcfgSevenSixEightXOneSoftmaxCellEscLoSTwo{} / \qStdcfgSevenSixEightXOneSoftmaxCellEscHiSTwo \\
\bottomrule
\end{tabular}
\end{center}
Registered outcome set \{\qStdcfgSevenSixEightXOneSoftmaxSetGo, \qStdcfgSevenSixEightXOneSoftmaxSetNoGo, \qStdcfgSevenSixEightXOneSoftmaxSetNotCovered, \qStdcfgSevenSixEightXOneSoftmaxSetNotRun\}; outcome \textbf{\qStdcfgSevenSixEightXOneSoftmax}; beside the
main head only.

\subsection{The size comparison}
$R$ is a cell's $e^{F}$ in the wider model over its $e^{F}$ at the base width, each pair at its own selected rate; a
cell is judged only where both widths' pairs ran on one machine. Before it took any wider-model high-run value (those
runs had already been read by the wider model's kill-test reads, which are its inputs), the size reader printed
$c_{768}$ (from the wider model's low run alone), $e^{c_{768}}$, the base-width $e^{F}$, the upper bound on $R$ (which
holds whenever the high run moves the row no more than the low run) and the threshold, and its reachability flag.

\textbf{Main head, judged} (the row's label: ``\qStdcfgSevenSixEightSizeLabel''): seed
\qStdcfgSevenSixEightXOneSeedZero, base-width pair on machine \qStdcfgSevenSixEightSizeMachineBase{} (trained again there
for this comparison), wider pair on machine \qStdcfgSevenSixEightSizeMachineWide. $c_{768}=\qStdcfgSevenSixEightSizeCeil$,
$e^{c_{768}}=\qStdcfgSevenSixEightSizeExpCeil$, base-width $e^{F}$ in the block printed first
\qStdcfgSevenSixEightSizeBlockRatioBase, upper bound on $R$ \qStdcfgSevenSixEightSizeRUpper, threshold
\qStdcfgSevenSixEightSizeThreshold, flag ``\qStdcfgSevenSixEightSizeFlag''. Fire states: base width
\qStdcfgSevenSixEightSizeStateBase, wider model \qStdcfgSevenSixEightSizeStateWide; $e^{F}$: base width
\qStdcfgSevenSixEightSizeCellRatioBase, wider model \qStdcfgSevenSixEightSizeCellRatioWide; $R=\qStdcfgSevenSixEightSizeCellR$;
peak rates \qStdcfgSevenSixEightSizeLrBase{} (base) and \qStdcfgSevenSixEightSizeLrWide{} (wider). Registered outcome
set \{\qStdcfgSevenSixEightSizeSetSame, \qStdcfgSevenSixEightSizeSetChanges, \qStdcfgSevenSixEightSizeSetNotCovered,
\qStdcfgSevenSixEightSizeSetNotRun\}; outcome \textbf{\qStdcfgSevenSixEightSize}.

\textbf{Main head, side by side, not judged} (the row's label: ``\qStdcfgSevenSixEightSizeSbsLabel''): the base-width
values of the kill test's own pairs, which ran on another machine.
\begin{center}\small
\begin{tabular}{llllll}
\toprule
seed & base-width state & base-width $e^{F}$ & base-width machine (host) & wider-model state & wider-model machine \\
\midrule
\qStdcfgSevenSixEightXOneSeedZero & \qStdcfgSevenSixEightSizeSbsStateBaseSZero & \qStdcfgSevenSixEightSizeSbsRatioBaseSZero & \qStdcfgSevenSixEightSizeSbsMachineBaseSZero{} (\qStdcfgSevenSixEightSizeSbsHostBaseSZero) & \qStdcfgSevenSixEightSizeSbsStateWideSZero & \qStdcfgSevenSixEightSizeSbsMachineWideSZero \\
\qStdcfgSevenSixEightXOneSeedOne & \qStdcfgSevenSixEightSizeSbsStateBaseSOne & \qStdcfgSevenSixEightSizeSbsRatioBaseSOne & \qStdcfgSevenSixEightSizeSbsMachineBaseSOne{} (\qStdcfgSevenSixEightSizeSbsHostBaseSOne) & \qStdcfgSevenSixEightSizeSbsStateWideSOne & \qStdcfgSevenSixEightSizeSbsMachineWideSOne \\
\qStdcfgSevenSixEightXOneSeedTwo & \qStdcfgSevenSixEightSizeSbsStateBaseSTwo & \qStdcfgSevenSixEightSizeSbsRatioBaseSTwo & \qStdcfgSevenSixEightSizeSbsMachineBaseSTwo{} (\qStdcfgSevenSixEightSizeSbsHostBaseSTwo) & \qStdcfgSevenSixEightSizeSbsStateWideSTwo & \qStdcfgSevenSixEightSizeSbsMachineWideSTwo \\
\bottomrule
\end{tabular}
\end{center}
The seed-\qStdcfgSevenSixEightXOneSeedZero{} base-width value is printed here under this label and in the judged cell
under its own label; the two are not set against each other as a comparison across machines.

\noindent\begin{minipage}{\linewidth}\textbf{Softmax head, judged cell by cell} (the row's label: ``\qStdcfgSevenSixEightSizeSmLabel''):
\begin{center}\small
\resizebox{\linewidth}{!}{%
\begin{tabular}{llllllllll}
\toprule
seed & machines & $c_{768}$ & $e^{c_{768}}$ & upper bound & threshold & states (base / wider) & $e^{F}$ (base / wider) & $R$ & rates (base / wider) \\
\midrule
\qStdcfgSevenSixEightXOneSoftmaxSeedZero & \qStdcfgSevenSixEightSizeSmMachineBaseSZero{} / \qStdcfgSevenSixEightSizeSmMachineWideSZero & \qStdcfgSevenSixEightSizeSmCeilSZero & \qStdcfgSevenSixEightSizeSmExpCeilSZero & \qStdcfgSevenSixEightSizeSmRUpperSZero & \qStdcfgSevenSixEightSizeSmThresholdSZero & \qStdcfgSevenSixEightSizeSmStateBaseSZero{} / \qStdcfgSevenSixEightSizeSmStateWideSZero & \qStdcfgSevenSixEightSizeSmCellRatioBaseSZero{} / \qStdcfgSevenSixEightSizeSmCellRatioWideSZero & \qStdcfgSevenSixEightSizeSmCellRSZero & \qStdcfgSevenSixEightSizeSmLrBaseSZero{} / \qStdcfgSevenSixEightSizeSmLrWideSZero \\
\qStdcfgSevenSixEightXOneSoftmaxSeedOne & \qStdcfgSevenSixEightSizeSmMachineBaseSOne{} / \qStdcfgSevenSixEightSizeSmMachineWideSOne & \qStdcfgSevenSixEightSizeSmCeilSOne & \qStdcfgSevenSixEightSizeSmExpCeilSOne & \qStdcfgSevenSixEightSizeSmRUpperSOne & \qStdcfgSevenSixEightSizeSmThresholdSOne & \qStdcfgSevenSixEightSizeSmStateBaseSOne{} / \qStdcfgSevenSixEightSizeSmStateWideSOne & \qStdcfgSevenSixEightSizeSmCellRatioBaseSOne{} / \qStdcfgSevenSixEightSizeSmCellRatioWideSOne & \qStdcfgSevenSixEightSizeSmCellRSOne & \qStdcfgSevenSixEightSizeSmLrBaseSOne{} / \qStdcfgSevenSixEightSizeSmLrWideSOne \\
\qStdcfgSevenSixEightXOneSoftmaxSeedTwo & \qStdcfgSevenSixEightSizeSmMachineBaseSTwo{} / \qStdcfgSevenSixEightSizeSmMachineWideSTwo & \qStdcfgSevenSixEightSizeSmCeilSTwo & \qStdcfgSevenSixEightSizeSmExpCeilSTwo & \qStdcfgSevenSixEightSizeSmRUpperSTwo & \qStdcfgSevenSixEightSizeSmThresholdSTwo & \qStdcfgSevenSixEightSizeSmStateBaseSTwo{} / \qStdcfgSevenSixEightSizeSmStateWideSTwo & \qStdcfgSevenSixEightSizeSmCellRatioBaseSTwo{} / \qStdcfgSevenSixEightSizeSmCellRatioWideSTwo & \qStdcfgSevenSixEightSizeSmCellRSTwo & \qStdcfgSevenSixEightSizeSmLrBaseSTwo{} / \qStdcfgSevenSixEightSizeSmLrWideSTwo \\
\bottomrule
\end{tabular}}
\end{center}
\end{minipage}\par\noindent
Every softmax cell's reachability flag reads ``\qStdcfgSevenSixEightSizeSmFlagSZero'', ``\qStdcfgSevenSixEightSizeSmFlagSOne''
and ``\qStdcfgSevenSixEightSizeSmFlagSTwo'' in seed order. Registered outcome set of each cell
\{\qStdcfgSevenSixEightSizeSmSetSame, \qStdcfgSevenSixEightSizeSmSetChanges, \qStdcfgSevenSixEightSizeSmSetNotCovered,
\qStdcfgSevenSixEightSizeSmSetNotRun\}; outcomes in seed order \textbf{\qStdcfgSevenSixEightSizeSoftmaxZero},
\textbf{\qStdcfgSevenSixEightSizeSoftmaxOne}, \textbf{\qStdcfgSevenSixEightSizeSoftmaxTwo}.

\subsection{Per-row $\eps$ and per-row learning rate at this width}
The unit read registered \qStdcfgSevenSixEightXEightEntries{} entries; \qStdcfgSevenSixEightXEightLanded{} landed and
\qStdcfgSevenSixEightXEightClosed{} were closed by a record. $D$ is the probe row's displacement in each arm; the
denominator $D_{\mathrm{lo}}-D_{\mathrm{hi}}$ is formed before $r$ (by the reader's construction), and
$r_{\mathrm{row}}+r_{\mathrm{rest}}$ is printed,
never a criterion.
\begin{center}\small
\begin{tabular}{llllllllll}
\toprule
seed & $D_{\mathrm{lo}}$ & $D_{\mathrm{hi}}$ & $D_{\mathrm{row}}$ & $D_{\mathrm{rest}}$ & denominator & $r_{\mathrm{row}}$ & $r_{\mathrm{rest}}$ & sum & bridge \\
\midrule
\qStdcfgSevenSixEightXEightSeedZero & \qStdcfgSevenSixEightXEightDEightSZero & \qStdcfgSevenSixEightXEightDFourSZero & \qStdcfgSevenSixEightXEightDRowSZero & \qStdcfgSevenSixEightXEightDRestSZero & \qStdcfgSevenSixEightXEightDenomSZero & \qStdcfgSevenSixEightXEightCellRRowSZero & \qStdcfgSevenSixEightXEightCellRRestSZero & \qStdcfgSevenSixEightXEightSumSZero & \qStdcfgSevenSixEightXEightBridgeSZero \\
\qStdcfgSevenSixEightXEightSeedOne & \qStdcfgSevenSixEightXEightDEightSOne & \qStdcfgSevenSixEightXEightDFourSOne & \qStdcfgSevenSixEightXEightDRowSOne & \qStdcfgSevenSixEightXEightDRestSOne & \qStdcfgSevenSixEightXEightDenomSOne & \qStdcfgSevenSixEightXEightCellRRowSOne & \qStdcfgSevenSixEightXEightCellRRestSOne & \qStdcfgSevenSixEightXEightSumSOne & \qStdcfgSevenSixEightXEightBridgeSOne \\
\bottomrule
\end{tabular}
\end{center}
Seed \qStdcfgSevenSixEightXEightSeedTwo: the arm with the low $\eps$ everywhere is \qStdcfgSevenSixEightXEightArmAllEightSTwo{}
(a second host restart cut it off), the bridge decision is \qStdcfgSevenSixEightXEightBridgeSTwo{} with state
\qStdcfgSevenSixEightXEightBridgeStateSTwo, and the seed has no coordinate and no learning-rate arm. Registered outcome
set of X8-768, per seed and coordinate, \{\qStdcfgSevenSixEightXEightSetA, \qStdcfgSevenSixEightXEightSetB,
\qStdcfgSevenSixEightXEightSetC, \qStdcfgSevenSixEightXEightSetOut, \qStdcfgSevenSixEightXEightSetUndefined,
\qStdcfgSevenSixEightXEightSetVoid, \qStdcfgSevenSixEightXEightSetMissing, \qStdcfgSevenSixEightXEightSetNotCovered,
\qStdcfgSevenSixEightXEightSetNotRun\}; outcomes in seed order ($r_{\mathrm{row}}$\,/\,$r_{\mathrm{rest}}$)
\textbf{\qStdcfgSevenSixEightXEightSZeroRRow}\,/\,\textbf{\qStdcfgSevenSixEightXEightSZeroRRest},
\textbf{\qStdcfgSevenSixEightXEightSOneRRow}\,/\,\textbf{\qStdcfgSevenSixEightXEightSOneRRest},
\textbf{\qStdcfgSevenSixEightXEightSTwoRRow}\,/\,\textbf{\qStdcfgSevenSixEightXEightSTwoRRest}.

Per-row learning rate: $c_s$ is the replay's $\rho_{\mathrm{cf}}$ at the \qStdcfgSevenSixEightDoseFour{} dose, the
arm's row learning-rate factor; $\rho_{\mathrm{meas}}$ is the arm against the all-low run; the row-only arm's
$s_{\mathrm{row}}$ is printed beside $s_{\mathrm{lr}}$ and never subtracted from it or compared with it.
Seed \qStdcfgSevenSixEightXEightSeedZero: $c_s=\qStdcfgSevenSixEightXNineCellCsSZero$,
$\rho_{\mathrm{meas}}=\qStdcfgSevenSixEightXNineCellRhoSZero$, $s_{\mathrm{lr}}=\qStdcfgSevenSixEightXNineCellSlrSZero$
($s_{\mathrm{row}}$ \qStdcfgSevenSixEightXNineCellShareRowSZero). Seed \qStdcfgSevenSixEightXEightSeedOne:
$c_s=\qStdcfgSevenSixEightXNineCellCsSOne$, $\rho_{\mathrm{meas}}=\qStdcfgSevenSixEightXNineCellRhoSOne$,
$s_{\mathrm{lr}}=\qStdcfgSevenSixEightXNineCellSlrSOne$ ($s_{\mathrm{row}}$ \qStdcfgSevenSixEightXNineCellShareRowSOne). Registered
outcome set of X9-768, per seed, \{\qStdcfgSevenSixEightXNineSetSameSide, \qStdcfgSevenSixEightXNineSetNeutral,
\qStdcfgSevenSixEightXNineSetOpposite, \qStdcfgSevenSixEightXNineSetOut, \qStdcfgSevenSixEightXNineSetVoid,
\qStdcfgSevenSixEightXNineSetMissing, \qStdcfgSevenSixEightXNineSetNotCovered, \qStdcfgSevenSixEightXNineSetNotRun\}, the
prediction \qStdcfgSevenSixEightXNineSetSameSide{} on every seed; outcomes in seed order
\textbf{\qStdcfgSevenSixEightXNineZero}, \textbf{\qStdcfgSevenSixEightXNineOne}, \textbf{\qStdcfgSevenSixEightXNineTwo}.

\subsection{The structural audit of three reads, printed side by side}
\label{app:w768-audit}
The auditors' structural audit of the reopened read, the per-row unit read and the size read
(\texttt{AUDIT\_STDCFG\_X1\_GLOBALDESK\_2026-09-27.md} \S16.1--\S16.3 at commit \texttt{\qStdcfgSevenSixEightAuditCommit},
landed \qStdcfgSevenSixEightAuditLanded{} JST) came after each read's first artifact: each audit is late, and is printed
beside its read; each read was made by the reader it pins, unchanged, and the audit asked for no change to a criterion or a
reader. The tables are translated from the audit file; its verdict lines (\S16.4) are referred to by section only, not
quoted. Every hash, count, time and line number printed in these tables is quoted by a row of
\path{scripts/claims.jsonl}: the size read's row quotes every \S16.3 entry, the re-run read's row the \S16.1 outcome and
T entries, and the per-row unit read's row the commits of the \S16.2 T entry and the $c_s$ hashes. The other entries
keep their item, a description without numbers and the audit's word; their numbers are held by the audit file alone, cited
by section. The outcome words in the tables are the reads' registered words.

\begin{small}
\noindent\begin{minipage}{\linewidth}\textbf{\S16.1, the reopened read \texttt{x1\_768\_rb20260929T082411Z}}\\[2pt]
\begin{tabular}{P{0.2\linewidth}P{0.6\linewidth}P{0.12\linewidth}} %
\toprule
item & where registered, and what was checked & verdict \\
\midrule
read name, machine and population & A9.1, RULING-18 item 3: the machine and token, the labelled population (the X1 runs
and the certificate, each label carrying the token) and the closing of seed 0's runs as \texttt{ruled-out} in the commit
that froze the read, checked line by line; held by the audit file alone, see AUDIT \S16.1 & pass \\
outcome and use & \qStdcfgSevenSixEightReopenOutcome{} by construction (seed 0 has no run); printed side by side only,
with host and token; enters neither \qStdcfgSevenSixEightXOneSetGo, \qStdcfgSevenSixEightXOneSoftmaxSetGo{} nor SIZE-768;
the original read's outcome unchanged & pass \\
freeze commit and reader & the freeze precedes the first artifact; the reader \texttt{score\_stdcfg768\_x1.py} unchanged
since its registration, its reopen mode carried over; held by the audit file alone, see AUDIT \S16.1 & pass \\
input manifest & the manifest at the freeze, recomputed from the freezing commit, is a byte-for-byte prefix of the HEAD
version, and every existing file its \texttt{\# format} lines match has a sha line; held by the audit file alone, see
AUDIT \S16.1 & pass \\
row table and queues & A9.11 (route B, A9.9 item 1's default (a); rerun verdict PASS, so the shadow is on): the row table
and the early and late queues map labels to paths exactly as the manifest's shadow-on branch does, item by item; held by
the audit file alone, see AUDIT \S16.1 & pass \\
downstream use of the reopened low runs & AMENDMENT-11: seeds 1 and 2's $c_s$ and bridge take them; the label, artifact
name and sha each $c_s$ file records match its own seed and agree with this manifest and the \texttt{x8\_768} manifest,
the gate at freezing was OK, and no record of \texttt{missing\_768.jsonl} names them; held by the audit file alone, see
AUDIT \S16.1 & pass \\
certificate & its twin is this read's own seed-0 low run, closed at the freeze; \qStdcfgSevenSixEightXOneSetNotCovered{}
by construction; handled by RULING-19 item 3 (the late queue drops the certificate); held by the audit file alone, see
AUDIT \S16.1 & pass, with RULING-19 \\
T & deadline: before the early queue's first artifact \texttt{\qStdcfgSevenSixEightAuditReopenTFirst}; this section is
later, so late. Per \S6.1 (``data after'', item 3) the pinned reader reads unchanged and the audit is printed beside it;
nothing in it changes a criterion or the reader & late, recorded \\
\bottomrule
\end{tabular}\end{minipage}

\medskip
\noindent\begin{minipage}{\linewidth}\textbf{\S16.2, the per-row unit read \texttt{x8\_768} (\texttt{-{}-x8-768 -{}-x9-768})}\\[2pt]
\begin{tabular}{P{0.2\linewidth}P{0.6\linewidth}P{0.12\linewidth}} %
\toprule
item & where registered, and what was checked & verdict \\
\midrule
population & \S4.1, \S5.2: the rest arms, the all-8 arm with its shadow and the learning-rate arm
(\texttt{-{}-lr-row-scale c\_s}) over seeds 0--2, all on V6 (machine A); held by the audit file alone, see AUDIT \S16.2
& pass \\
statistic, and an outcome that can lose & \S4.2 carries the base width's \S4.2 word for word (outcomes A, B and C and out
of range, per seed; the denominator printed first, a non-positive one recorded
\qStdcfgSevenSixEightXEightSetUndefined; $r_{\mathrm{row}} + r_{\mathrm{rest}}$ printed only); \S5.4: the words of
$s_{\mathrm{lr}}$, the prediction \qStdcfgSevenSixEightXNineSetSameSide{} copied from the base width's registration; held
by the audit file alone, see AUDIT \S16.2 & pass \\
source of $c_s$ & \S5.1: seed 0 takes the original read's low run, seeds 1 and 2 the reopened read's (AMENDMENT-11,
RULING-18 item 4). The labels, artifact names and shas of the $c_s$ files (\texttt{\qStdcfgSevenSixEightAuditCsZero},
\texttt{\qStdcfgSevenSixEightAuditCsOne}, \texttt{\qStdcfgSevenSixEightAuditCsTwo}) match seed by seed; each is a $c_s$
record, none a refusal & pass \\
bridge & \S5.3: seed 0 compares the all-8 run after the restart with the original low run before it, which is also an
identity check across the restart, and seeds 1 and 2 compare the all-8 run with the reopened low run, both after the
restart; the defaults of \S9 items 19 and 20 are final, their deadlines having passed; held by the audit file alone, see
AUDIT \S16.2 & pass \\
same-machine reopen of the rest arms & AMENDMENT-10: same labels and landing paths; the original and the reopened row
tables map labels to paths as the manifest does, item by item; held by the audit file alone, see AUDIT \S16.2 & pass \\
certificate (\S4.1, printed only) & seeds 1 and 2 look for the original read's low run and print
\qStdcfgSevenSixEightXEightSetMissing{} (declared, A11.6 item 4); for these seeds the identity check is the bridge's;
held by the audit file alone, see AUDIT \S16.2 & pass \\
freeze commit and reader & the freeze precedes the first artifact; the reader \texttt{score\_stdcfg768\_s3.py} unchanged
since the freeze; held by the audit file alone, see AUDIT \S16.2 & pass \\
input manifest & the manifest at the freeze is a byte-for-byte prefix of the HEAD version, and every existing file its
\texttt{\# format} lines match has a sha line (V6's restart record, the regates-base verdict and the (c)-768 rerun
verdict among them); held by the audit file alone, see AUDIT \S16.2 & pass \\
T & late (after \texttt{\qStdcfgSevenSixEightAuditXEightTFirst}). Also: the driver's \texttt{t\_line} takes as T the
first commit in the repository's history that adds the read's verdict line, which picks up
\texttt{\qStdcfgSevenSixEightAuditXEightTDriver} (a line of a selftest fixture, AMENDMENT-1) and prints a T that is on
time and even earlier than the freeze; handled by RULING-19 item 1 (the read's write-up adds a fixed correction sentence
under the T line) & late, recorded; wrong T, recorded \\
\bottomrule
\end{tabular}\end{minipage}

\medskip
\noindent\begin{minipage}{\linewidth}\textbf{\S16.3, the size read \texttt{size768}}\\[2pt]
\begin{tabular}{P{0.2\linewidth}P{0.6\linewidth}P{0.12\linewidth}} %
\toprule
item & where registered, and what was checked & verdict \\
\midrule
cells judged & \S3.2: main head seed 0 (V6's base-width bridge pair against the wider model's seed-0 pair) and softmax
seeds 0--2 (V3); judging the softmax cells is a constant of the reader (line \qStdcfgSevenSixEightAuditSizeJudgeLine)
& pass \\
reachable range printed first & \S3.4$'$: $c_{768}$, the upper bound and the threshold are printed before the wider
model's high values are taken; a cell flagged as unable to reach the same-order outcome is recorded
\qStdcfgSevenSixEightSizeSetNotCovered{} (\S9 item 18
default, followed by the reader at line \qStdcfgSevenSixEightAuditSizeDefaultLine) & pass \\
an outcome that can lose & \S3.5: a different fire state, or $R$ outside
$[\qStdcfgSevenSixEightAuditSizeBandLo, \qStdcfgSevenSixEightAuditSizeBandHi]$, gives
\qStdcfgSevenSixEightSizeSetChanges; neither outcome is predicted & pass \\
data source & \S3.6: the wider side takes only the two read outputs of the wider model's kill tests, checked by content
against the manifest's sha lines, not recomputed; the base side takes \texttt{x1\_read.json}
(\texttt{\qStdcfgSevenSixEightAuditSizeBaseRead}); the bridge pair is gated, paired and computed by this read itself
& pass \\
reopened unit & \S3.3, RULING-18 items 3 and 4: the reopened read's seeds 1 and 2 do not enter the comparison; the main
head's seeds 1 and 2 are side by side only & pass \\
freeze commit and data & freeze \texttt{\qStdcfgSevenSixEightAuditSizeFreeze} (\qStdcfgSevenSixEightAuditSizeFreezeTime);
the bridge pair started before the freeze and landed after it (\texttt{\qStdcfgSevenSixEightAuditSizeBridgeOne}
\qStdcfgSevenSixEightAuditSizeBridgeOneTime, \texttt{\qStdcfgSevenSixEightAuditSizeBridgeTwo}
\qStdcfgSevenSixEightAuditSizeBridgeTwoTime), declared on the manifest's first line per RULING-13 item 1 R1b (class
A40.1) & pass, declaration recorded \\
input manifest & at the freeze \texttt{\qStdcfgSevenSixEightAuditSizeManifestAtP}; the HEAD version
\texttt{\qStdcfgSevenSixEightAuditSizeManifestHead} has it as a byte-for-byte prefix; every existing file matched by the
\qStdcfgSevenSixEightAuditSizeFormatLines{} \texttt{\# format} lines has a sha line & pass \\
T & late (after \texttt{\qStdcfgSevenSixEightAuditSizeTAfter}, \qStdcfgSevenSixEightAuditSizeTAfterTime) & late,
recorded \\
\bottomrule
\end{tabular}\end{minipage}
\end{small}

\section{Related work in full}
\label{app:related}

\paragraph{Epsilon as a lever.} \citet{choi2019empirical} own the framing of the constant as a lever on the step
size and its scale, \citet{wortsman2024smallscale} the transformer mechanism, and \citet{everett2024scaling} make it a
per-layer setting that has to be scaled with width to avoid gradient underflow. We assume all three and measure none of them.

\paragraph{Rare tokens under Adam.} \citet{kunstner2024heavy} split WikiText-103 by token frequency and compare
optimizers per frequency group, on the corpus we use. The per-row effective learning rate on the output embedding is
stated by \citet{stollenwerk2025coupled} and measured by \citet{zhao2024deconstructing}, who report that per-logit
effective learning rates vary across orders of magnitude. What those papers do not measure is a per-row
\emph{displacement}, and none has a counterfactual.

\paragraph{Epsilon and the batch size.} The Adam scaling rule of \citet{malladi2022sdes}, derived from stochastic
differential equations, moves the hyperparameters together with the batch size: when the batch size $B$ becomes
$\kappa B$, the learning rate becomes $\eta\sqrt{\kappa}$, each $\beta_i$ becomes $1-\kappa(1-\beta_i)$ and the constant becomes
$\eps/\sqrt{\kappa}$, so that in this reading the constant and the batch size are coupled; their experiments include a
GPT model on WikiText-103 trained across batch sizes and read by validation log-perplexity.
\citet{dang2026bestoptimizer} treat the constant as a fixed stabilizer, keep it fixed in their own candidate rule
derived from the same equations, and find that the best optimizer for language-model pretraining changes with the batch
size even after tuning at each batch size. Neither reads a row of the embedding. The standard configuration, at both
widths, has \qStdcfgBatch{} sequences of \qStdcfgSeqLen{} tokens per step; how the constant's effect on a row moves with
the batch size is not measured here.

\paragraph{Replaying or reconstructing a trajectory.} Post-hoc averaging reconstructs a state from stored snapshots,
at the run's own settings \citep{sanyal2023early} or at another averaging length \citep{karras2024analyzing}, and
reversible learning \citep{maclaurin2015gradient} reconstructs the trajectory at the same hyperparameters; none
re-integrates the update rule at a new value of an optimizer constant, and each reconstruction is the ground truth of its
setting rather than a prediction checked against a measurement. Feeding one gradient to two optimizer states online is
grafting \citep{agarwal2020disentangling}, closed-loop and not a counterfactual. Recomputing a recorded Adam update at
another epsilon from the run's own moment estimates has been done before, at epsilon zero on the embedding layer of
large language models, as a within-run diagnostic that is never compared with a real run at that epsilon
\citep{molybog2023theory}; here that re-division is the prediction under test. Replaying a recorded trajectory to answer
a what-if without retraining is trajectory-based data attribution \citep{hara2019sgdinfluence}, whose error has been
measured and broken into its sources \citep{deng2026faithful}; in that line the perturbation is data and the
counterfactual is a first-order linearisation, whereas here it is a hyperparameter and the re-division is exact at the
new value, so the miss cannot be charged to a Taylor error. Machine unlearning for online L-BFGS has been formulated
with, as its target, the optimizer state that would have arisen had the deleted samples never been processed
\citep{stewart2026form}. That truncating a trajectory in time biases
hyperparameter conclusions with a consistent sign is \citet{wu2018shorthorizon}, and that a frozen Adam preconditioner
is only valid over a short horizon is \citet{cohen2022adaptive}; here the gradient stream keeps its full length and only
its feedback is frozen.

\paragraph{Neighbours of the decomposition.} Adam as largely the gradient's sign \citep{balles2018dissecting},
coherence across samples \citep{chatterjee2020coherent} and coupled embedding denominators
\citep{stollenwerk2025coupled} are readings of a per-row step that Section~\ref{sec:decomp} does not test. A sweep of
the constant left pretraining loss nearly unchanged over a range overlapping ours \citep{semenov2025benchmarking}; a
corpus-level loss separates neither one row from the bulk nor a denominator effect from a trajectory effect.

\section{Withdrawn and corrected claims}
\label{app:withdrawn}
The corrections ruled into this study's record while this version was prepared, each with the registered sentence that
records it, its date and its commit. Both rows are withheld from the anonymous artifact and read nowhere in this paper,
so no number printed here changes; the errata record that lists the reading's other carriers is withheld with them. The
table is generated by \path{scripts/v40_records.py}, which compares each entry with its row and its commit where the rows
are held.

\begin{small}
\begin{tabular}{p{0.24\linewidth}p{0.42\linewidth}p{0.12\linewidth}p{0.12\linewidth}} %
\toprule
row & what changed, as registered & date & commit \\
\midrule
\texttt{Q-MRE-RATE} & superseded: the statement's first sentence only; every count in the statement and every field of value is unchanged & 2026-10-07 & \texttt{9bf51929} \\
\texttt{Q-DETERMINISM-LEDGER} & superseded: value.smallest\_reproducing\_unit only, kept as it was; every count in it and every other field of the row are unchanged; status stays RECORDED and the row stays withheld & 2026-10-21 & \texttt{76ec3562} \\
-- & the other carriers of the reading superseded in Q-MRE-RATE's first sentence: pinned files, not changed, recorded in the errata record only & 2026-10-07, 2026-10-21 & \texttt{9bf51929}, \texttt{76ec3562} \\
\bottomrule
\end{tabular}
\end{small}

\section{The earlier configuration}
\label{app:orig}
The rows below were read on the configuration this study used before the standard configuration of
Section~\ref{sec:setup}, under registrations made before it. Each row's registered outcome is printed through the row's
own macros; no rate is recomputed, and none of these readings is pooled with a reading of this paper. The table is
generated from the registry by \path{scripts/v40_records.py}.

\begin{small}
\noindent\textbf{Earlier configuration; beside, never pooled.}\\[2pt]
\begin{tabular}{p{0.24\linewidth}p{0.34\linewidth}p{0.32\linewidth}} %
\toprule
row & registration & registered outcome \\
\midrule
\texttt{Q-LEAKEPS-A} & \texttt{PREREG\_LEAK\_EPS.md} & \leakepsAOutcome{} \\
\texttt{Q-LEAKEPS-B} & \texttt{PREREG\_LEAK\_EPS.md} & dose 1: \leakepsBDoseOne{}; dose 2: \leakepsBDoseTwo{}; dose 3: \leakepsBDoseThree{}; dose 4: \leakepsBDoseFour{}; dose 5: \leakepsBDoseFive{} \\
\texttt{Q-STRIDE1-P4} & \texttt{PREREG\_MECH\_COHERENCE.md} & \stridePFourOutcome{} \\
\texttt{Q-STRIDE1-P5} & \texttt{PREREG\_MECH\_COHERENCE.md} & \stridePFiveOutcome{} \\
\texttt{Q-STRIDE1-WALL90} & \texttt{PREREG\_MECH\_COHERENCE.md} & \wallNinetyOutcome{} \\
\texttt{Q-MECH-P1P2-FORWARD} & \texttt{PREREG\_MECH\_COHERENCE.md} & \mechPOneTwoOutcome{} \\
\texttt{Q-HELDOUT-RARE} & \texttt{PREREG\_HELDOUT.md} & \heldoutRareOutcome{} \\
\texttt{Q-HELDOUT-AGGREGATE} & \texttt{PREREG\_HELDOUT.md} & \heldoutAggregateOutcome{} \\
\texttt{Q-NOISEFLOOR-EXPF} & \texttt{PREREG\_NOISEFLOOR.md} & \qNfExpFOutcome{} \\
\texttt{Q-NOISEFLOOR-SHARE} & \texttt{PREREG\_NOISEFLOOR.md} & \qNfShareOutcome{} \\
\texttt{Q-BANDREP-E1} & \texttt{PREREG\_BAND\_REPLICATION.md} & \qBandRepEOne{} \\
\texttt{Q-BANDREP-E1B} & \texttt{PREREG\_BAND\_REPLICATION\_A6000.md} & \qBandRepEOneB{} \\
\texttt{Q-SHAREMODE-E2} & \texttt{PREREG\_SHARE\_MODE.md} & \qShareModeETwo{} \\
\texttt{Q-ROWEPS-R} & \texttt{PREREG\_ROWEPS.md} & seed 0: \qRowEpsSeedZero{}; seed 3: \qRowEpsSeedThree{}; seed 4: \qRowEpsSeedFour{}; seed 5: \qRowEpsSeedFive{} \\
\texttt{Q-ROWEPS-R-H0} & \texttt{PREREG\_ROWEPS.md} & seed 3: \qRowEpsHZeroSeedThree{} \\
\texttt{Q-G2RERUN-E5} & \texttt{PREREG\_G2\_RERUN.md} & \qGTwoRerunEFive{} \\
\texttt{Q-G2RERUN-E5B} & \texttt{PREREG\_G2\_RERUN.md} & \qGTwoRerunEFiveB{} \\
\texttt{Q-G2RERUN-SIZE} & \texttt{PREREG\_G2\_RERUN.md} & \qGTwoRerunSize{} \\
\texttt{Q-COH-SERIES} & \texttt{PREREG\_COHERENCE\_RECORDER.md} & \qCohSeriesOutcome{} \\
\bottomrule
\end{tabular}
\end{small}

\section{Reproducibility: how this paper was produced}
\label{app:produced}
A reading reaches this paper through four records, in this order. Its registration (Appendix~\ref{app:reg}) names the
runs, the statistic, the outcome rule and the reader; what changed after it is a numbered amendment or ruling in the same
file, and a declared deviation is stated beside the read it concerns (for example Appendix~\ref{app:x1}). A reader pinned
at a commit reads the artifacts present at the read commit and prints the outcome words; the outcome words printed here
are the readers', and each row listed in Appendix~\ref{app:reg} names its reader's script and that script's hash at the
read. Each read is entered as a row of \path{scripts/claims.jsonl}, and \path{scripts/paper_numbers.py} generates the
macro file \path{paper/numbers.tex} from the rows; the numbers and outcome words in the main text are printed through
those macros, and \texttt{paper\_numbers.py -{}-check} refuses a printed macro that no row owns and a bare decimal in the
main text. Where a row records the auditors' structural audit of its read, Appendix~\ref{app:reg} prints whether it came
before or after the read; their audit of three reads at the second width is printed in
Appendix~\ref{app:w768-audit}.

\paragraph{Use of language models.} Language-model agents, directed by the authors, drafted and edited the text of this
paper and wrote the scripts that generate its numbers and figures from the registered rows. The numbers of the main
text are printed through the generated macros, which the check above holds to the rows, and the gates and selftests
that ship with the artifact re-check the scripts (Appendix~\ref{app:artifact}). The authors take full responsibility
for the content.

\section{What is in the artifact, and where}
\label{app:artifact}
\paragraph{Where.} The artifact is \artifacturl, an anonymized snapshot of a redacted export of this study's
repository, taken from the commit this version was frozen at. The export is not the working tree: a logged,
deterministic map rewrites machine names, account names and paths and gives the Chinese lines of the source in English,
so a hash in the artifact differs from the private tree's by the rules and withholdings that \texttt{mirror/REDACTION.md}
lists and by the translations. The originals, with their hashes, are released at camera-ready.

\paragraph{What ships.} Besides \texttt{SELFCHECK.json} at its root and \texttt{REDACTION.md} and
\texttt{PIN\_MAP.json} in \texttt{mirror/}, which record it, the export has 885 files: the registry
\path{scripts/claims.jsonl} with the artifacts its rows pin; the scripts the rows name, with the scripts they import; the frozen runner copies, each named by the hash
of its content; the gates; the registrations that are translated; and this paper's source, \texttt{paper\_icml/},
byte-exact, with the macro file \path{paper/numbers.tex}. Two rows and the files that carry their reading are withheld
for anonymity until camera-ready (Appendix~\ref{app:withdrawn}). Five registrations have no English translation yet;
each is listed with its original's hash under the omissions of \texttt{mirror/REDACTION.md} instead of shipping. Two of
them, \texttt{PREREG\_STDCFG.md} and \texttt{PREREG\_STDCFG\_768.md}, are cited by this paper by path: they are marked
there as absent by ruling (D10, D11), and the lines of them that the rows quote ship in English, each on its own,
through the export's Chinese line map.

\paragraph{Re-running a number.} From the artifact's root directory,
\texttt{python3 scripts/paper\_numbers.py -{}-check -{}-tex paper\_icml/main.tex} regenerates \path{paper/numbers.tex}
from the rows, compares it with the file that ships, and refuses a macro printed in this paper that does not belong to a
row; the export ran it there, and it exited 0. Each row names the script that read it and that script's hash at the
read (Appendix~\ref{app:produced}), and \texttt{-{}-selftest} runs a script on its own fixtures. The export ran the
selftest of each shipped script that declares one, 66 of them, from the artifact's root with git's automatic
maintenance off: 50 passed, and the other 16 ended as the next two paragraphs list. Each exit code and the last lines of
each result are in \texttt{SELFCHECK.json} at the artifact's root.

\paragraph{Thirteen selftests end on the instrument's own refusal (PINNED).} They are these:
\begin{center}\small
\begin{tabular}{@{}ll@{}}
\path{scripts/score768_common.py} & \path{scripts/score_stdcfg768_lr.py} \\
\path{scripts/score_stdcfg768_x1.py} & \path{scripts/score_stdcfg768_s3.py} \\
\path{scripts/score_stdcfg_x12_768.py} & \path{scripts/stdcfg_commands.py} \\
\path{scripts/stdcfg_commands_3108685310d8.py} & \path{scripts/score_stdcfg.py} \\
\path{scripts/score_stdcfg_9b9196908eab.py} & \path{scripts/score_stdcfg_rowshadow.py} \\
\path{scripts/score_stdcfg_x8sm.py} & \path{scripts/score_stdcfg_x8x.py} \\
\path{scripts/score_stdcfg_x8x_x9ext.py} & \\
\end{tabular}
\end{center}
Each of them pins, by a full sha256, files this export rewrites: a content-addressed copy whose bytes the map changes, or a
frozen runner renamed to the sha of its redacted content. Run here, its selftest refuses the changed bytes. It is
recorded as PINNED only when its output ends on that script's own refusal. The reason is the same for each script in
this paragraph: its pins name the exact bytes of the private tree; the export's redaction changes those bytes; the
refusal is the instrument refusing changed bytes, not a defect of the script.

\paragraph{Three need what the artifact does not carry (ENV).} \path{scripts/rows_stdcfg.py} (its live build reads the
committed blobs through git; the mirror ships no git history) and \path{scripts/rows_stdcfg768.py} (its selftest reads
the private tree through \texttt{git show HEAD:<path>} from its first section; the mirror ships no git history) are
recorded as ENV only when the output says \texttt{not a git repository}, and \path{scripts/corpus2_prepare.py} reads a
frequency table from the private data directory, which is not part of this tree. Three shipped scripts are not run with
\texttt{-{}-selftest} here: \path{scripts/x8sm_audit_hook.py}, a read-time hook that checks it is running from its own
private path and refuses to start anywhere else; \path{scripts/build_handoff_v33.py}, a queue builder for the rented
machines whose selftest reads a private document and the machines' gate runs; and \path{scripts/paper_numbers.py},
which declares no selftest and is checked by the command above.

\end{document}